\documentclass[10pt,twocolumn,letterpaper]{article}

\usepackage{iccv}
\usepackage{times}
\usepackage{epsfig}
\usepackage{graphicx}
\usepackage{amsmath}
\usepackage{amssymb}
\usepackage{textcomp}
\usepackage{multirow}
\usepackage{subcaption}
\usepackage[dvipsnames]{xcolor}

\usepackage[pagebackref=true,breaklinks=true,colorlinks,bookmarks=false]{hyperref}

\iccvfinalcopy 

\def\iccvPaperID{12} 

\ificcvfinal\fi
\begin{document}


\title{Monocular 3D Object Detection with Pseudo-LiDAR Point Cloud}

\author{Xinshuo Weng\\
Carnegie Mellon University\\
{\tt\small xinshuow@cs.cmu.edu}
\and
Kris Kitani\\
Carnegie Mellon University\\
{\tt\small kkitani@cs.cmu.edu}
}

\maketitle

\begin{abstract}
\vspace{-0.3cm}
Monocular 3D scene understanding tasks, such as object size estimation, heading angle estimation and 3D localization, is challenging. Successful modern day methods for 3D scene understanding require the use of a 3D sensor. On the other hand, single image based methods have significantly worse performance. In this work, we aim at bridging the performance gap between 3D sensing and 2D sensing for 3D object detection by enhancing LiDAR-based algorithms to work with single image input. Specifically, we perform monocular depth estimation and lift the input image to a point cloud representation, which we call pseudo-LiDAR point cloud. Then we can train a LiDAR-based 3D detection network with our pseudo-LiDAR end-to-end. Following the pipeline of two-stage 3D detection algorithms, we detect 2D object proposals in the input image and extract a point cloud frustum from the pseudo-LiDAR for each proposal. Then an oriented 3D bounding box is detected for each frustum. To handle the large amount of noise in the pseudo-LiDAR, we propose two innovations: (1) use a 2D-3D bounding box consistency constraint, adjusting the predicted 3D bounding box to have a high overlap with its corresponding 2D proposal after projecting onto the image; (2) use the instance mask instead of the bounding box as the representation of 2D proposals, in order to reduce the number of points not belonging to the object in the point cloud frustum. Through our evaluation on the KITTI benchmark, we achieve the top-ranked performance on both bird's eye view and 3D object detection among all monocular methods, effectively quadrupling the performance over previous state-of-the-art. 
Our code is available at \url{https://github.com/xinshuoweng/Mono3D_PLiDAR}.
\end{abstract}


\begin{figure}[t]
\begin{center}
\begin{subfigure}{0.235\textwidth}
    \includegraphics[trim=0cm 2cm 17cm 2cm, clip=true, 
    width=\textwidth]{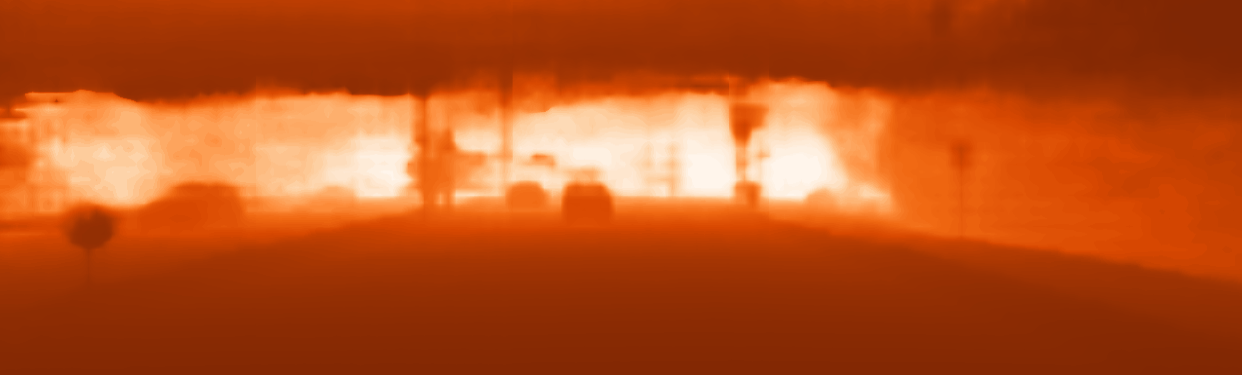}
    \vspace{-0.6cm}
    \caption{Monocular depth estimation}
    \label{fig:depth_teasor} 
\end{subfigure}
\begin{subfigure}{0.235\textwidth}
    \includegraphics[trim=0cm 2cm 17cm 2cm, clip=true, width=\textwidth]{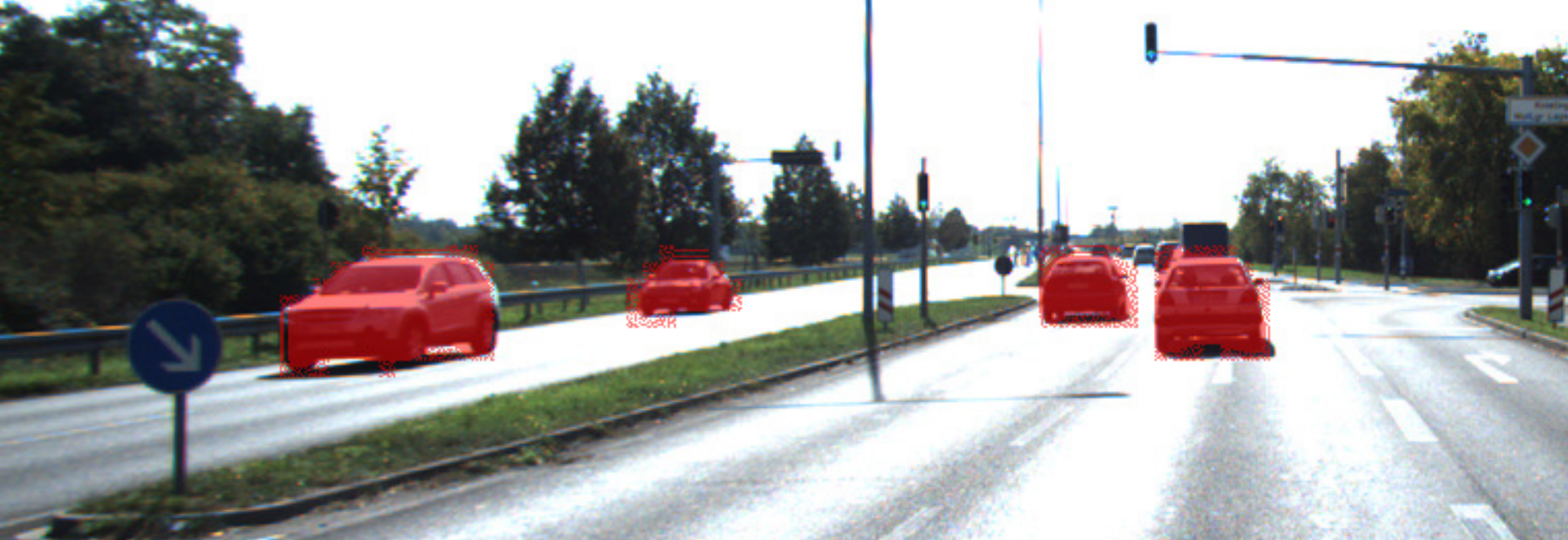}
    \vspace{-0.6cm}
    \caption{Instance segmentation}
    \label{fig:mask_teasor} 
\end{subfigure}
\\
   \includegraphics[trim=0cm 13cm 23.4cm 0cm, clip=true,
   width=0.06\linewidth]{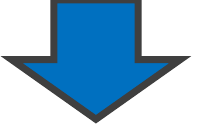}\\

\begin{subfigure}{0.4\textwidth}
    \includegraphics[trim=4cm 6.5cm 11cm 4cm, clip=true,
   width=\linewidth]{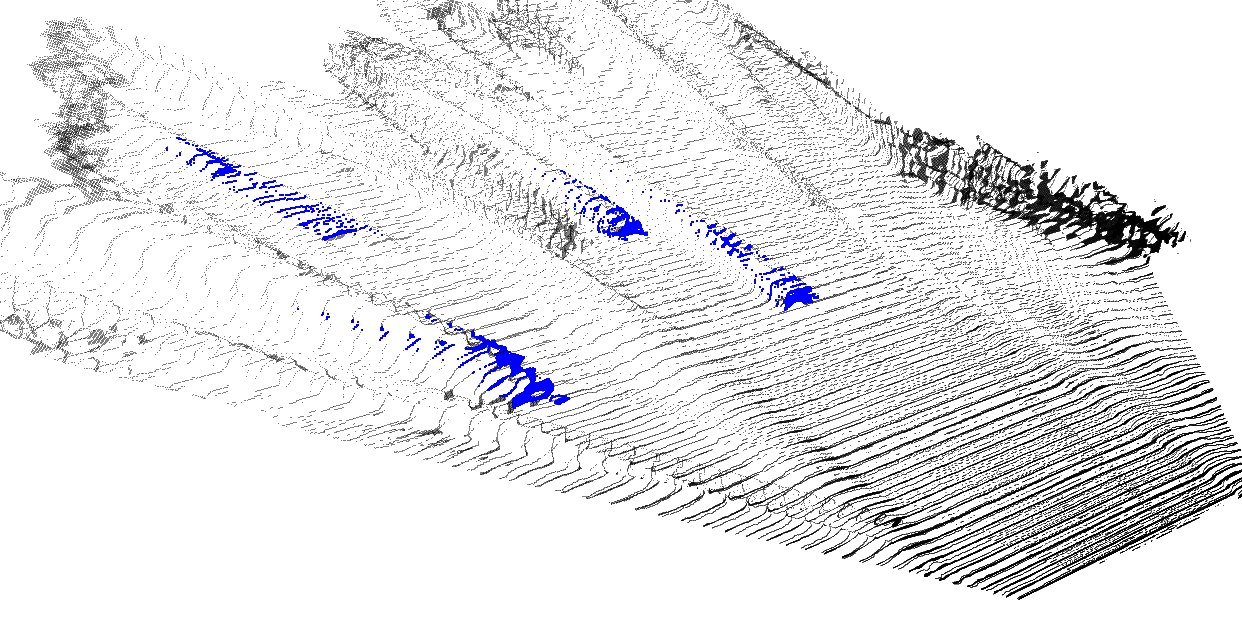}\\
    \vspace{-0.6cm}
    \caption{Point cloud frustums overlaid on the pseudo-LiDAR}
    \label{fig:frustum_teasor} 
\end{subfigure}
   
   \includegraphics[trim=0cm 13cm 23.4cm 0cm, clip=true,
   width=0.06\linewidth]{images/teaser/arrow.pdf}\\
   
\begin{subfigure}{0.235\textwidth}
   \includegraphics[trim=4cm 5cm 10cm 4cm, clip=true,
   width=\linewidth]{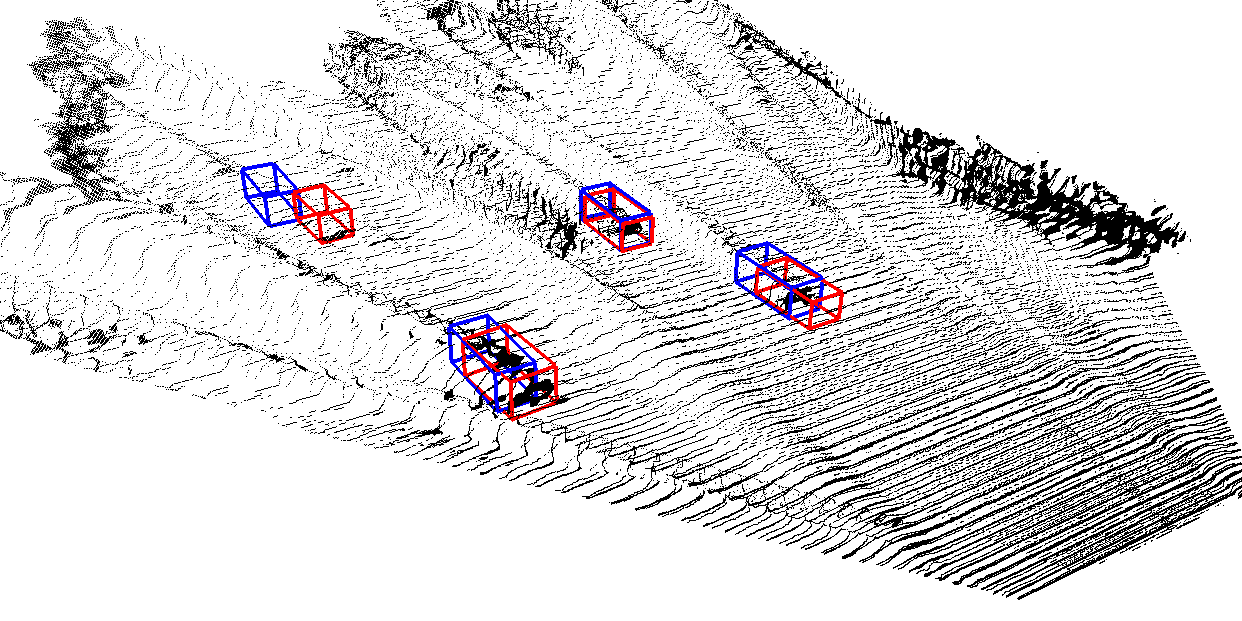}
    \vspace{-0.6cm}
    \caption{Results without BBC}
    \label{fig:results_before_bbc_teasor} 
\end{subfigure}
\begin{subfigure}{0.235\textwidth}
   \includegraphics[trim=4cm 5cm 10cm 4cm, clip=true,
   width=\linewidth]{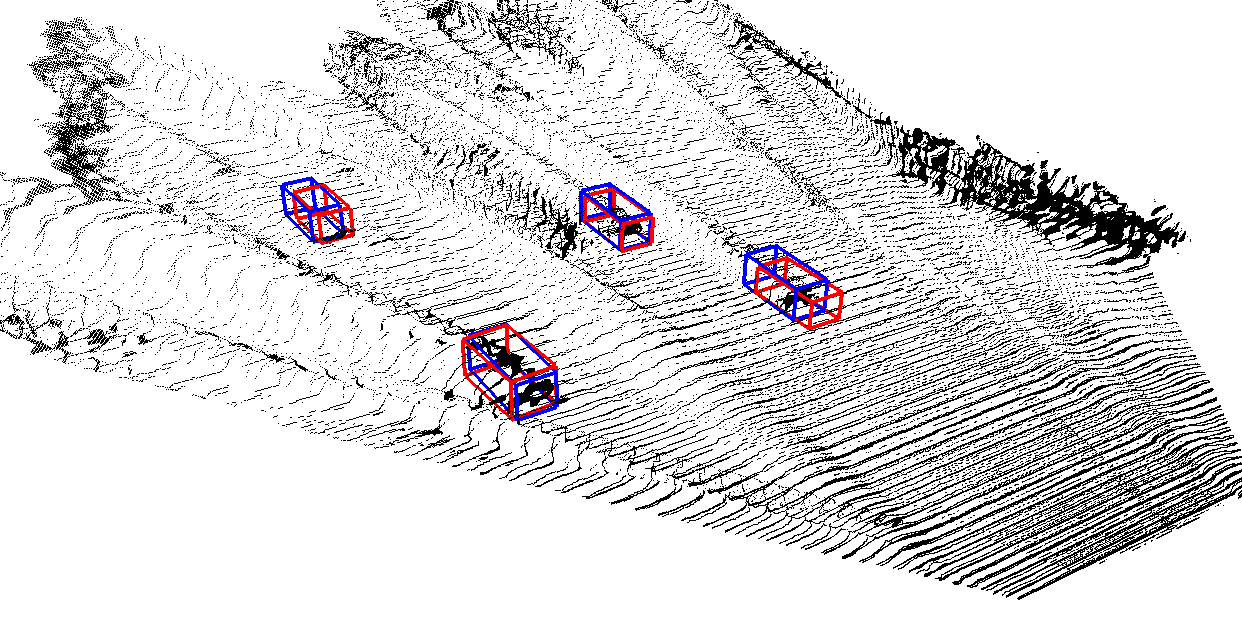}
    \vspace{-0.6cm}
    \caption{Results with BBC}
    \label{fig:results_after_bbc_teasor} 
\end{subfigure}

\end{center}

\vspace{-0.6cm}
\caption{(a) Monocular depth estimation and (b) Instance segmentation from a single input image; (c) Extracted point cloud frustums \textcolor{blue}{\textbf{(blue)}} overlaid on the pseudo-LiDAR \textbf{(black)}; 3D bounding box detection (\textcolor{blue}{\textbf{blue}}) results (d) without bounding box consistency (BBC) and (e) with BBC. Ground truth shown in \textcolor{red}{\textbf{red}}.}
\label{fig:teaser}
\vspace{-0.3cm}
\end{figure}


\vspace{-0.5cm}
\section{Introduction}
\vspace{-0.1cm}

3D object detection from a single image (monocular vision) is an indispensable part of future autonomous driving \cite{Wang2018} and robot vision \cite{Manglik2019} because a single cheap onboard camera is readily available in most modern cars. 
Successful modern day methods for 3D object detection heavily rely on 3D sensors, such as a depth camera, a stereo camera or a laser scanner (\emph{i.e.}, LiDAR), which can provide explicit 3D information about the entire scene.
The major disadvantages of this category of methods are: (1) the limited working range of the depth camera depending on the baseline; (2) the calibration and synchronization process of the stereo camera, causing it hard to scale on most modern cars; (3) the high cost of the LiDAR, especially when a high-resolution LiDAR is needed for detecting faraway objects accurately. 

On the other hand, a single camera, although cannot provide explicit depth information, is several orders of magnitude cheaper than the LiDAR and can capture the scene clearly up to approximately 100 meters.
Although people have explored the possibility of monocular 3D object detection for a decade \cite{Zia2015, Chen2016, Zia2013, PayendeLaGaranderie2018, Zia2014, Song2015, Fidler2012, Xu2018, Mousavian2017, Oberweger2018, Kundu2018}, state-of-the-art monocular methods can only yield drastically low performance in contrast to the high performance achieved by the LiDAR-based methods (\emph{e.g.}, 13.6\% average precision (AP) \cite{Xu2018} vs. 86.5\% AP \cite{Ku2018} on the moderate set of cars of KITTI \cite{Geiger2012} dataset). 

In this paper, we aim at bridging this performance gap between 3D sensing and 2D sensing for 3D object detection by extending LiDAR-based algorithms to work with single image input, without using the stereo camera, the depth camera, or the LiDAR. We introduce an intermediate 3D point cloud representation of the data, referred to as \emph{``pseudo-LiDAR''}\footnote{We use the same term as in \cite{Wang2019} for virtual LiDAR but we emphasize that this work is developed independently from \cite{Wang2019} and finished before \cite{Wang2019} is published. Also, it contains significant innovations beyond \cite{Wang2019}.}. Intuitively, we first perform monocular depth estimation and generate the pseudo-LiDAR for the entire scene by lifting every pixel within the image into its 3D coordinate given the estimated depth. Then we can train any LiDAR-based 3D detection network with the pseudo-LiDAR. 
Specifically, we extend a popular two-stage LiDAR-based 3D detection algorithm, Frustum PointNets \cite{Qi2018}. Following the same pipeline, we detect 2D object proposals in the input image and extract a point cloud frustum from the pseudo-LiDAR for each 2D proposal. Then an oriented 3D bounding box is detected for each frustum.

In addition, we observe that there is a large amount of noise in the pseudo-LiDAR compared to the precise LiDAR point cloud due to the inaccurate monocular depth estimation. This noise often reflects in two ways: (1) The extracted point cloud frustum might be largely off and there is a \emph{local misalignment} with respect to the LiDAR point cloud
.This may result in a poor estimate of the object center location, especially for the faraway objects with more severe misalignment; (2) The extracted point cloud frustum always has a \emph{long tail} -- depth artifacts around the periphery of an object stretching back into the 3D space to form a tail shape -- because the estimated depth is not accurate around the boundaries of the object. Therefore, predicting the object's size in 3D becomes challenging.

We propose two innovations to handle the above issues: (1) To alleviate the local misalignment, we use a 2D-3D bounding box consistency constraint, adjusting the predicted 3D bounding box to have a high overlap with its corresponding 2D detected proposals after projecting onto the image. During training, we formulate this constraint as a \emph{bounding box consistency loss} (BBCL) to supervise the learning. During testing, a \emph{bounding box consistency optimization} (BBCO) is solved subject to this constraint using a global optimization method to further improve the prediction results. (2) To cut off the long tail and reduce the number of points not belonging to the object in the point cloud frustum, we use the \emph{instance mask} as the representation of the 2D proposals as opposed to using the bounding box in \cite{Qi2018}. We argue that, in this way, the extracted point cloud frustum is much cleaner, and thus making it easier to predict the object's size.

\begin{figure*}
\begin{center}

\includegraphics[trim=0cm 3.3cm 0cm 0cm, clip=true, width=0.95\linewidth]{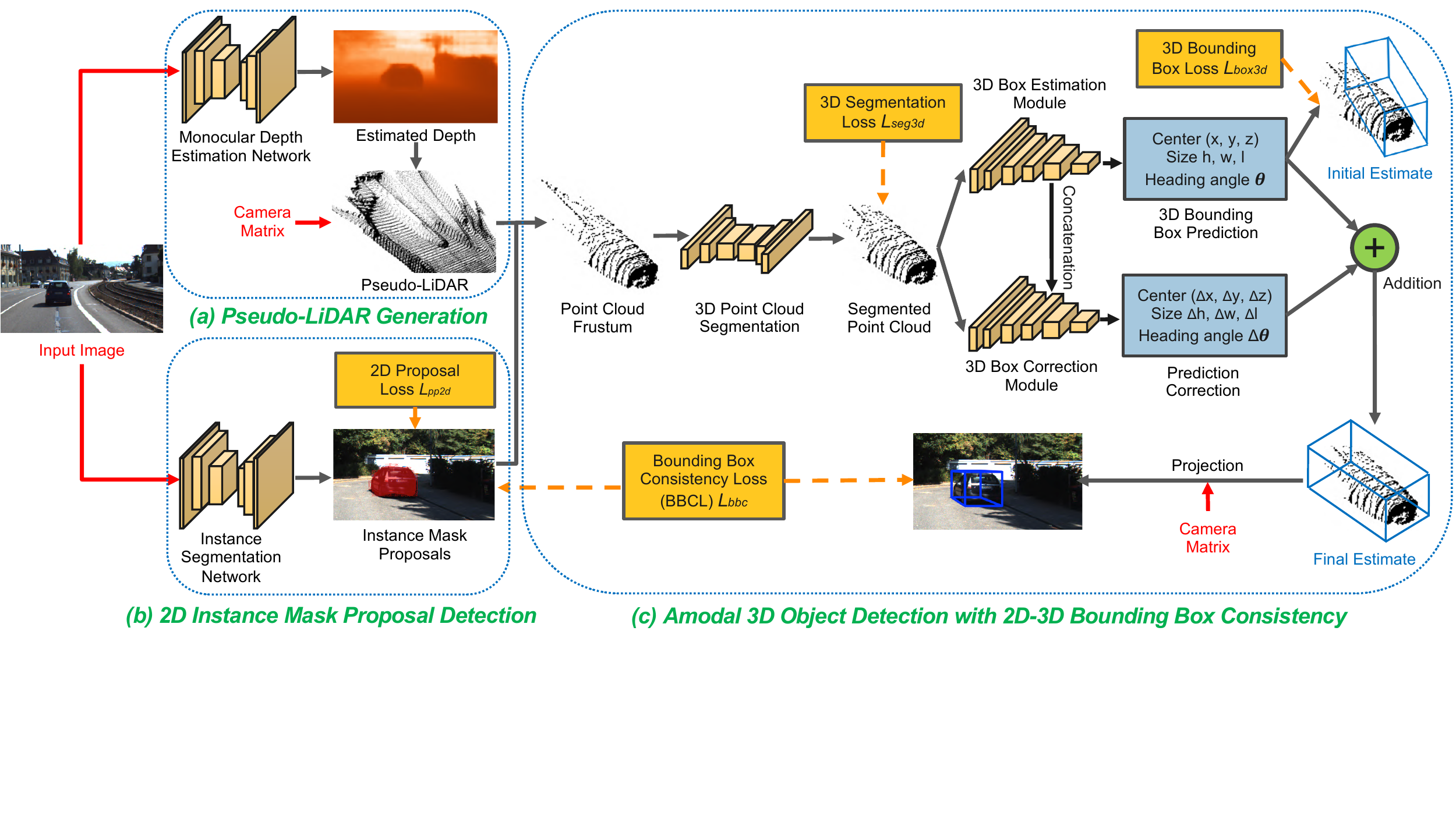}
\end{center}
\vspace{-0.6cm}
\caption{\textbf{Proposed Pipeline.} (a) Lift every pixel of input image to 3D coordinates given estimated depth to generate pseudo-LiDAR; (b) Instance mask proposals detected for extracting point cloud frustum; (c) 3D bounding box estimated \textcolor{blue}{\textbf{(blue)}} for each point cloud frustum made to be consistent with corresponding 2D proposal. Inputs and losses are in \textcolor{red}{\textbf{red}} and \textcolor{orange}{\textbf{orange}}.}
\label{fig:pipeline}
\vspace{-0.3cm}
\end{figure*}

Our pipeline is shown in Figure \ref{fig:pipeline}. To date, we achieve the top-ranked performance on bird's eye view and 3D object detection among all monocular methods on the KITTI dataset. For 3D detection in moderate class with IoU of 0.7, we raise the accuracy by up to \textbf{15.3\%} AP, nearly \textbf{quadrupling} the performance over the prior art \cite{Xu2018} (from 5.7\% by \cite{Xu2018} to 21.0\% by ours). We emphasize that we also achieve an improvement by up to \textbf{6.0\%} (from 42.3\% to 48.3\%) AP over the best concurrent work \cite{Wang2019} (its monocular variant), in moderate class with IoU of 0.5. 

Our contributions are summarized as follows: (1) We propose a pipeline of monocular 3D object detection, enhancing the LiDAR-based methods to work with single image input; (2) We show empirically that the bottleneck of the proposed pipeline is the noise in the pseudo-LiDAR due to inaccurate monocular depth estimation; (3) We propose to use a bounding box consistency loss during training and a consistency optimization during testing to adjust the 3D bounding box prediction; (4) We demonstrate the benefit of using instance mask as the representation of the 2D detected proposals; (5) We achieve the state-of-the-art performance and show an unprecedented improvement over all monocular methods on standard 3D object detection benchmark.


\vspace{-0.2cm}
\section{Related Work}
\vspace{-0.1cm}
\noindent\textbf{LiDAR-Based 3D Object Detection.}
Existing works have explored three ways of processing the LiDAR data for 3D object detection: (1) As the convolutional neural networks (CNNs) can naturally process images, many works focus on projecting the LiDAR point cloud into the bird's eye view (BEV) images as a pre-processing step and then regressing the 3D bounding box based on the features extracted from the BEV images \cite{Beltran2018, Wirges2018, Wirges2019, Liang2018, Ku2018, Yangb2018, Xiaozhi2017, Yangbin2018}; (2) On the other hand, one can divide the LiDAR point cloud into equally spaced 3D voxels and then apply 3D CNNs for 3D bounding box prediction \cite{Luo2018, Yan2018, Zhou2018}; (3) The most popular approach so far is to directly process the LiDAR point cloud through the neural network without pre-processing \cite{Lahoud2017, Du2018, Song2016, Yang2018, Xud2018, Shi2019, Shin2018, Song2014, Engelcke2017, Zhao2019, Gustafsson2018, Wangz2019, Qi2018, Lang2018}. To this end, novel neural networks that can directly consume the point cloud are developed \cite{Cherabier2017, Qi2017, Su2018, Yu2018, Misra2018, Wangyue2018, Graham2018}. Although LiDAR-based methods can achieve remarkable performance, they require that the high-resolution and precise LiDAR point cloud is available.

\vspace{2mm}\noindent\textbf{Monocular 3D Object Detection.} Unlike LiDAR-based methods requiring the precise LiDAR point cloud, monocular methods only require a single image, posing the task of 3D object detection more challenging. \cite{Chen2016} proposes to sample candidate bounding boxes in 3D and score their 2D projection based on the alignment with multiple semantic priors: shape, instance segmentation, context, and location. \cite{Manhardt2019} introduces a differentiable ROI lifting layer to predict the 3D bounding box based on features extracted from the input image and depth estimate. On the other hand, instead of estimating the pixel-wise depth for the entire scene, \cite{Qin2018} proposes a novel instance depth estimation module to predict the depth of the targeting 3D bounding box's center. In order to avoid using a coarse approximation (\emph{i.e.}, 3D bounding box) to the true 3D extent of objects, previous works \cite{Zia2015, Fidler2012, Oberweger2018, Zia2013, Chabot2017, Xiang2015, Zia2014, Kundu2018} have built fine-grained part-based models or leverage the existing CAD model collections \cite{Chang2015} in order to exploit rich 3D shape priors and reason about occlusion in 3D. \cite{PayendeLaGaranderie2018} enhances monocular 3D object detection algorithm to work with the image captured by 360\textdegree\ panoramic cameras. 

Models leveraging the 2D-3D bounding box consistency constraint are also related to our work. \cite{Mousavian2017} proposes to train a 2D CNN to estimate a subset of 3D bounding box parameters (\emph{i.e.}, the object's size and orientation). During testing, they combine these estimates with the constraint to compute the remaining of parameters, namely the object center location. As a result, the prediction of the object center location highly relies on the accuracy of the orientation and object size estimates. In contrast, we train a successful PointNet-based 3D detection network and learn to predict the complete set of parameters.
Also, we formulate the bounding box consistency constraint as a \emph{differentiable loss} during training and a \emph{constrained optimization} during testing to adjust 3D bounding box prediction. More importantly, we achieve an absolute AP improvement by up to 26.1\% over \cite{Mousavian2017} (from 5.6\% to an unprecedented 31.7\%)  --  a surprising 5$\times$ improvement in performance. 

The work of \cite{Wang2019} and \cite{Xu2018} both estimate the depth and generate a pseudo-LiDAR point cloud from the single image input for 3D detection.
We go one step beyond them by observing the local misalignment and long tail issues in the noisy pseudo-LiDAR and propose to use bounding box consistency constraint as a supervision signal and instance mask as the representation of the 2D proposals to mitigate the issues. We also show an absolute AP improvement by up to 21.2\% and 6.0\% over \cite{Xu2018} and \cite{Wang2019} respectively.

\vspace{2mm}\noindent\textbf{Supervision via Consistency.} Formulating a well-known geometry constraint to a differentiable loss for training not only provides a supervision signal for free but also makes the outputs of the model geometrically consistent with each other. \cite{Dong2018} proposes a registration loss to train a facial landmark detector, forcing the outputs are consistent across adjacent frames. \cite{Man2018, Yangz2018, Mahjourian2018, Yang2018aaai, Qixiao2018} jointly predict the depth and surface normal with a consistency loss forcing two outputs are compatible with each other. The multi-view supervision loss is proposed in \cite{Tulsiani2018, Sermanet2017, Tulsiani2017, Yao2018, Zhang2018, Jafarian2018}, making the prediction consistent across viewpoints. In addition, \cite{Zhu2017, Song2018, Weng2018, Chang2018, Bansal2018, Zhou2016} propose the cycle consistency loss, in the sense that if we translate our prediction into other domain and translate back, we should arrive back to the original input. In terms of consistency across dimensions, \cite{Tung2017, Kundu2018, Moreno2016} propose an inverse-graphics framework, which makes the prediction in 3D and ensures its 2D projection consistent with the 2D input. Similarly, our proposed BBCL forces the projection of the predicted 3D bounding box to be consistent with its 2D detected proposal.

\begin{figure}[t]
\begin{center}
\includegraphics[trim=0cm 2.5cm 0cm 0cm, clip=true, width=0.75\linewidth]{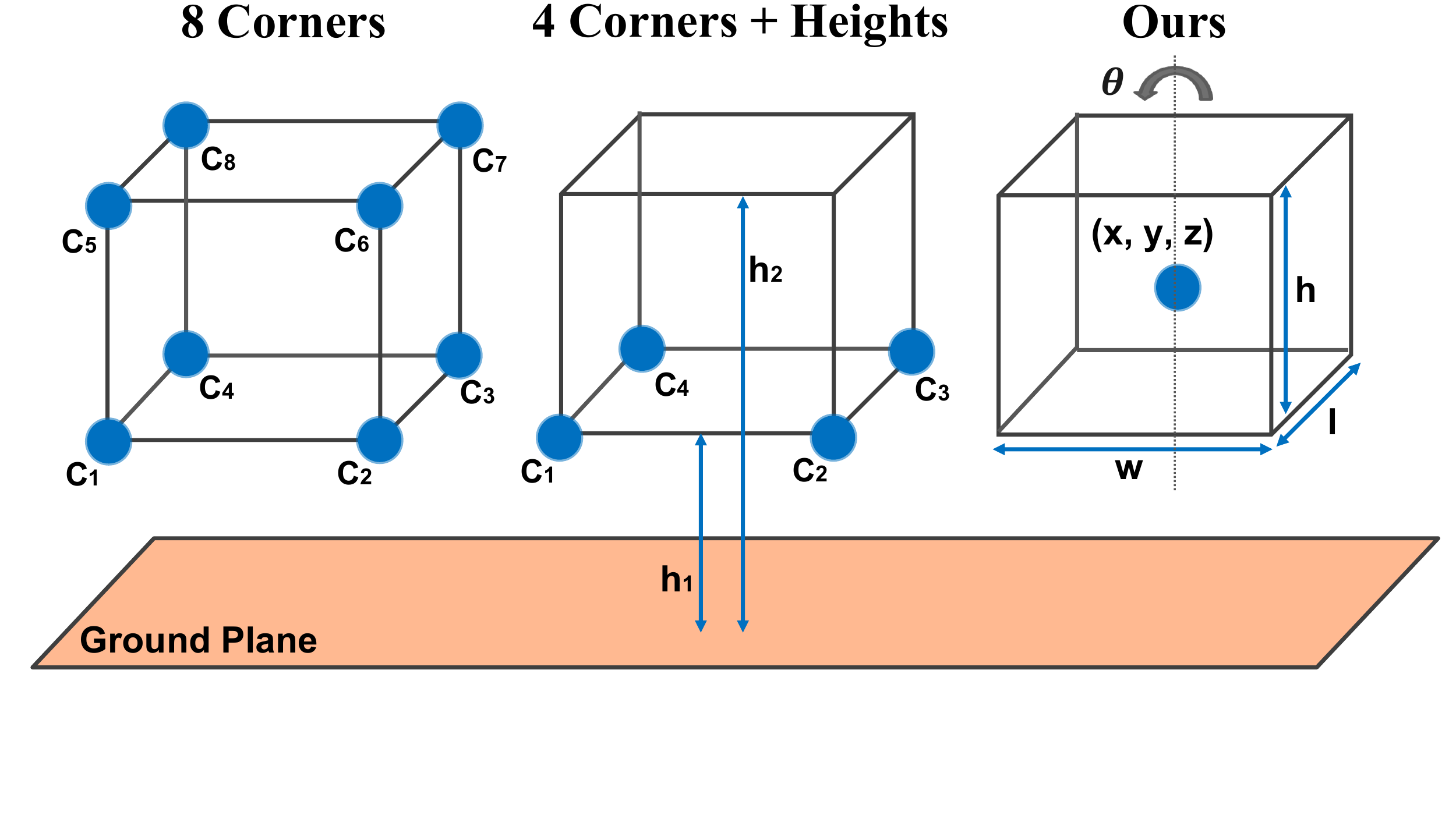}
\end{center}
\vspace{-0.6cm}
\caption{Comparison of the 3D bounding box parameterization between 8 corners \cite{Xiaozhi2017}, 4 corners with heights \cite{Ku2018} and ours. Our compact parameterization requires minimal number of parameters for an oriented 3D bounding box.}
\label{fig:parameterization}
\vspace{-0.3cm}
\end{figure}

\vspace{-0.2cm}
\section{Approach}
\vspace{-0.1cm}
Our goal is to estimate the oriented 3D bounding box of objects from only a single RGB image. During both training and testing, we do not require any data from the LiDAR, stereo and depth camera. The only assumption is that the camera matrix is known. Following \cite{Qi2018}, we parameterize our 3D bounding box output as a set of seven parameters, including the 3D coordinate of the object center ($x$, $y$, $z$), object's size $h$, $w$, $l$ and its heading angle $\theta$. Visualization of our parameterization compared to others is illustrated in Figure \ref{fig:parameterization}. We argue that our compact parameterization requires the minimal number of parameters for an oriented 3D bounding box.

In Figure \ref{fig:pipeline}, our pipeline consists of: (1) pseudo-LiDAR generation, (2) 2D instance mask proposal detection and (3) amodal 3D object detection with 2D-3D bounding box consistency. Based on the pseudo-LiDAR and instance mask proposals, point cloud frustums can be extracted, which are passed to train the amodal 3D detection network. The bounding box consistency loss and bounding box consistency optimization are used to adjust the 3D box estimate. 

\subsection{Pseudo-LiDAR Generation}
\noindent\textbf{Monocular Depth Estimation.}
To lift the input image to the pseudo-LiDAR point cloud, a depth estimate is needed. Thanks to the successful work called \emph{DORN} \cite{Fu2018},
we directly adopt it as a sub-network in our pipeline and initialize it using pre-trained weights. For convenience, we do not update the weights of the depth estimation network during training, and it can be regarded as an off-line module to provide the depth estimate. As our pipeline is agnostic to the choice of monocular depth estimation network, we can replace it with other networks if necessary.

\vspace{2mm}\noindent\textbf{Pseudo-LiDAR Generation.}
Our proposed pipeline can enhance the LiDAR-based 3D detection network to work with single image input, without the need for 3D sensors.
To this end, generating a point cloud from the input image that can mimic the LiDAR data is the essential step. Given the depth estimate and camera matrix, deriving the 3D location $(X_c, Y_c, Z_c)$ in the camera coordinate for each pixel $(u, v)$ is simply as:
\vspace{-0.2cm}
\begin{equation}
    X_c = \frac{(u - c_x)Z_c}{f_x}
\vspace{-0.35cm}
\end{equation}

\begin{equation}
    Y_c = \frac{(v - c_y)Z_c}{f_y}
\vspace{-0.1cm}
\end{equation}
where $Z_c$ is the estimated depth of the pixel in the camera coordinate and $(c_x, c_y)$ is the pixel location of the camera center. $f_x$ and $f_y$ are the focal length of the camera along $x$ and $y$ axes. Given the camera extrinsic matrix $C = [R \ t]$, one can also obtain the 3D location of the pixel in the world coordinate $(X, Y, Z)$ by computing $C^{-1} [X_c, Y_c, Z_c]^T$ and dividing by the last element. 
We refer to this generated 3D point cloud as \emph{pseudo-LiDAR}.

\begin{figure}[t]
\begin{center}
\includegraphics[trim=0cm 0cm 17cm 0cm, clip=true,width=\linewidth]{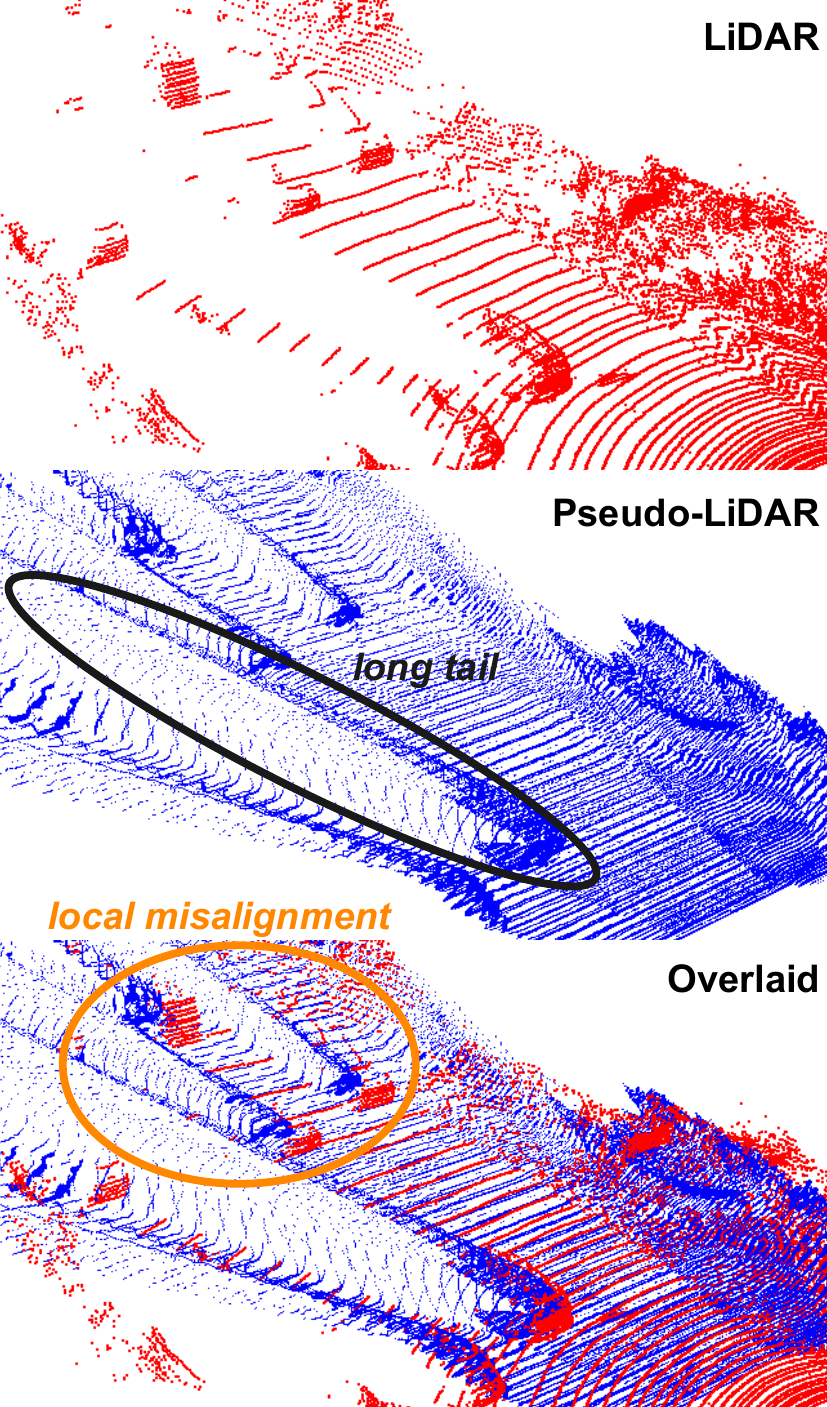}
\end{center}

\vspace{-0.5cm}
\caption{Comparison between the \textcolor{red}{\textbf{LiDAR}} (top), \textcolor{blue}{\textbf{pseudo-LiDAR}} (middle) and an overlaid version (bottom). Two types of noise discussed in Section \ref{sec:pseudovslidar} are indicated in \textcolor{orange}{\textbf{orange}} (local misalignment) and \textbf{black} (long tail) ellipses.}
\label{fig:pseudovslidar}
\vspace{-0.3cm}
\end{figure}

\vspace{2mm}\noindent\textbf{Pseudo-LiDAR vs. LiDAR Point Cloud.} \label{sec:pseudovslidar}
To make sure the pseudo-LiDAR is compatible with the LiDAR-based algorithms, it is natural to compare the pseudo-LiDAR with the LiDAR point cloud via visualization. An example is shown in Figure \ref{fig:pseudovslidar}. We observe that, although the generated pseudo-LiDAR aligns well with the precise LiDAR point cloud in terms of the \emph{global} structure, there is a large amount of \emph{local noise} in the pseudo-LiDAR due to inaccurate monocular depth estimation. This noise often reflects in two ways: (1) The extracted point cloud frustum might be largely off and there is a \emph{local misalignment} with respect to the LiDAR point cloud. This may result in a poor estimate of the object center location, especially for the faraway objects with more severe misalignment. For example, in the \textcolor{orange}{\textbf{orange}} eclipse of Figure \ref{fig:pseudovslidar}, the point cloud frustums fall behind their LiDAR counterpart; (2) The point cloud frustum extracted from the pseudo-LiDAR often has a \emph{long tail} because the estimated depth is not accurate around the boundaries of the object. Therefore, predicting the size of the objects becomes challenging. An example of point cloud frustum with the long tail is shown in the \textbf{black} eclipse of Figure \ref{fig:pseudovslidar}.

In addition, a distinction of the pseudo-LiDAR from the LiDAR point cloud is the density of the point cloud. Although a high-cost LiDAR can provide high-resolution point cloud, the number of LiDAR points is still at least one order of magnitude less than the pseudo-LiDAR point cloud. We will show how the density of the point cloud affects the performance in the experiment section.

\begin{figure}[t]
\begin{center}
\includegraphics[trim=4.2cm 2.6cm 15.2cm 4cm, clip=true, width=0.49\linewidth]{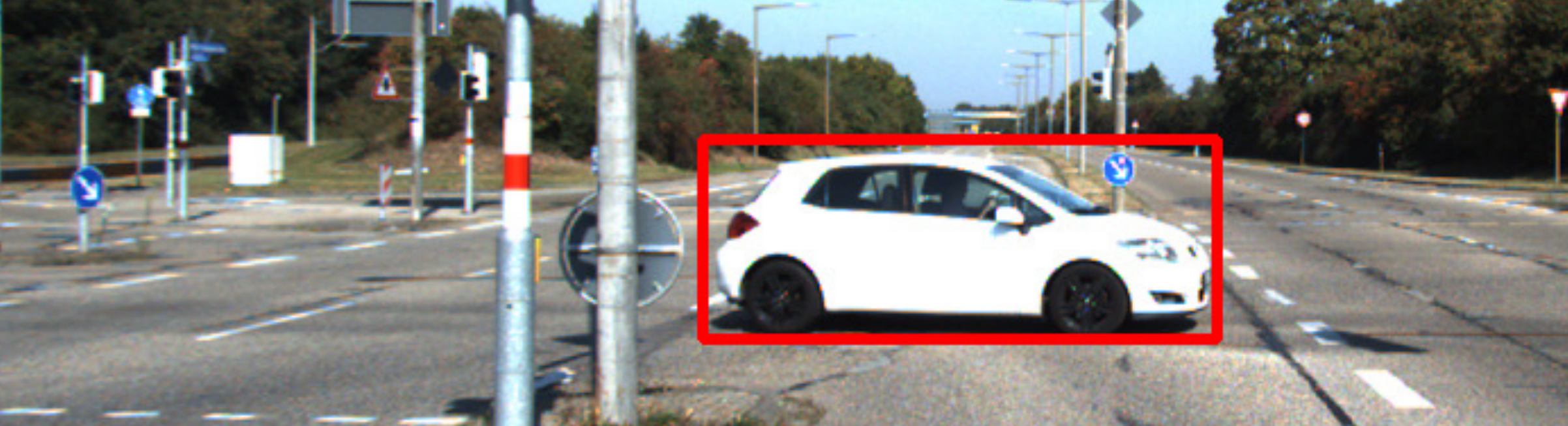}
\includegraphics[trim=4.2cm 2.6cm 15.2cm 4cm, clip=true, width=0.49\linewidth]{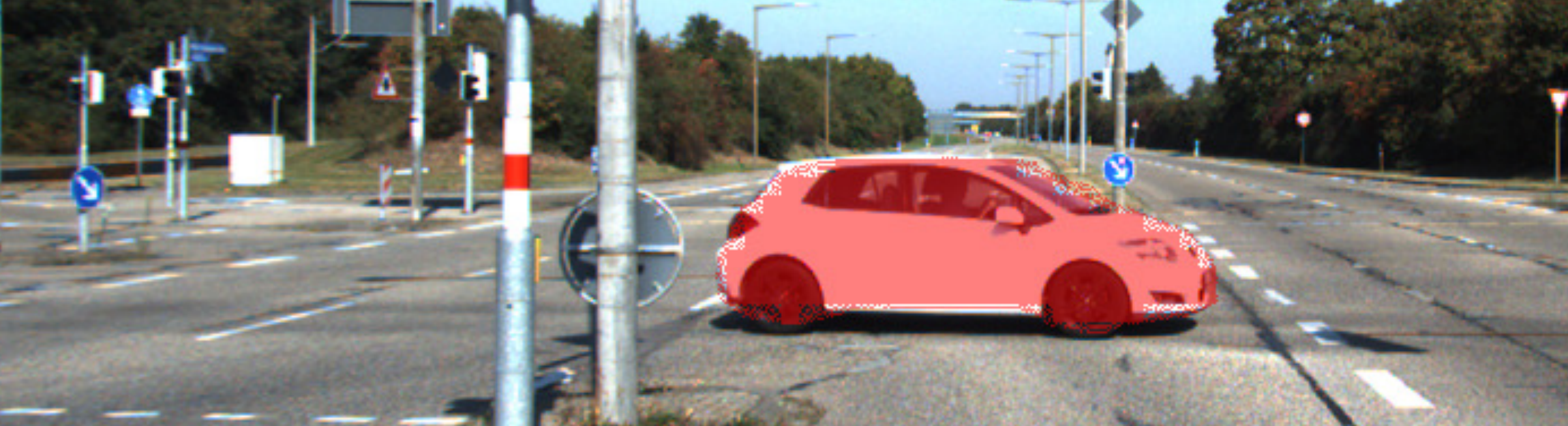}
\includegraphics[trim=18cm 5.2cm 6cm 8cm, clip=true, width=0.49\linewidth]{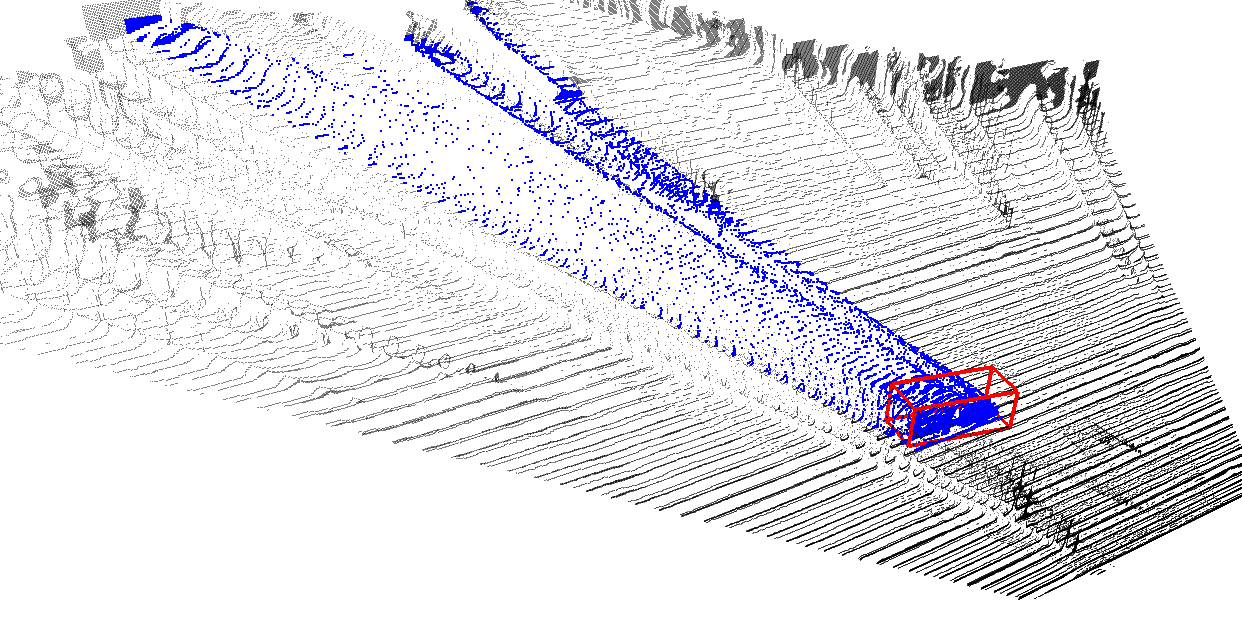}
\includegraphics[trim=18cm 5.2cm 6cm 8cm, clip=true,
width=0.49\linewidth]{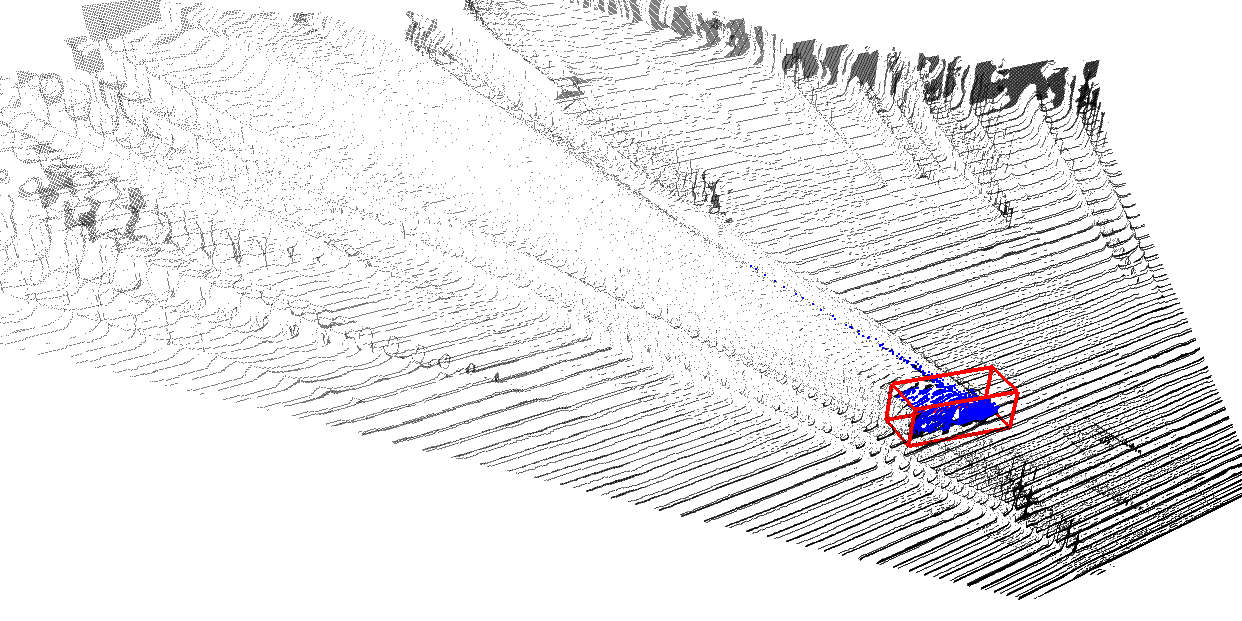}
\end{center}
\vspace{-0.5cm}
\caption{\textbf{Effectiveness of Instance Mask Proposal.} Top left: 2D box proposal. Top right: Instance mask proposal. Bottom left: Point cloud frustum lifted from 2D box proposal with noisy long tail. Bottom right: Point cloud frustum lifted from instance mask proposal with no tail. Ground truth box corresponding to the frustum shown in \textcolor{red}{\textbf{red}}.}
\label{fig:frustum}
\vspace{-0.3cm}
\end{figure}

\subsection{2D Instance Mask Proposal Detection}
In order to generate a point cloud frustum for each object, we first detect an object proposal in 2D. Unlike previous works using the bounding box as the representation of the 2D proposals \cite{Wangz2019, Qi2018, Wang2019, Xu2018}, we claim that it is better to use the instance mask, especially when the point cloud frustum is extracted from the noisy pseudo-LiDAR and thus has a large number of redundant points. We compare the generated point cloud frustum corresponding to the bounding box and instance mask proposal in Figure \ref{fig:frustum}. In the left column, we demonstrate that, when we lift all the pixels within the 2D bounding box proposal into 3D, the generated point cloud frustum has the \emph{long tail} issue as discussed in Section \ref{sec:pseudovslidar}. On the other hand, in the right column of Figure \ref{fig:frustum}, lifting only the pixels within the instance mask proposal significantly removes the points not being enclosed by the ground truth box, resulting in a point cloud frustum with no tail. Specifically, we consider the Mask R-CNN \cite{He2017} as our instance segmentation network. 

\subsection{Amodal 3D Object Detection}
Based on the generated pseudo-LiDAR and 2D instance mask proposals, we can extract a set of point cloud frustums, which are then passed to train a two-stage LiDAR-based 3D detection algorithm for 3D bounding box prediction. In this paper, we experiment with Frustum PointNets \cite{Qi2018}.
In brief, we segment the point cloud frustum in 3D to further remove the points not belonging to the objects. Then we sample a fixed number of points from the segmented point cloud for 3D bounding box estimation, including estimating the center ($x$, $y$, $z$), size $h$, $w$, $l$ and heading angle $\theta$. Please refer to the Frustum PointNets \cite{Qi2018} for details.

\begin{figure}[t]
\begin{center}
\includegraphics[trim=1.2cm 0.6cm 15.2cm 4cm, clip=true, width=0.49\linewidth]{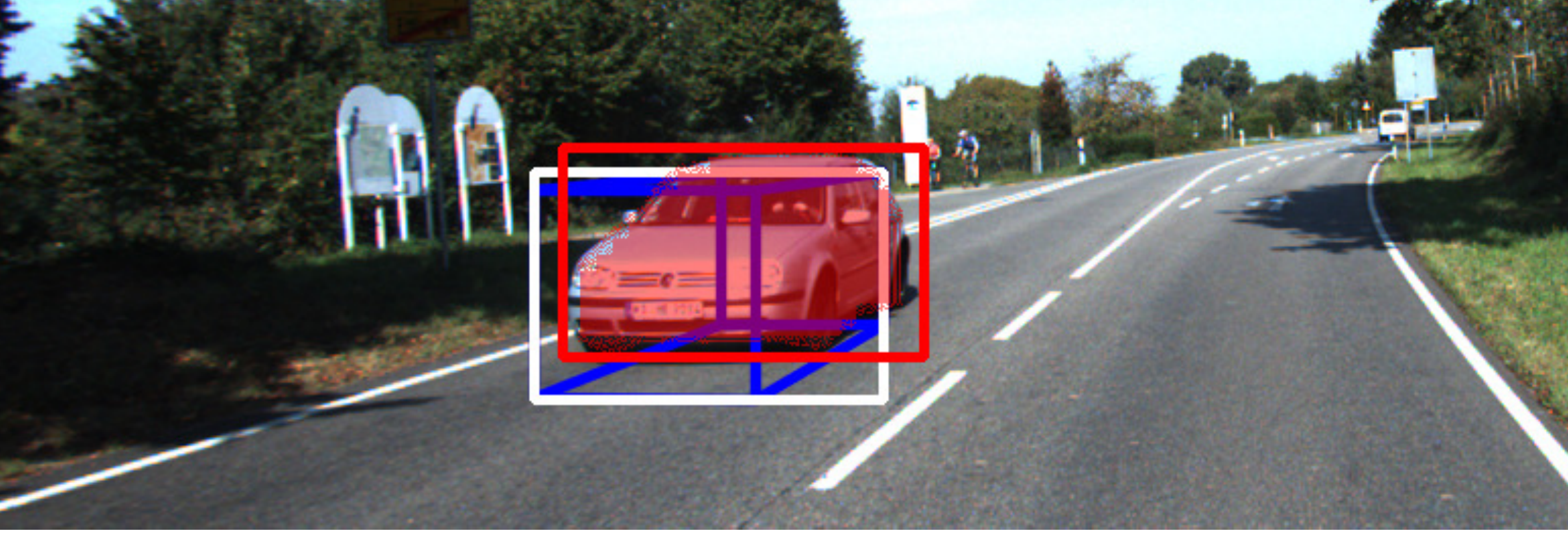}
\includegraphics[trim=1.2cm 0.6cm 15.2cm 4cm, clip=true, width=0.49\linewidth]{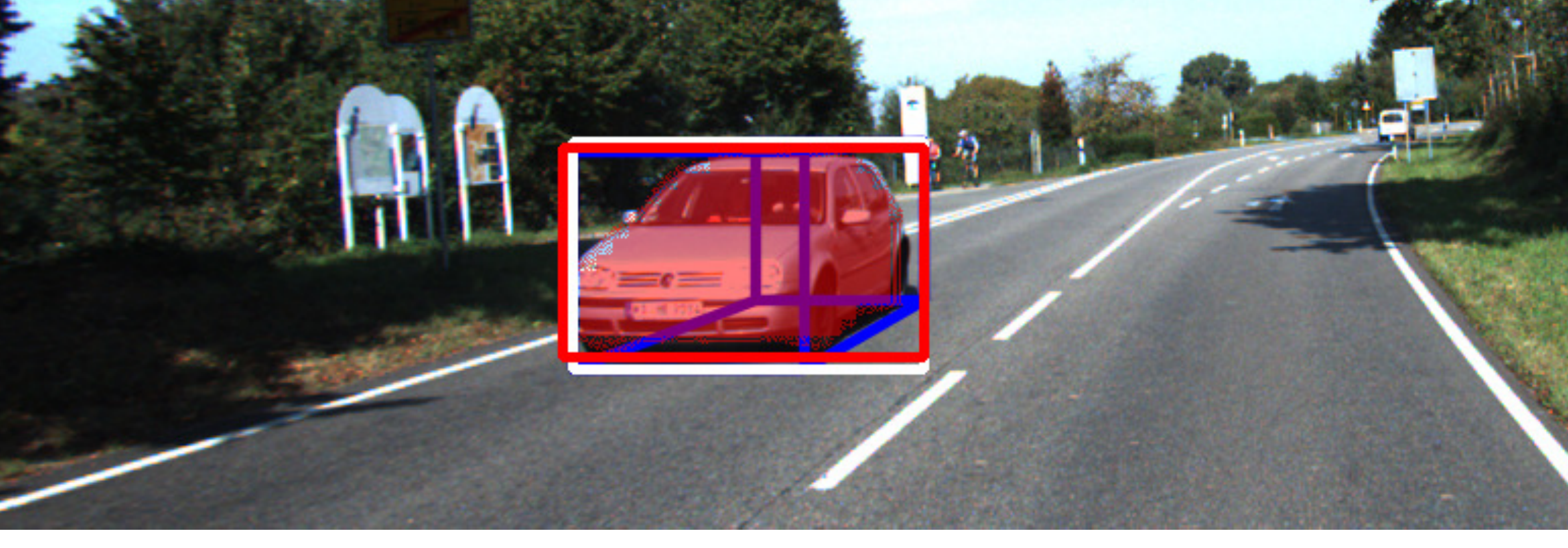}
\begin{subfigure}{0.235\textwidth}
    \includegraphics[trim=18cm 2.5cm 5cm 12cm, clip=true, width=\linewidth]{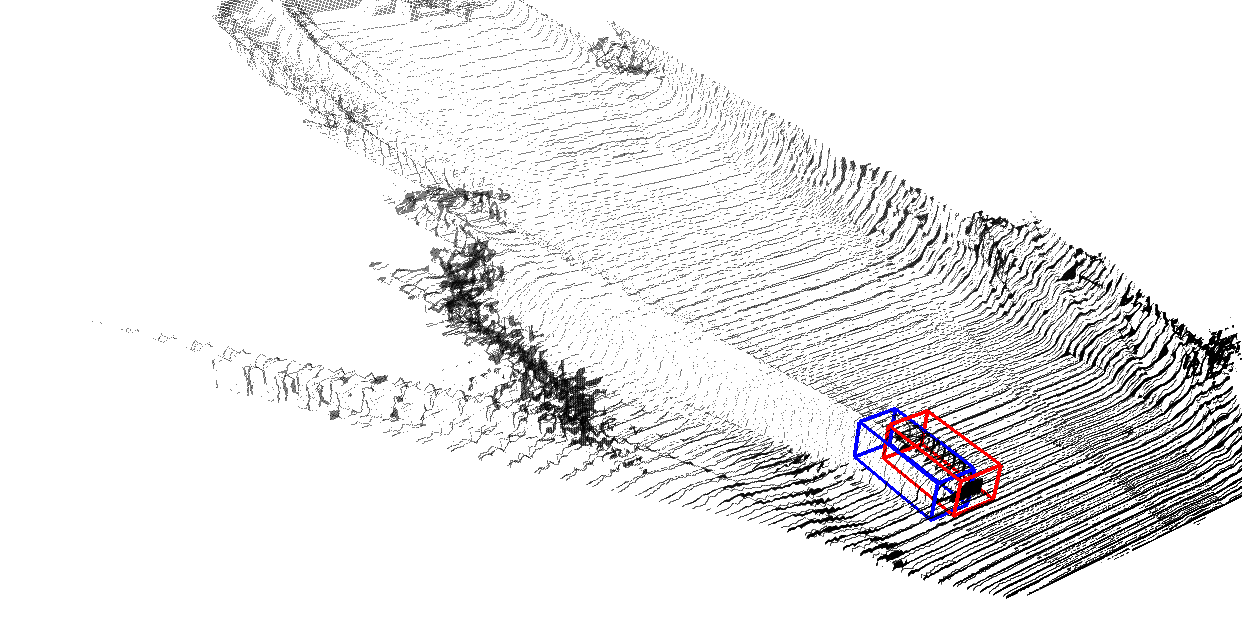}
    \vspace{-0.6cm}
    \caption{Results without BBC}
    \label{fig:before_bbc} 
\end{subfigure}
\begin{subfigure}{0.235\textwidth}
    \includegraphics[trim=18cm 2.5cm 5cm 12cm, clip=true, width=\linewidth]{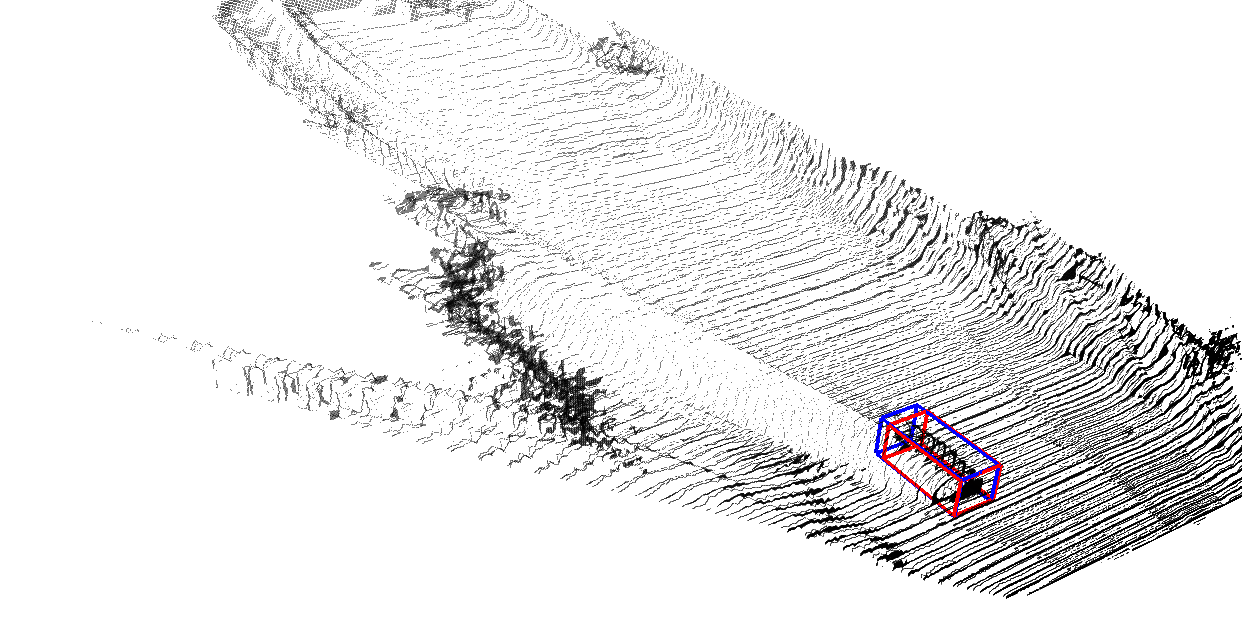}
    \vspace{-0.6cm}
    \caption{Results with BBC}
    \label{fig:after_bbc} 
\end{subfigure}
\end{center}
\vspace{-0.6cm}
\caption{\textbf{Effect of Bounding Box Consistency (BBC).} Top Row: Minimum bounding rectangle (MBR) of \textcolor{blue}{3D box estimate} (white), \textcolor{red}{instance mask} \textcolor{red}{(\textbf{red})}. Bottom left: Poor 3D box estimate without BBC. Bottom right: Improved 3D box estimate with BBC. Ground truth shown in \textcolor{red}{\textbf{red}}.
}
\label{fig:bbc}
\vspace{-0.3cm}
\end{figure}

\subsection{2D-3D Bounding Box Consistency (BBC)} \label{sec:bbc}
To alleviate the local misalignment issue, we use the geometry constraint of the bounding box consistency to refine our 3D bounding box estimate. Given an inaccurate 3D bounding box estimate, it is highly possible that its 2D projection also does not match well with the corresponding 2D proposal. An example is shown in Figure \ref{fig:before_bbc}. By adjusting the 3D bounding box estimate in 3D space so that its 2D projection can have a higher 2D Intersection of Union (IoU) with the corresponding 2D proposal, we demonstrate that the 3D IoU of 3D bounding box estimate with its ground truth can be also increased, shown in Figure \ref{fig:after_bbc}.

\vspace{1mm}
Formally, we first convert the 3D bounding box estimate ($x$, $y$, $z$, $h$, $w$, $l$, $\theta$) to the 8 corner representation $\left \{ (p_x^{n}, p_y^{n}, p_z^{n}) \right \}_{n=1}^8$. Then its 2D projection $\left \{ (u^{n}, v^{n}) \right \}_{n=1}^8$ can be computed given the camera projection matrix. From that, we can compute the minimum bounding rectangle (MBR), which is a tuple $t^e = (t^e_{x}, t^e_{y}, t^e_{w}, t^e_{h})$, representing the smallest axis-aligned 2D bounding box that can enclose the 2D point set $\left \{ (u^{n}, v^{n}) \right \}_{n=1}^8$. Similarly, we can obtain the MBR of the 2D mask proposal $t^p = (t^p_{x}, t^p_{y}, t^p_{w}, t^p_{h})$. The goal of the BBC is to increase the 2D IoU between the 2D bounding box $t^e$ and $t^p$.

\vspace{2mm}
\noindent\textbf{Bounding Box Consistency Loss (BBCL).}
During training, we propose a PointNet-based 3D box correction module\footnote{Details of the specific architectures are described in the supplementary} for bounding box refinement. 
The 3D box correction module takes the segmented point cloud and features extracted from the 3D box estimation module as the input, and outputs a correction of the 3D bounding box parameters (\emph{i.e.}, a residual).
Then our final estimate $E_f$ is the summation over the initial estimate $E_i$ and the residual. The loss can be formulated as follows:
\vspace{-0.2cm}
\begin{equation}\label{eq:bbc}
    L_{bbc} = \sum_{i\in{\left \{ x, y, w, h\right \}}} \text{smooth}_{L_1} (t^e_{i} - t^p_i)
\vspace{-0.2cm}
\end{equation}
Where $t^e$ and $t^p$ can be computed deterministically from the final estimate $E_f$ and 2D mask proposal respectively as described in Section \ref{sec:bbc}. As the gradients can be back-propagated through the entire network, we can thus train our 3D detection network with BBCL end-to-end.

\vspace{2mm}
\noindent\textbf{Bounding Box Consistency Optimization (BBCO).} 
During testing, we further refine the final estimate with the BBC constraint as a post-processing step. For each pair of the 3D bounding box estimate and its 2D proposal, we solve the same optimization problem and minimize the $L_{bbc}$ in Equation \ref{eq:bbc} using a global search optimization method.

\vspace{-0.1cm}
\section{Experiments}
\subsection{Settings}
\noindent\textbf{Dataset.}
We evaluate on the KITTI bird's eye view and 3D object detection benchmark \cite{Geiger2012}, containing $7481$ training and $7518$ testing images as well as the corresponding LiDAR point clouds, stereo images, and full camera matrix. We use the same training and validation split as \cite{Qi2018}. We emphasize again, during training and testing, our approach does not use any LiDAR point cloud or stereo image data.

\vspace{2mm}\noindent\textbf{Evaluation Metric.}
We use the evaluation toolkit provided by KITTI, which computes the precision-recall curves and average precision (AP) with the IoU thresholds at 0.5 and 0.7. We denote the AP for the bird's eye view (BEV) and 3D object detection as $AP_{BEV}$ and $AP_{3D}$ respectively.

\vspace{2mm}
\noindent\textbf{Baselines.}
We compare our method with previous state-of-the-art: Mono3D \cite{Chen2016}, Deep3DBox \cite{Mousavian2017} and MLF-MONO \cite{Xu2018}. To show the superiority of our method, we also compare with three recent concurrent works: ROI-10D \cite{Manhardt2019}, MonoGRNet \cite{Qin2018} and PL-MONO \cite{Wang2019}.

\begin{table*}[ht]
\caption{Quantitative comparison on KITTI \textbf{val} set. We report the average precision (in \%) of car category on bird's eye view and 3D object detection as $\text{AP}_{\text{BEV}}$ and $\text{AP}_{\text{3D}}$. Top three rows are previous state-of-the-art methods and middle three rows colored in \textcolor{Green}{green} are concurrent works developed independently from our work. We outperform all monocular methods.}
\vspace{-0.2cm}
\centering
\resizebox{\textwidth}{!}{
\begin{tabular}{|c|c|c|c|c|c|c|c|}
    \hline
    \multirow{2}{*}{Method} & \multirow{2}{*}{Input} & \multicolumn{3}{c|}{$\text{AP}_{\text{BEV}}$ / $\text{AP}_{\text{3D}}$ (in \%), \textbf{IoU = 0.5}} & \multicolumn{3}{c|}{$\text{AP}_{\text{BEV}}$ / $\text{AP}_{\text{3D}}$ (in \%), \textbf{IoU = 0.7}}\\
    \cline{3-8}
    & & Easy & Moderate & Hard & Easy & Moderate & Hard \\
    \hline
    Mono3D \cite{Chen2016} & Monocular & 30.5 / 25.2 & 22.4 / 18.2 & 19.2 / 15.5 & 5.2 / 2.5  & 5.2 / 2.3 & 4.1 / 2.3\\ 
    Deep3DBox \cite{Mousavian2017} & Monocular & 30.0 / 27.0 & 23.8 / 20.6 & 18.8 / 15.9 & 10.0 / 5.6 & 7.7 / 4.1 & 5.3 / 3.8\\ 
    MLF-MONO \cite{Xu2018} & Monocular & 55.0 / 47.9 & 36.7 / 29.5 & 31.3 / 26.4 & 22.0 / 10.5 & 13.6 / 5.7 & 11.6 / 5.4\\ 
    \hline
    \textcolor{Green}{ROI-10D} \cite{Manhardt2019} & Monocular & 46.9 / 37.6 & 34.1 / 25.1 & 30.5 / 21.8 & 14.5 / 9.6 & 9.9 / 6.6 & 8.7 / 6.3\\ 
    \textcolor{Green}{MonoGRNet} \cite{Qin2018} & Monocular & - / 50.5 & - / 37.0 & - / 30.8 & - / 13.9 & - / 10.2 & - / 7.6\\ 
    \textcolor{Green}{PL-MONO} \cite{Wang2019} & Monocular & 70.8 / 66.3 & 49.4 / 42.3 & 42.7 / 38.5 & 40.6 / 28.2 & 26.3 / 18.5 & 22.9 / 16.4\\ 
    \hline
    \textbf{Ours} & Monocular & \textbf{72.1} / \textbf{68.4} & \textbf{53.1} / \textbf{48.3} & \textbf{44.6} / \textbf{43.0} & \textbf{41.9} / \textbf{31.5} & \textbf{28.3} / \textbf{21.0} & \textbf{24.5} / \textbf{17.5}\\ 
    \hline
\end{tabular}}
\label{tab:quantitative}
\vspace{-0.3cm}
\end{table*}

\begin{table}
\caption{$\text{AP}_{\text{BEV}}$ / $\text{AP}_{\text{3D}}$ performance on KITTI \textbf{val} set for pedestrians and cyclists at IoU = 0.5.}
\vspace{-0.6cm}
\begin{center}
\begin{tabular}{|c|c|c|c|}
\hline
Category & Easy & Moderate & Hard\\
\hline
Pedestrian & 14.4 / 11.6 & 13.8 / 11.2 & 12.0 / 10.9\\
Cyclist & 11.0 / 8.5 & 7.7 / 6.5 & 6.8 / 6.5\\
\hline
\end{tabular}
\end{center}
\label{tab:ped_val}
\vspace{-0.8cm}
\end{table}

\begin{figure*}[ht]
\begin{center}
    \includegraphics[width=0.32\linewidth]{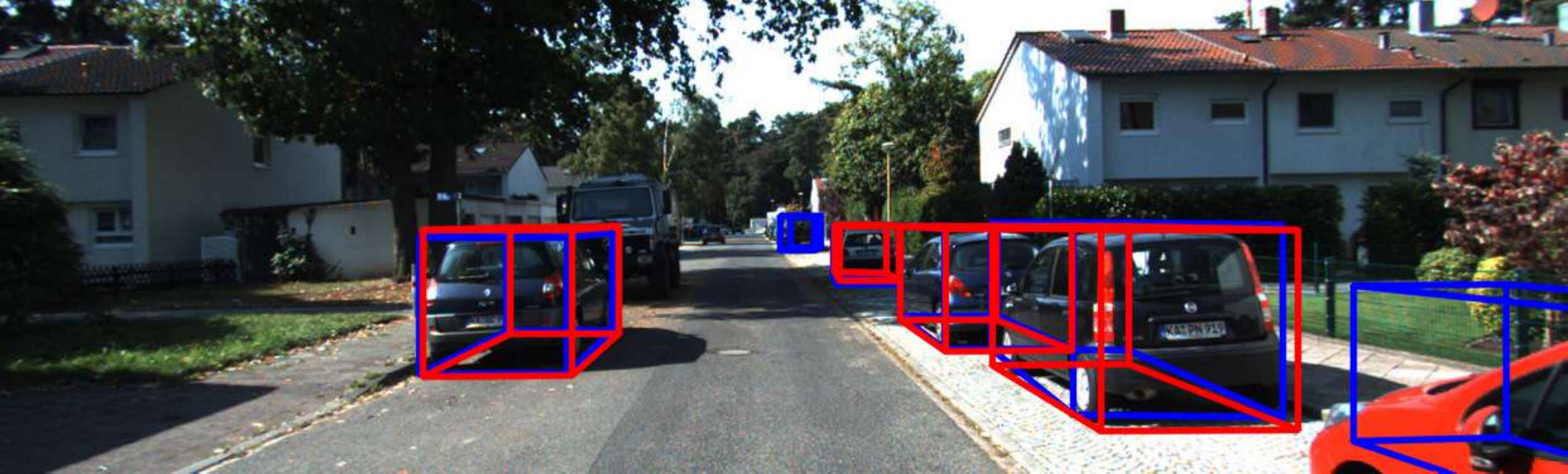}
   \includegraphics[width=0.32\linewidth]{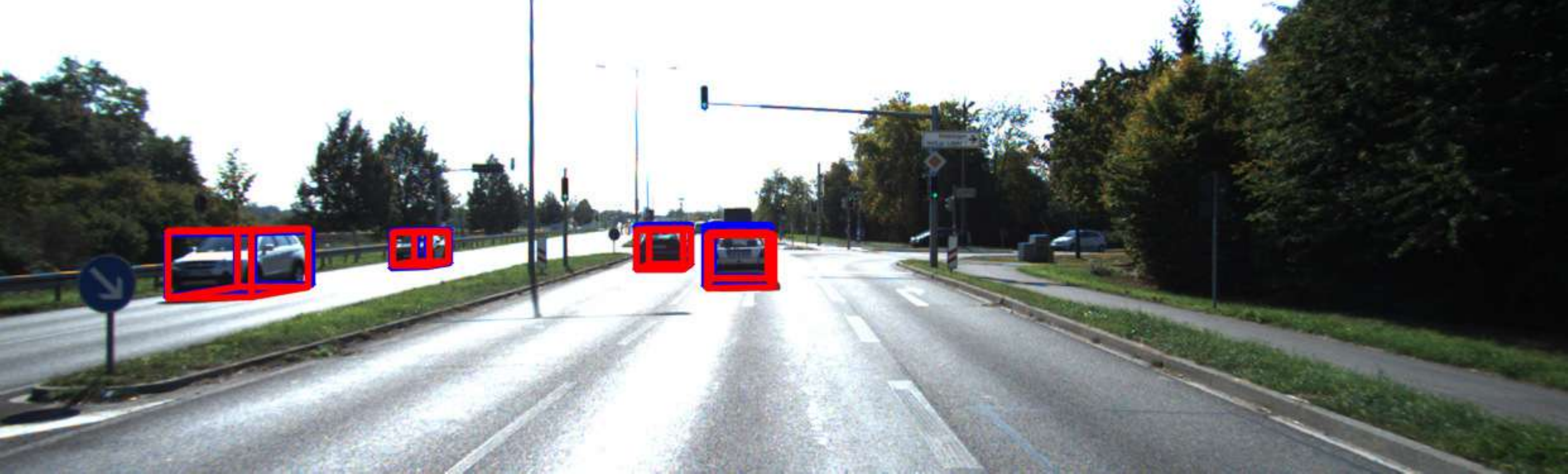}
   \includegraphics[width=0.32\linewidth]{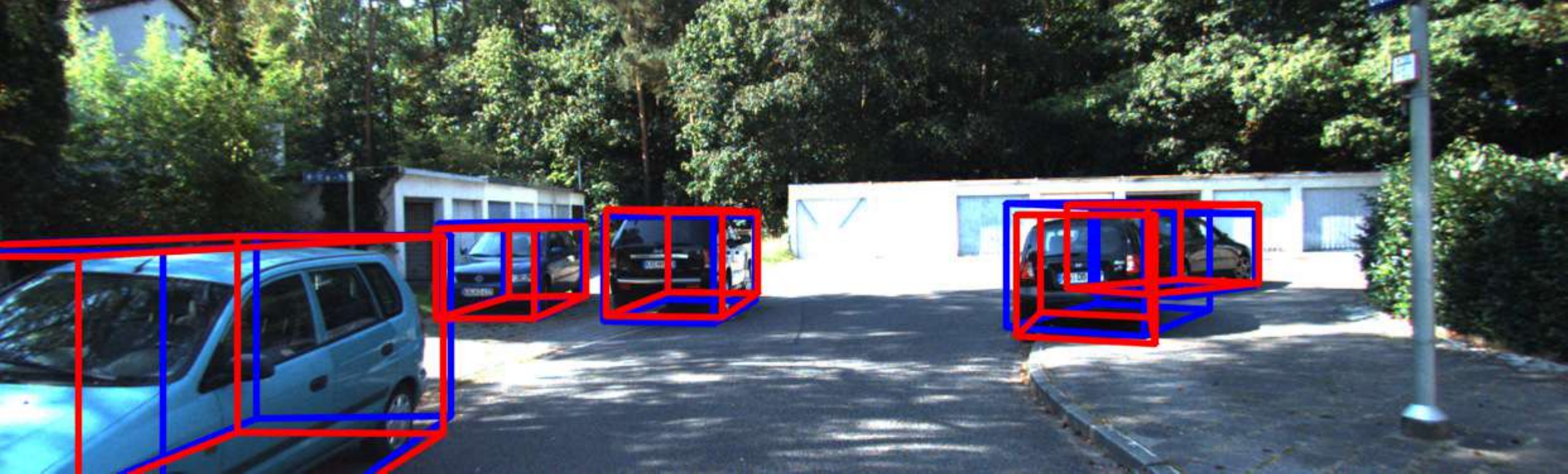}
   
   \includegraphics[trim=4cm 2cm 0cm 3cm, clip=true,
   width=0.32\linewidth]{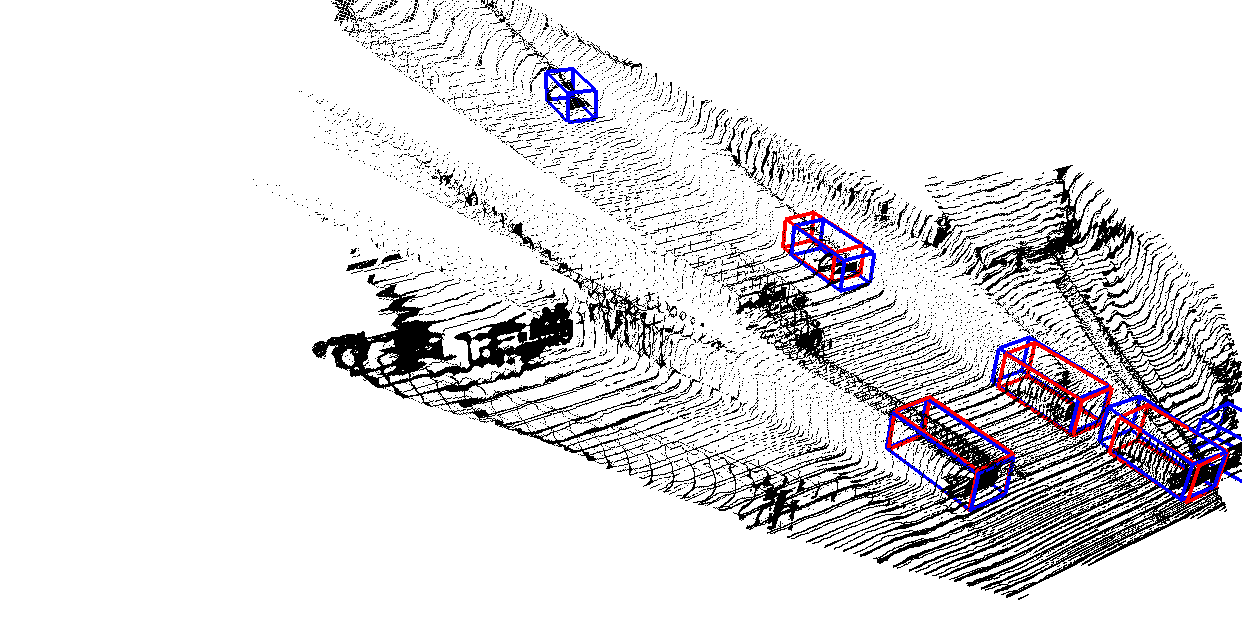}
   \includegraphics[trim=4cm 2cm 0cm 3cm, clip=true,
   width=0.32\linewidth]{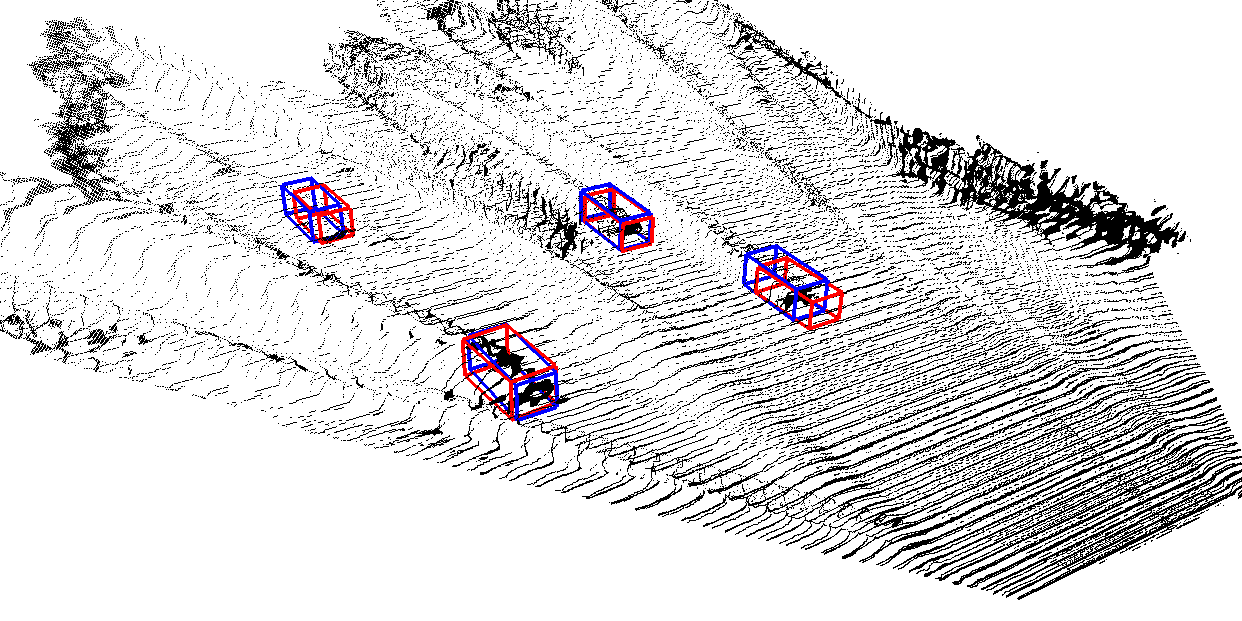}
   \includegraphics[trim=4cm 2cm 0cm 3cm, clip=true,
   width=0.32\linewidth]{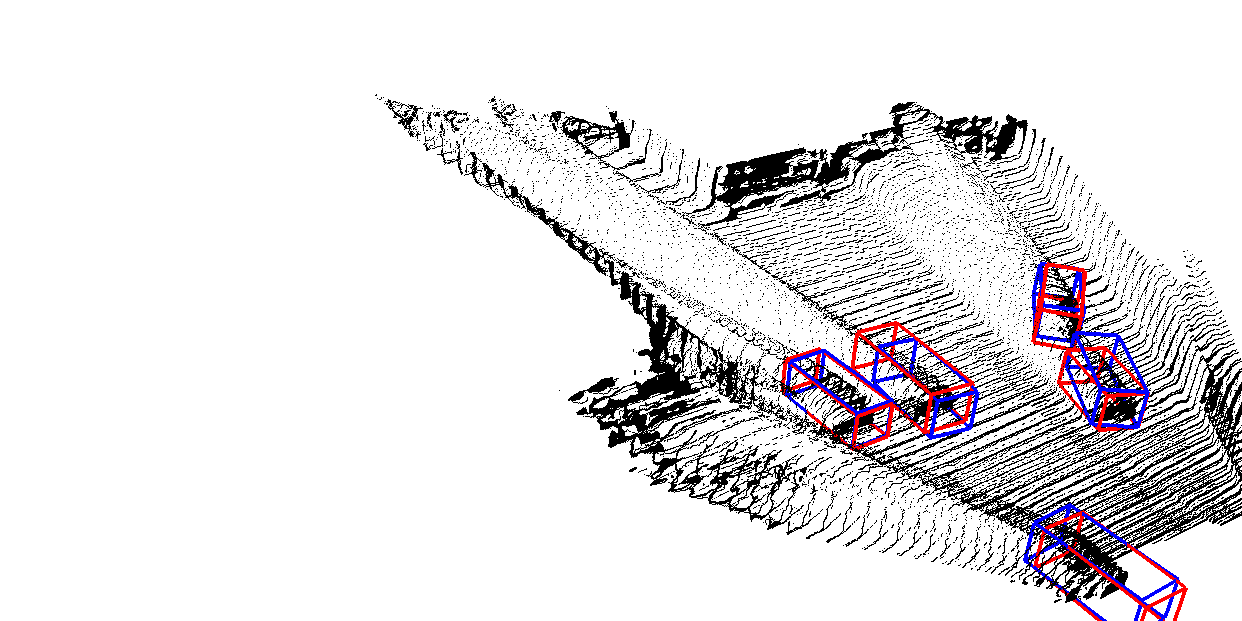}
   
   \vspace{0.1cm}
   \includegraphics[width=0.32\linewidth]{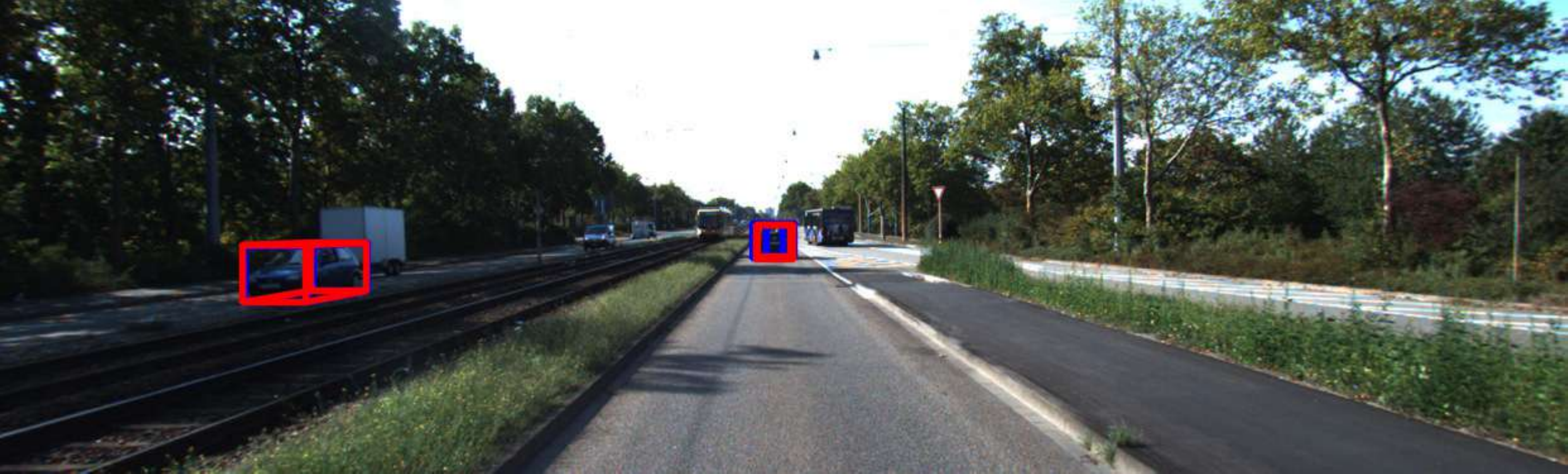}
   \includegraphics[width=0.32\linewidth]{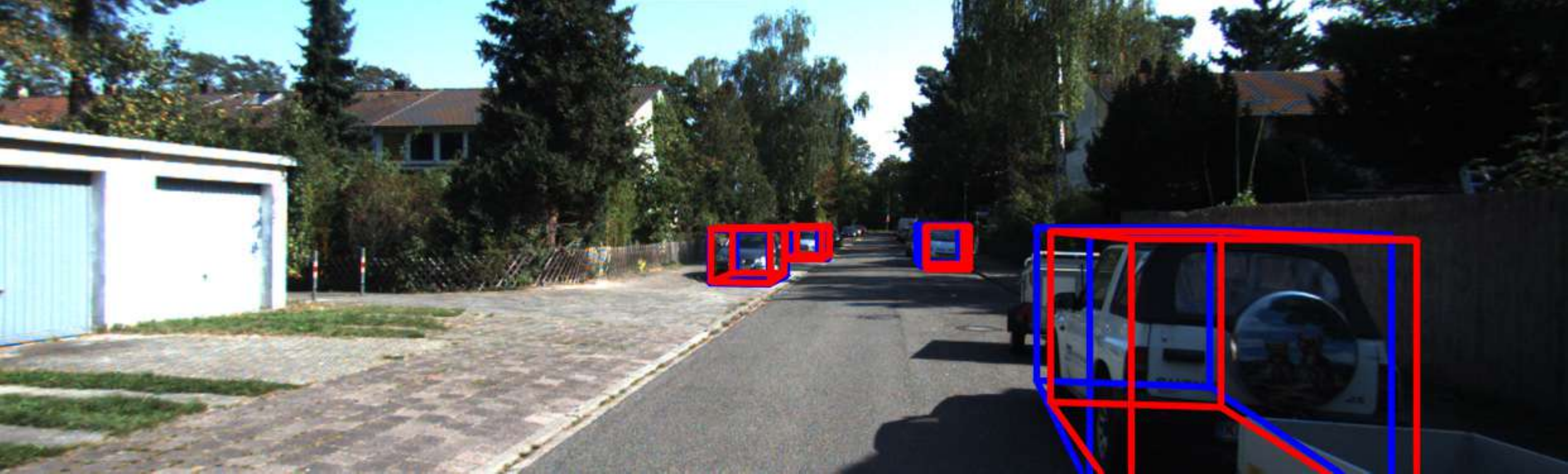}
   \includegraphics[width=0.32\linewidth]{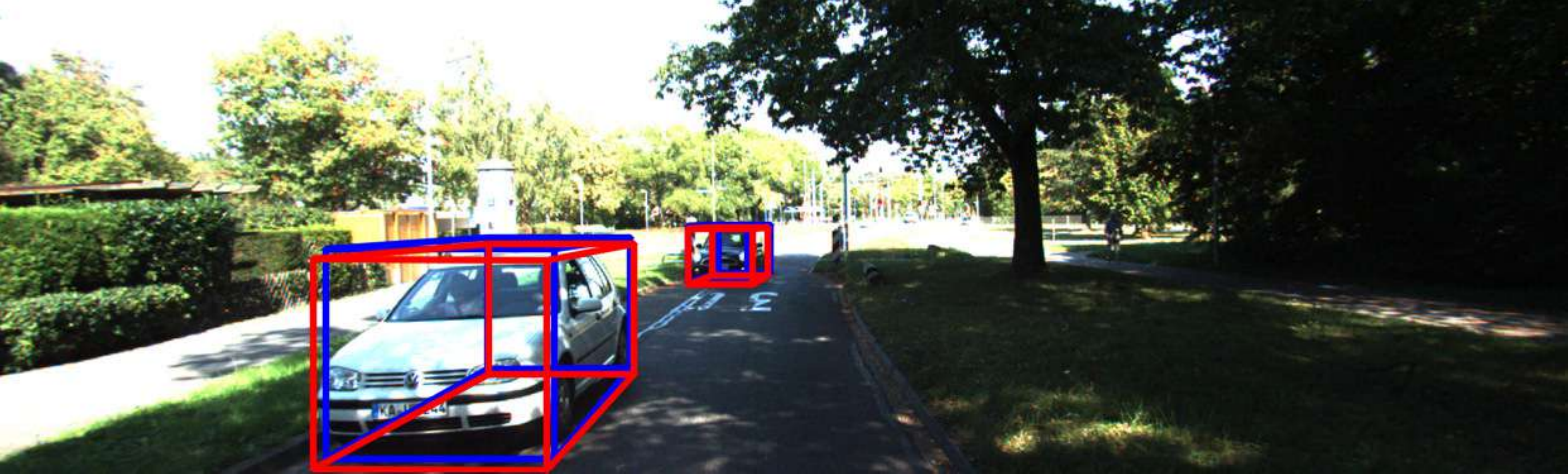}
   
   \includegraphics[trim=4cm 2cm 0cm 3cm, clip=true,
   width=0.32\linewidth]{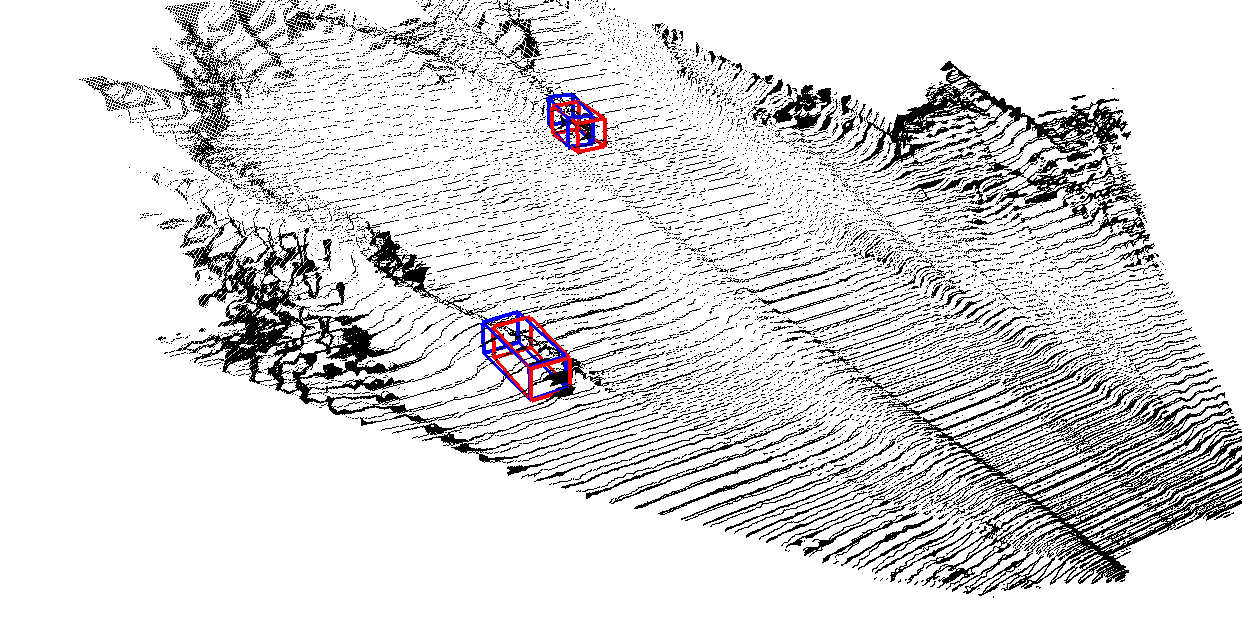}
   \includegraphics[trim=4cm 2cm 0cm 3cm, clip=true,
   width=0.32\linewidth]{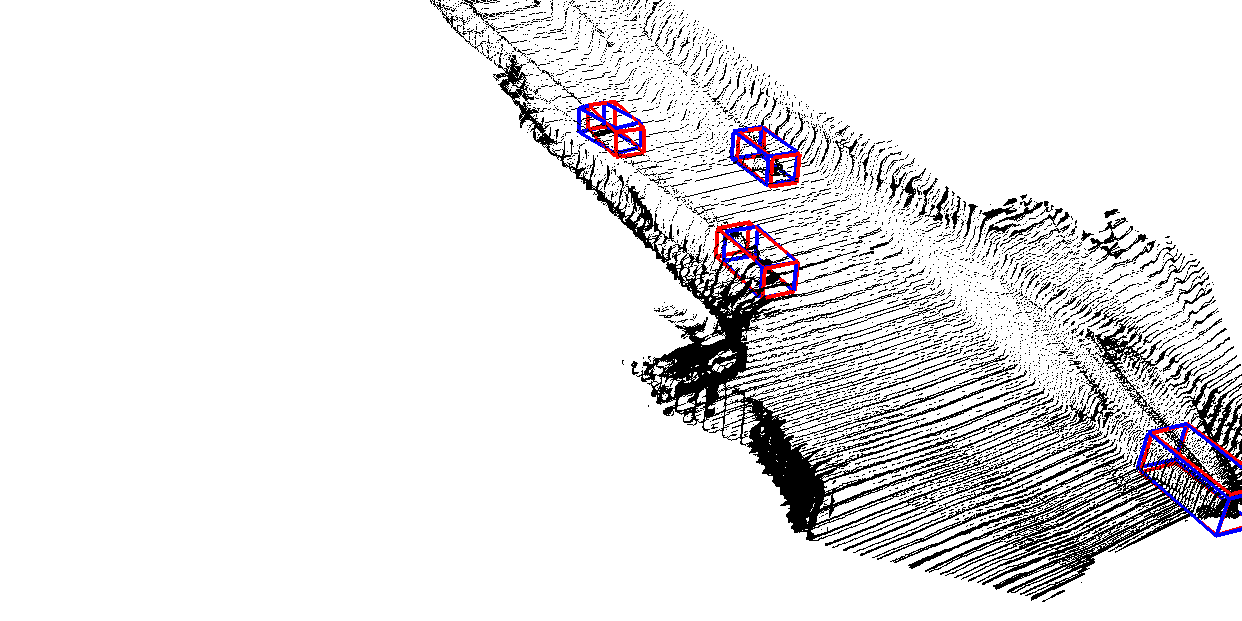}
   \includegraphics[trim=4cm 2cm 0cm 3cm, clip=true,
   width=0.32\linewidth]{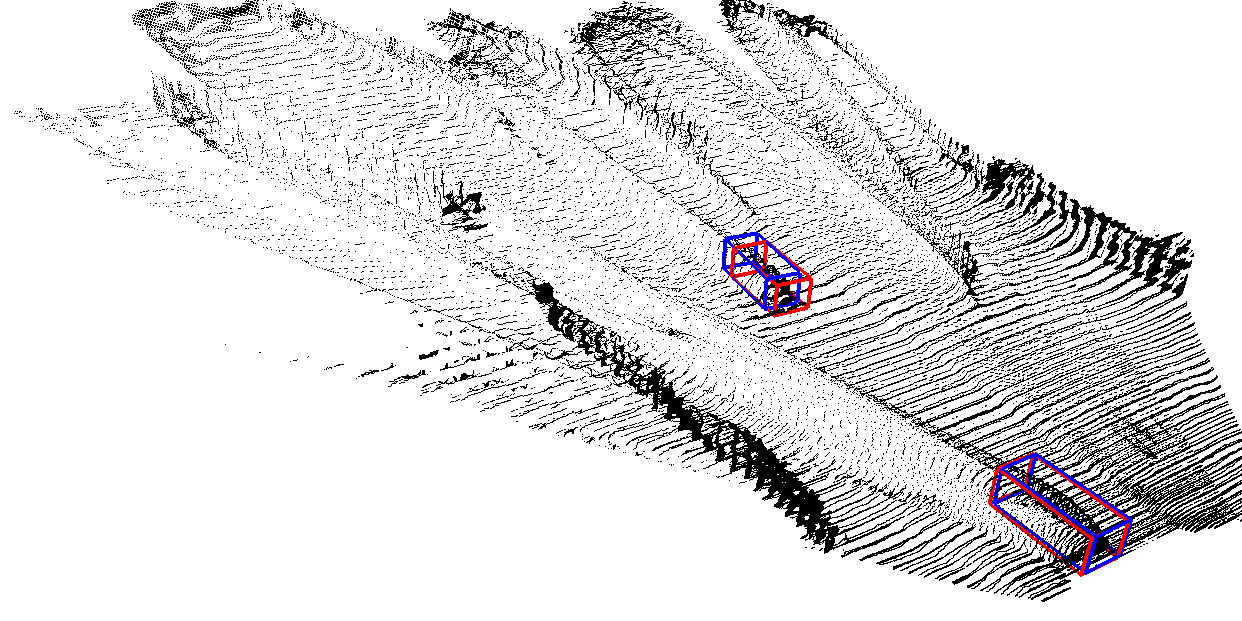}
   
\end{center}
\vspace{-0.6cm}
\caption{Qualitative results of our proposed method on KITTI \textbf{val} set. We visualize our 3D bounding box estimate (in \textcolor{blue}{blue}) and ground truth (in \textcolor{red}{red}) on the frontal images (1st and 3rd rows) and pseudo-LiDAR point cloud (2nd and 4th rows).}
\vspace{-0.3cm}
\label{fig:qua}
\end{figure*}

\begin{table*}
\caption{Summarized \textbf{positive} ablative analysis on KITTI \textbf{val} set. We show individual and combined effects of using pseudo-LiDAR (+PLiDAR), using instance mask (+Mask), training with bounding box consistency loss (+BBCL), testing with bounding box consistency optimization (+BBCO) and removing the TNet from the amodal 3D detection network (-TNet).}
\vspace{-0.2cm}
\centering
\resizebox{\textwidth}{!}{
\begin{tabular}{|c|c|c|c|c|c|c|}
    \hline
    \multirow{2}{*}{Method} & \multicolumn{3}{c|}{$\text{AP}_{\text{BEV}}$ / $\text{AP}_{\text{3D}}$ (in \%), \textbf{IoU = 0.5}} & \multicolumn{3}{c|}{$\text{AP}_{\text{BEV}}$ / $\text{AP}_{\text{3D}}$ (in \%), \textbf{IoU = 0.7}}\\
    \cline{2-7}
    & Easy & Moderate & Hard & Easy & Moderate & Hard \\
    \hline
    +PLiDAR & 71.4 / 66.2 & 49.8 / 42.5 & 42.8 / 38.6 & 40.4 / 28.9 & 26.5 / 18.2 & 22.9 / 16.2\\ 
    +PLiDAR+Mask & 70.8 / 64.7 & 51.4 / 44.5 & 44.4 / 40.4 & 41.2 / 29.4 & 27.8 / 19.8 & 24.2 / 17.5\\ 
    +PLiDAR+BBCO & 71.9 / 68.2 & 50.4 / 46.6 & 43.3 / 40.9 & \textbf{42.0} / \textbf{31.7} & 27.4 / 20.8 & 23.3 / 17.1\\ 
    +PLiDAR+BBCL & 71.7 / \textbf{68.5} & 50.3 / 46.5 & 43.2 / 40.5 & 41.6 / 31.3 & 27.0 / 20.8 & 23.1 / 17.1\\ 
    +PLiDAR-TNet & 70.4 / 66.0 & 49.8 / 42.6 & 42.7 / 38.6 & 41.7 / 29.4 & 26.4 / 18.5 & 23.0 / 16.4\\ 
    +PLiDAR+Mask+BBCO & 71.1 / 67.7 & 52.1 / 48.2 & \textbf{44.8} / 42.3 & 40.7 / 28.9 & 27.4 / 20.0 & 24.0 / 17.1\\ 
    +PLiDAR+Mask+BBCO-TNet & 71.1 / 68.1 & 52.3 / \textbf{48.3} & \textbf{44.8} / 42.2 & 41.5 / 28.5 & 28.3 / 20.3 & 24.1 / 17.2\\ 
    \textbf{Ours} (+PLiDAR+Mask+BBCO-TNet+BBCL) & \textbf{72.1} / 68.4 & \textbf{53.1} / \textbf{48.3} & 44.6 / \textbf{43.0} & 41.9 / 31.5 & \textbf{28.3} / \textbf{21.0} & \textbf{24.5} / \textbf{17.5}\\ 
    \hline
\end{tabular}}
\label{tab:ablation}
\vspace{-0.3cm}
\end{table*}

\subsection{Implementation Details}
\noindent\textbf{2D Instance Mask Proposal Detection.} As only 200 training images with pixel-wise annotation are provided by KITTI instance segmentation benchmark, it is not enough for training an instance segmentation network from scratch. Therefore, we first train our instance segmentation network\footnote{Details about the performance of our instance segmentation network are in the supplementary material.} on Cityscapes dataset \cite{Cordts2016} with 3475 training images and then fine-tune on the KITTI dataset.

\vspace{2mm}
\noindent\textbf{Amodal 3D Object Detection.}
To analyze the full potential of the Frustum PointNets \cite{Qi2018} for 3D object detection with pseudo-LiDAR, we experiment with its different variants in our ablation study: (1) Removing the intermediate supervision from the 3D segmentation loss $L_{seg3d}$ so that network can only implicitly learn to segment point cloud via minimizing the 3D bounding box loss $L_{box3d}$;
(2) Removing the TNet proposed in \cite{Qi2018} for object center regression and learning to predict the object center location using the 3D box estimation module; 
(3) Varying number of points sampled from the segmented point cloud to show the effect of point cloud density.

\vspace{2mm}
\noindent\textbf{Bounding Box Consistency Optimization (BBCO).} We use the differential evolution \cite{Storn1997} as our global search optimization method to refine our 3D bounding box estimate during testing. The final estimate from the network is used as the initialization of the optimization method. The bounds of the 3D bounding box parameters are linearly increasing based on the object's depth, \emph{i.e.}, the further the objects are, the more their 3D bounding box can be adjusted.

\subsection{Experimental Results}
\noindent\textbf{Comparison with State-of-the-Art Methods.}
We summarize the bird's eye view and 3D object detection results ($\text{AP}_{\text{BEV}}$ and $\text{AP}_{\text{3D}}$) on KITTI val set in Table \ref{tab:quantitative}. Our method consistently outperforms all monocular methods by a large margin on all levels of difficulty with different evaluation metrics. We highlight that, at IoU = 0.7 (moderate) -- the metric used to rank algorithms on the KITTI leader board -- we nearly \textbf{quadruple} the $\text{AP}_{\text{3D}}$ performance over previous state-of-the-art \cite{Xu2018} (from 5.7 by MLF-MONO \cite{Xu2018} to 21.0 by ours). We emphasize that we also achieve an improvement by up to \textbf{6.0\%} (from 42.3\% by PL-MONO \cite{Wang2019} to 48.3\% by ours) absolute $\text{AP}_{\text{3D}}$ over the best-performed concurrent work \cite{Wang2019} on the moderate set at IoU = 0.5. Examples of our 3D bounding box estimate on KITTI val set are visualized in Figure \ref{fig:qua}.


\vspace{2mm}\noindent\textbf{Results on Pedestrian and Cyclist.}
We report $\text{AP}_{\text{BEV}}$ and $\text{AP}_{\text{3D}}$ results on KITTI val set for pedestrians and cyclists at IoU = 0.5 in Table \ref{tab:ped_val}. We emphasize that the bird's eye view and 3D object detection from a single image for pedestrians and cyclists are much more challenging than cars due to the small sizes of the objects. Therefore, none\footnote{To avoid confusion, we note that \cite{Wang2019} is the first to present results on pedestrians and cyclists from stereo input instead of monocular input.} of prior monocular works has ever reported the results for pedestrians and cyclists. Although our reported $\text{AP}_{\text{BEV}}$ and $\text{AP}_{\text{3D}}$ performance for pedestrians and cyclists are significantly worse than for cars, we argue that this is a good starting point for future monocular work. 


\subsection{Ablation Study}
Unless otherwise mentioned, we conduct all the ablative analysis by progressively including modules in the network. In the most basic setting, we use only the proposed pseudo-LiDAR (+PLiDAR in Table \ref{tab:ablation}) generated from the DORN \cite{Fu2018}, without using the instance mask as the representation of the 2D proposal and bounding box consistency to refine the 3D bounding box estimate. Instead, it (\emph{i.e.}, +PLiDAR) uses 2D bounding boxes detected by the Faster R-CNN \cite{Ren2015} as the 2D proposals and follows the original Frustum PointNet \cite{Qi2018} for 3D bounding box estimation. We train the network from scratch by random initializing its weights and sample 512 points from the segmented point cloud for 3D bounding box estimation. All \textbf{positive} ablative analysis is summarized in Table \ref{tab:ablation} and negative analysis is in Table \ref{tab:3dseg} \ref{tab:density} and \ref{tab:finetune}. The best-performed model, also illustrated in Figure \ref{fig:pipeline}, is the combination of using pseudo-LiDAR, instance mask proposals, training with BBCL, testing with BBCO and removing the TNet from the Frustum PointNets.

\vspace{2mm}
\noindent\textbf{Instance Mask vs. Bounding Box Proposal.} We replace the bounding box proposals in +PLiDAR with our proposed instance mask proposals in +PLiDAR+Mask. In Table \ref{tab:ablation}, we observe that +PLiDAR+Mask consistently outperforms +PLiDAR about 1-2\% AP on all subsets except for the easy set at IoU = 0.5.

\vspace{2mm}
\noindent\textbf{Effect of Bounding Box Consistency.} In Table \ref{tab:ablation}, we compare +PLiDAR with +PLiDAR+BBCL (training the network with bounding box consistency loss) and +PLiDAR+BBCO (applying bounding box consistency optimization during testing). We show that either BBCL or BBCO improves the performance significantly, \emph{e.g.}, $\text{AP}_{\text{3D}}$ from 42.5\% to 46.6\% in the moderate set at IoU = 0.5.

\vspace{2mm}
\noindent\textbf{Removing the TNet.} We observe a mild improvement when comparing +PLiDAR-TNet with +PLiDAR at IoU = 0.7 in Table \ref{tab:ablation}. On the other hand, removing the TNet does not make any obvious difference on all sets at IoU = 0.5.

\begin{table}
\caption{Effect of 3D segmentation loss $L_{seg3d}$. $\text{AP}_{\text{BEV}}$ and $\text{AP}_{\text{3D}}$ results on KITTI \textbf{val} set for car category at IoU = 0.7.}
\vspace{-0.6cm}
\begin{center}
\begin{tabular}{|c|c|c|c|}
\hline
loss $L_{seg3d}$ & Easy & Moderate & Hard\\
\hline
w/ (+PLiDAR) & 40.4 / 28.9 & 26.5 / 18.2 & 22.9 / 16.2\\
w/o & 32.9 / 21.8 & 22.4 / 15.5 & 20.4 / 14.8\\
\hline
\end{tabular}
\end{center}
\vspace{-0.5cm}
\label{tab:3dseg}
\end{table}

\vspace{2mm}
\noindent\textbf{Effect of 3D Segmentation Loss.} In Table \ref{tab:3dseg}, we also compare +PLiDAR with the variant trained without the 3D segmentation loss $L_{seg3d}$. We observe a significant performance drop, meaning that it is difficult to learn the point cloud segmentation network without direct supervision.

\begin{table}
\caption{Effect of point cloud density. $\text{AP}_{\text{BEV}}$ and $\text{AP}_{\text{3D}}$ results on KITTI \textbf{val} set for car category at IoU = 0.7.}
\vspace{-0.6cm}
\begin{center}
\begin{tabular}{|c|c|c|c|}
\hline
Num. of Points & Easy & Moderate & Hard\\
\hline
4096 & 41.1 / 29.0 & 26.9 / 18.4 & 23.1 / 16.4\\
2048 & 41.1 / 28.9 & 26.3 / 18.2 & 22.9 / 16.2\\
1024 & 40.7 / 29.2 & 26.0 / 18.2 & 22.9 / 16.1\\
512 (+PLiDAR) & 40.4 / 28.9 & 26.5 / 18.2 & 22.9 / 16.2\\
256  & 41.8 / 29.1 & 26.5 / 18.3 & 23.0 / 16.2\\
\hline
\end{tabular}
\end{center}
\vspace{-0.7cm}
\label{tab:density}
\end{table}

\vspace{2mm}
\noindent\textbf{Effect of Point Cloud Density.} In Table \ref{tab:density}, we compare models trained with the different number of points sampled from the segmented point cloud before feeding into the 3D box estimation module. 
Surprisingly, it turns out increasing the point cloud density (\emph{e.g.}, from 512 to 4096 points) does not improve the performance.

\begin{table}
\caption{Fine-tuning vs. training from scratch. $\text{AP}_{\text{BEV}}$ and $\text{AP}_{\text{3D}}$ results on KITTI \textbf{val} set for car category at IoU = 0.7.}
\vspace{-0.6cm}
\begin{center}
\resizebox{0.48\textwidth}{!}{
\begin{tabular}{|c|c|c|c|}
\hline
Initialization & Easy & Moderate & Hard\\
\hline
random (+PLiDAR) & 40.4 / 28.9 & 26.5 / 18.2 & 22.9 / 16.2\\
pre-trained & 40.6 / 27.1 & 26.1 / 18.1 & 22.6 / 16.0\\
\hline
\end{tabular}}
\end{center}
\vspace{-0.7cm}
\label{tab:finetune}
\end{table}

\vspace{2mm}
\noindent\textbf{Fine-Tuning vs. Training from Scratch.} In Table \ref{tab:finetune}, we compare +PLiDAR (\emph{i.e.}, training with randomly initialized weights) with its variant, which initializes the weights from the pre-trained model of Frustum PointNets. Surprisingly, training with the pre-trained weights slightly drops the performance. We argue that it is because the pre-trained model provided by Frustum PointNets might have over-fitted on the LiDAR point cloud data and cannot be easily adapted to consume our pseudo-LiDAR input.



\vspace{-0.1cm}
\section{Conclusion}
In this paper, we propose a novel monocular 3D object detection pipeline that can enhance LiDAR-based algorithms to work with single image input, without the need of 3D sensors (\emph{e.g.}, the stereo camera, the depth camera or the LiDAR). The essential step of the proposed pipeline is to lift the 2D input image to a 3D point cloud, which we call \emph{pseudo-LiDAR} point cloud. To handle the \emph{local misalignment} and \emph{long tail} issues caused by the noise in the pseudo-LiDAR, we propose to (1) use a 2D-3D bounding box consistency constraint to refine our 3D box estimate; (2) use the instance mask proposal to generate the point cloud frustum. Importantly, our method achieves the top-ranked performance on KITTI bird's eye view and 3D object detection benchmark among all monocular methods, quadrupling the performance over previous state-of-the-art. Although our focus is monocular 3D object detection, our method can be easily extended to work with stereo image input.


{\small
\bibliographystyle{ieee}
\bibliography{main}
}

\end{document}


\title{Monocular 3D Object Detection with Pseudo-LiDAR Point Cloud \\ Supplementary Material}

\author{Xinshuo Weng\\
Carnegie Mellon University\\
{\tt\small xinshuow@cs.cmu.edu}
\and
Kris Kitani\\
Carnegie Mellon University\\
{\tt\small kkitani@cs.cmu.edu}
}

\maketitle

\section{Overview}
This document provides additional technical details, extra experiments, more visualization and justification of our idea. Each section in this document corresponds to the subsection of the approach section in the main paper.

\vspace{-0.1cm}
\section{Pseudo-LiDAR Generation}

\vspace{-0.1cm}\noindent\textbf{Additional Visualization of LiDAR vs. Pseudo-LiDAR}

\begin{figure*}
\begin{center}

\includegraphics[trim=0cm 0cm 0cm 0cm, clip=true, width=\linewidth]{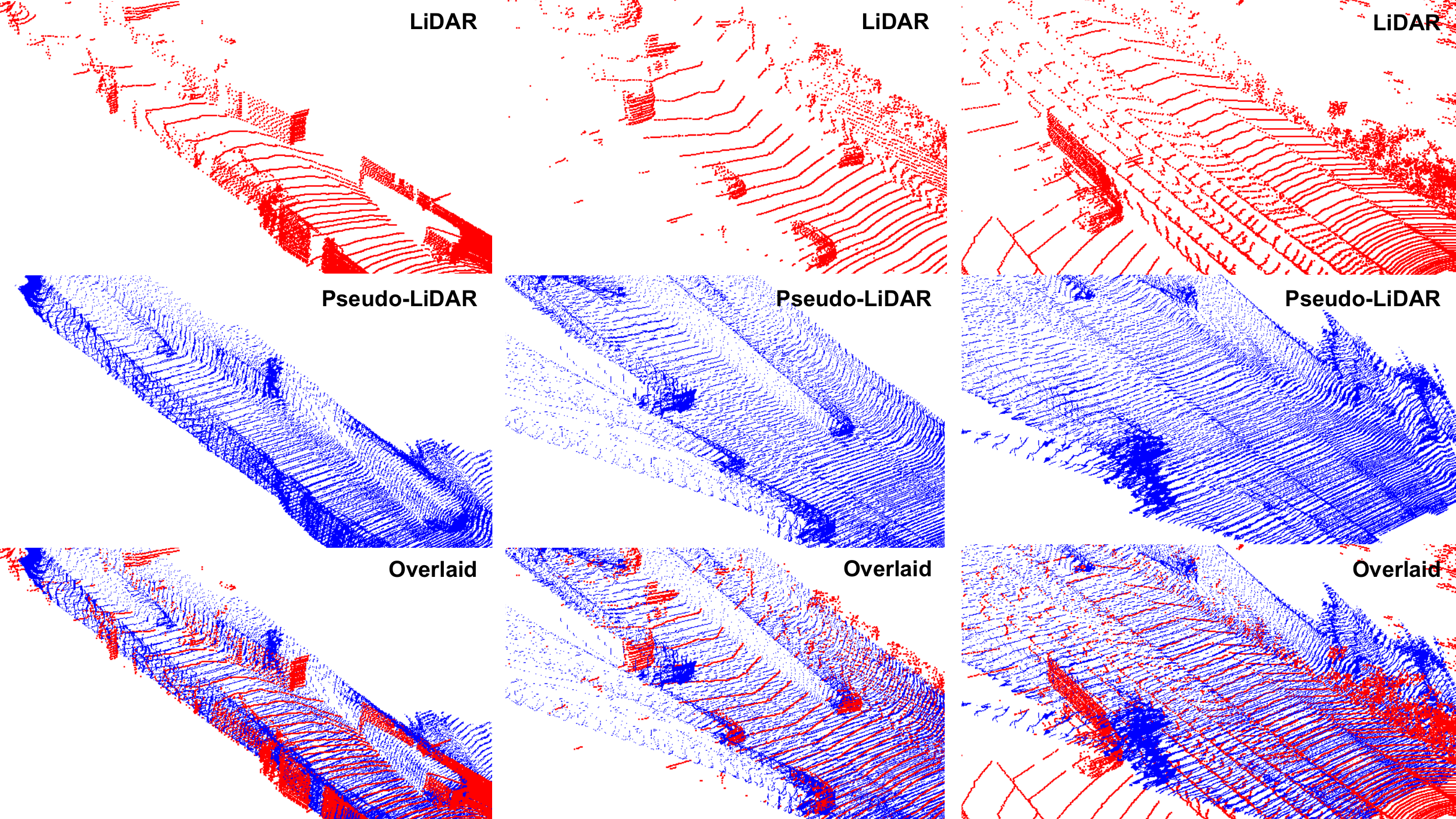}
\end{center}
\vspace{-0.5cm}
\caption{Additional visual comparison between the \textcolor{red}{\textbf{LiDAR}} (top), \textcolor{blue}{\textbf{pseudo-LiDAR}} (middle) and an overlaid version (bottom). 
}
\label{fig:lidarvspseudo}
\vspace{-0.4cm}
\end{figure*}

We provide the additional visual comparison between the LiDAR and pseudo-LiDAR point cloud in Figure \ref{fig:lidarvspseudo}, demonstrating again the \emph{local misalignment} and \emph{long tail} issues we have observed in the pseudo-LiDAR point cloud.

\vspace{-0.1cm}
\section{2D Instance Mask Proposal Detection}


\vspace{-0.1cm}\noindent\textbf{Justification of Using Instance Mask Proposal for 3D Point Cloud Segmentation and 3D Box Estimation}

In the main paper, we justify the effectiveness of using instance mask proposal to generate the point cloud frustum with no tail. We provide further details here about how the generated point cloud frustum with no tail can improve the results in the subsequent 3D point cloud segmentation and 3D bounding box estimation module.

\begin{figure}[t]
\captionsetup[subfigure]{labelformat=empty}
\begin{center}
\begin{subfigure}{0.235\textwidth}
    \caption{Bounding Box Proposal}
    \vspace{-0.2cm}
\end{subfigure}
\begin{subfigure}{0.235\textwidth}
    \caption{Instance Mask Proposal}
    \vspace{-0.2cm}
\end{subfigure}
\begin{subfigure}{0.48\textwidth}
\centering
\includegraphics[trim=12cm 13cm 21cm 3cm, clip=true, width=0.49\linewidth]{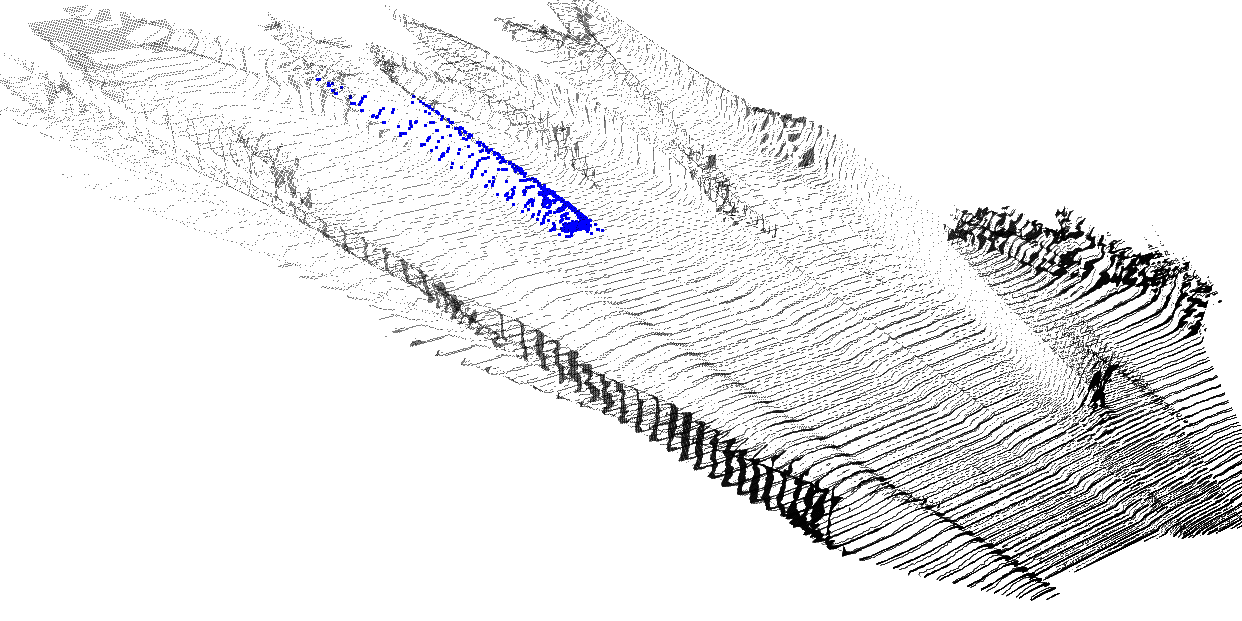}
\includegraphics[trim=12cm 13cm 21cm 3cm, clip=true, width=0.49\linewidth]{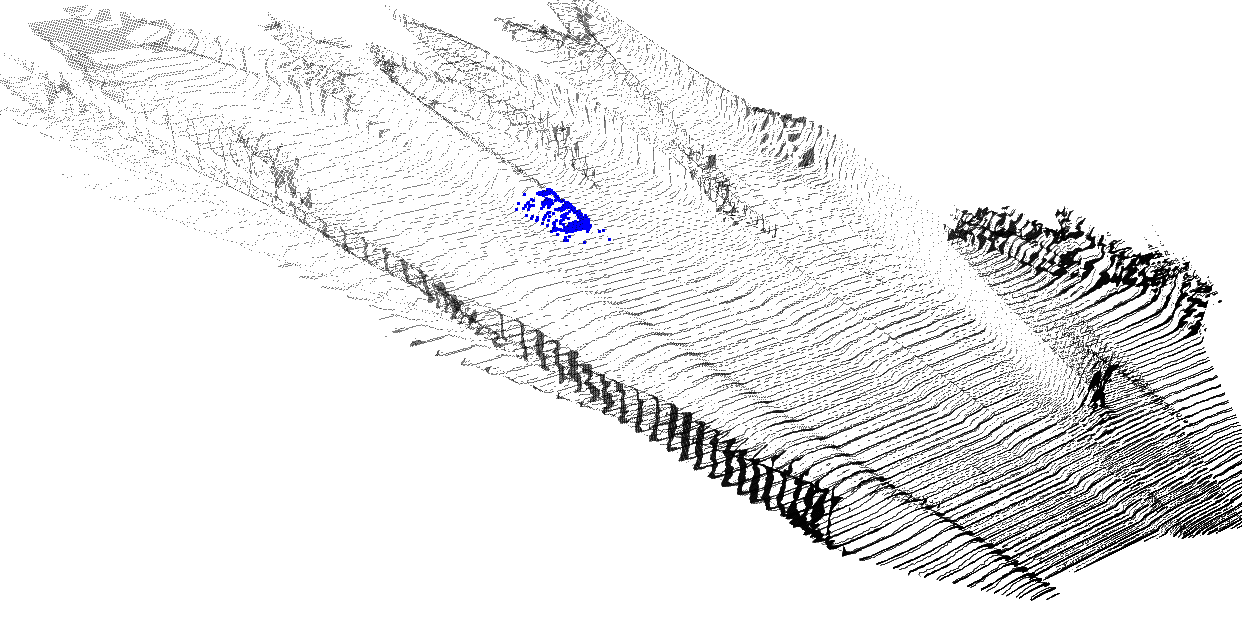}
    \vspace{-0.15cm}
    \caption{(a) Generated point cloud frustum (\textbf{\textcolor{blue}{blue}})}
\end{subfigure}
\begin{subfigure}{0.48\textwidth}
\centering
\includegraphics[trim=12cm 13cm 21cm 3cm, clip=true, width=0.49\linewidth]{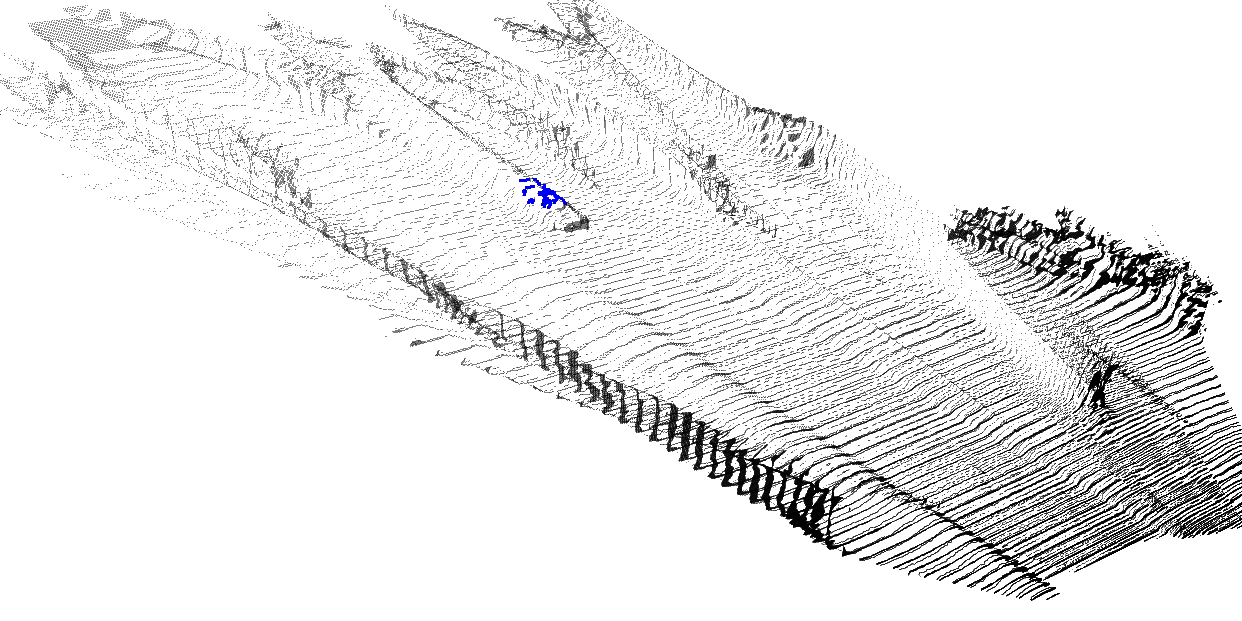}
\includegraphics[trim=12cm 13cm 21cm 3cm, clip=true,
width=0.49\linewidth]{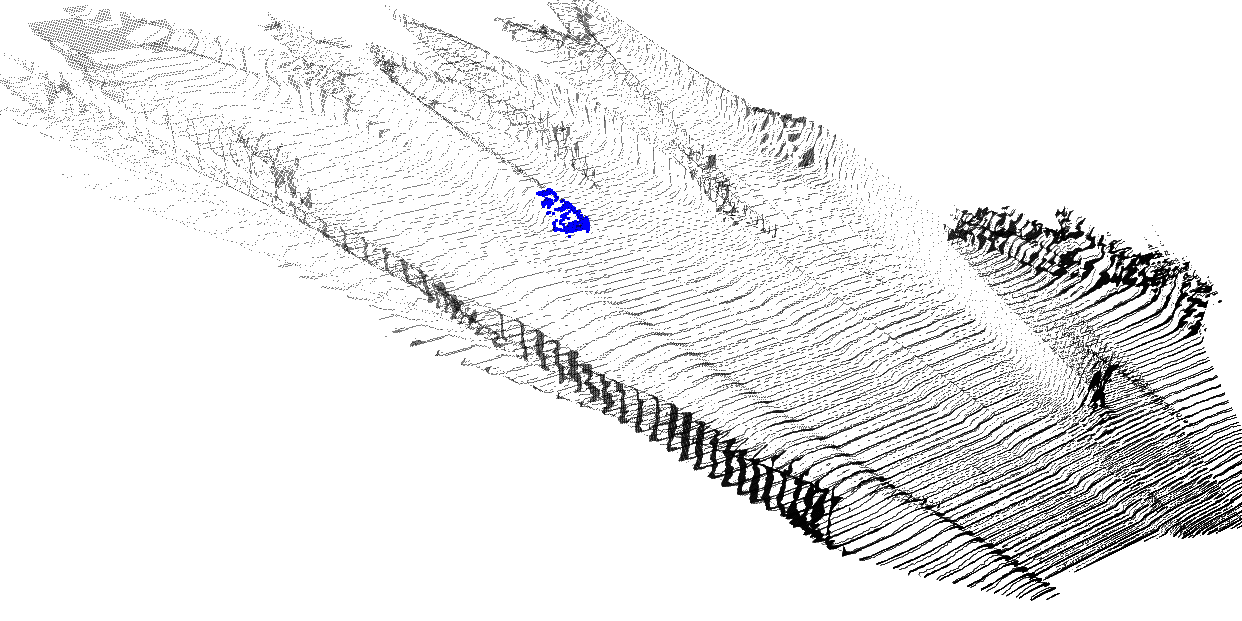}
    \vspace{-0.15cm}
    \caption{(b) Segmented point cloud (\textbf{\textcolor{blue}{blue}})}
\end{subfigure}
\begin{subfigure}{0.48\textwidth}
    \centering
    \includegraphics[trim=12cm 13cm 21cm 3cm, clip=true, width=0.49\linewidth]{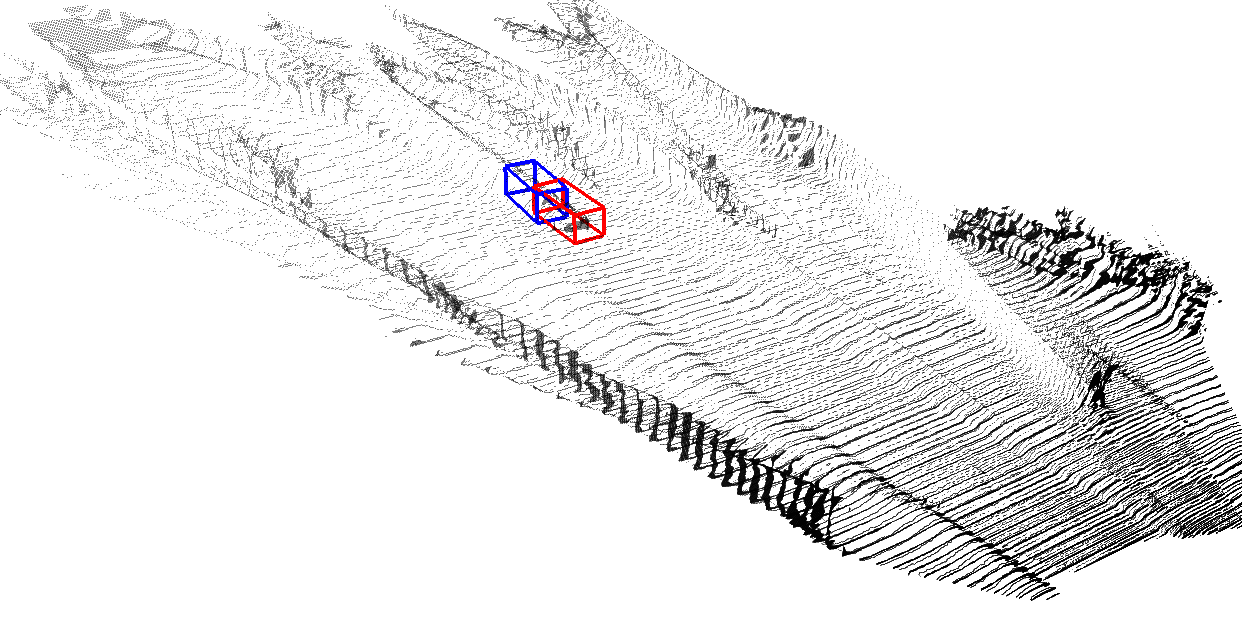}
    \includegraphics[trim=12cm 13cm 21cm 3cm, clip=true,
    width=0.49\linewidth]{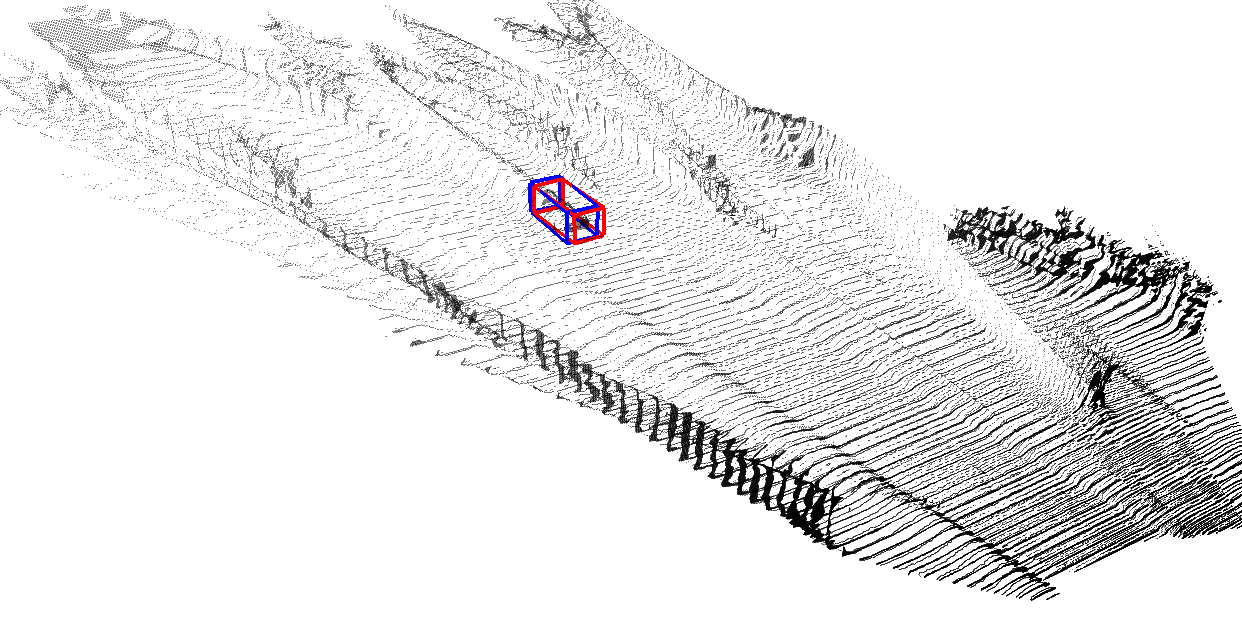}
    \vspace{-0.15cm}
    \caption{(c) 3D bounding box estimation (\textbf{\textcolor{blue}{blue}}) and ground truth (\textbf{\textcolor{red}{red}})}
\end{subfigure}

\end{center}
\vspace{-0.6cm}
\caption{\textbf{Justification of Using Instance Mask Proposal.} We visualize the generated point cloud frustum (top row), segmented point cloud (middle row) and the 3D box prediction (bottom row) from the bounding box and instance mask proposal respectively. We show that the frustum from the instance mask with no tail makes the 3D point cloud segmentation easier and results in a better 3D box estimate.
}
\label{fig:frustum}
\vspace{-0.2cm}
\end{figure}

An example of visualization is shown in Figure \ref{fig:frustum}. In the left column, the point cloud frustum is generated from the bounding box proposal and has a long tail, making the 3D point cloud segmentation task difficult (\emph{e.g.}, in the middle left of the figure, the segmented point cloud misses lots of points belonging to the object and still contains background points). This further causes a poor 3D box estimation, especially a poor object center estimate. On the other hand, the point cloud frustum generated from the instance mask proposal with no tail, shown in the right column, can reduce a large number of background points so that the subsequent point cloud segmentation and 3D box estimation can be more accurate.


\begin{figure*}
\begin{center}

\includegraphics[trim=0cm 3cm 0cm 0cm, clip=true, width=0.8\linewidth]{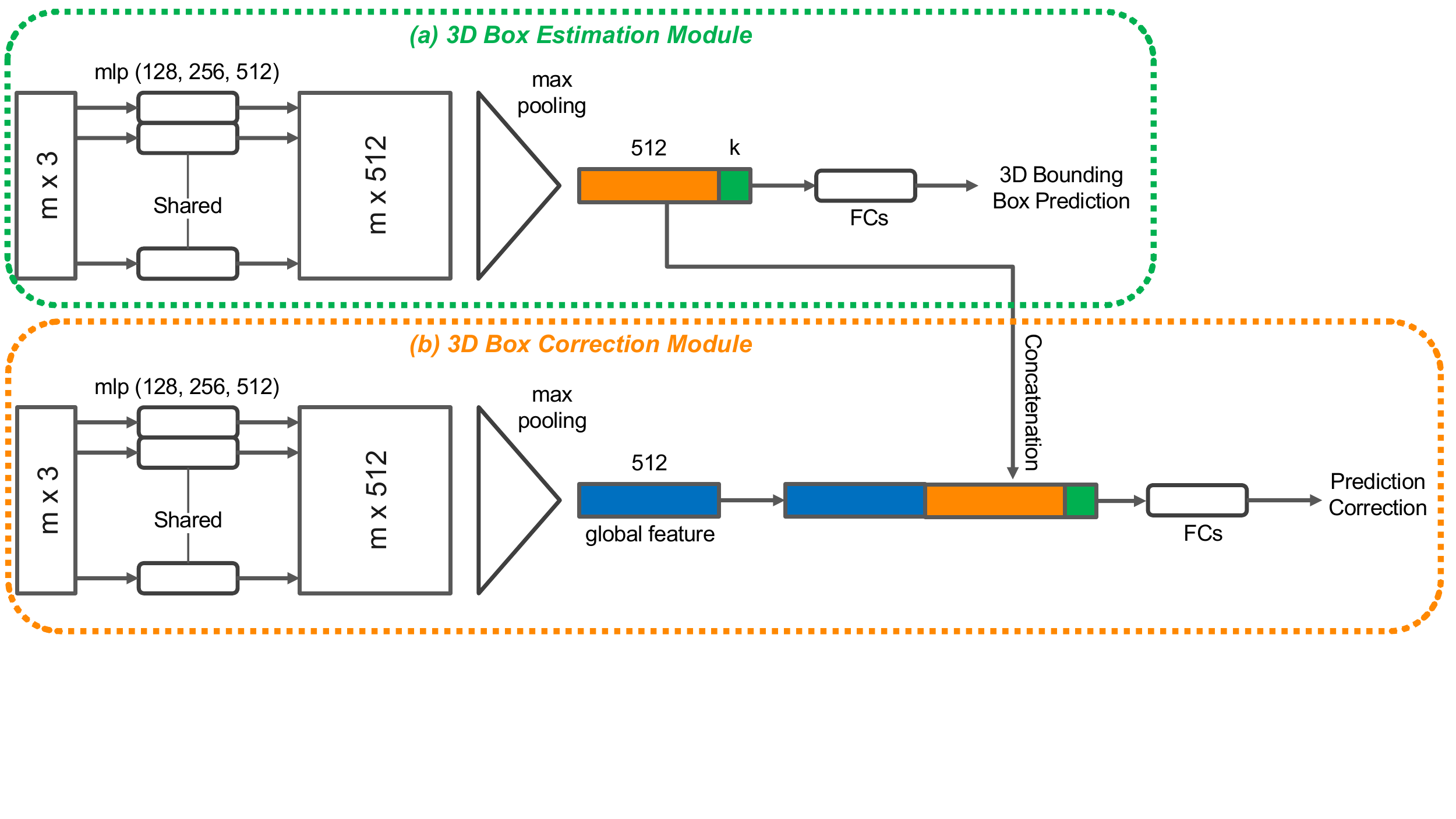}
\end{center}
\vspace{-0.7cm}
\caption{\textbf{Network Architecture of the 3D Box Estimation and 3D Box Correction Module.} Both modules take the point cloud as the input. The 3D box estimation module outputs the full 3D box parameters and the 3D box correction module outputs the residual of the parameters. $k$ is the number of class (\emph{e.g.}, 3 in KITTI). The length-$k$ vector (in \textbf{\textcolor{Green}{green}}) is a one-hot vector denotes which class the input point cloud belongs to. $\text{mlp}$ denotes the multi-layer perceptron.}
\label{fig:architecture}
\vspace{-0.3cm}
\end{figure*}

\vspace{2mm}
\noindent\textbf{Quantitative Comparison of the 2D Instance Mask and Bounding Box Proposal}


To compare our instance mask proposals with the bounding box proposals used in the baseline, we compute the minimum bounding rectangle (MBR) of our 2D mask proposals. We report the average precision (in \%) of car category on val set of KITTI \cite{Geiger2012} 2D object detection benchmark as $\text{AP}_{\text{2D}}$ in Table \ref{tab:2d_performance}. IoU thresholds of 0.5 and 0.7 are used.

\begin{table}[t]
\caption{2D proposal evaluation. $\text{AP}_{\text{2D}}$ performance on KITTI \textbf{val} set for car category at IoU = 0.5 / 0.7.}
\vspace{-0.6cm}
\begin{center}
\begin{tabular}{|c|c|c|c|}
\hline
\multirow{2}{*}{Proposal Type} & \multicolumn{3}{c|}{$\text{AP}_{\text{2D}}$ (in \%), \textbf{IoU = 0.5 / 0.7}}\\
    \cline{2-4}

& Easy & Moderate & Hard\\
\hline
Bounding Box & 97.2 / 96.5 & 97.3 / 90.3 & 90.0 / 87.6\\
Instance Mask & 96.0 / 87.6 & 89.6 / 75.7 & 80.3 / 59.4\\
\hline
\end{tabular}
\end{center}
\vspace{-0.8cm}
\label{tab:2d_performance}
\end{table}

Unsurprisingly, we find that the MBR of our mask proposal performs worse than the 2D bounding box proposal due to the lack of the pixel-level instance segmentation annotation on KITTI (only 200 images annotated with instance masks compared to 7500 images annotated with bounding boxes). However, the performance of the bird eye view and 3D object detection when using these 2D mask proposals is surprisingly higher than when using 2D bounding box proposals, which is shown in Ours (baseline) and Ours+Mask of Table 3 in the main paper. This further strengthens the effectiveness of using the instance mask proposal for 3D box estimation, \emph{i.e.} detecting the 3D bounding box from the frustum with no tail is much easier.

\begin{figure*}[ht]
\begin{center}
\includegraphics[width=0.32\linewidth]{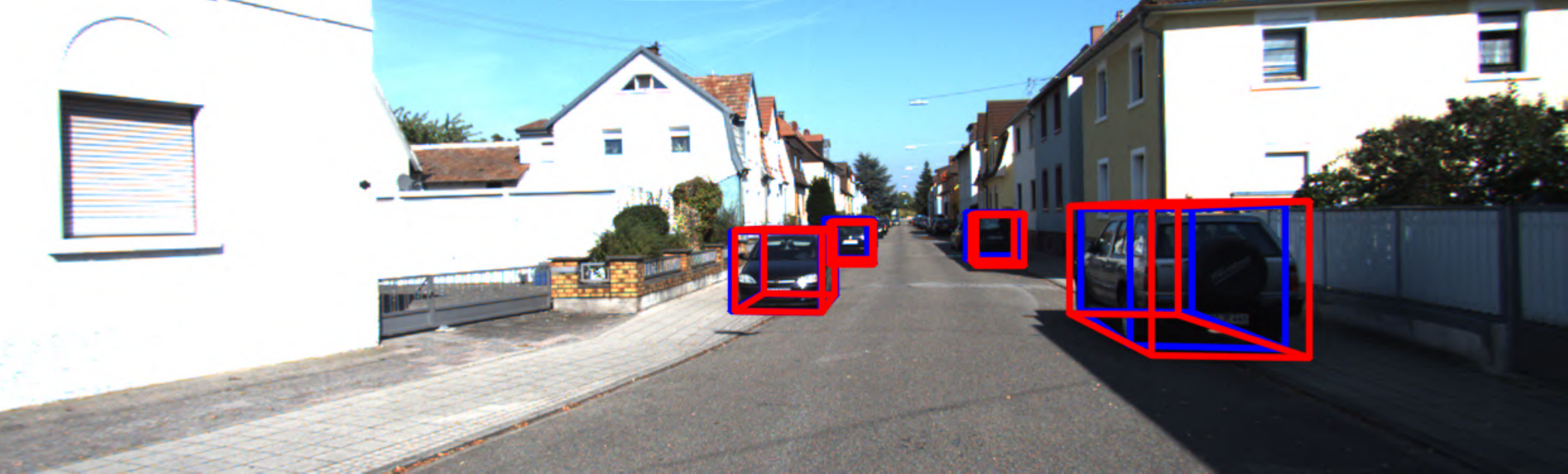}
\includegraphics[width=0.32\linewidth]{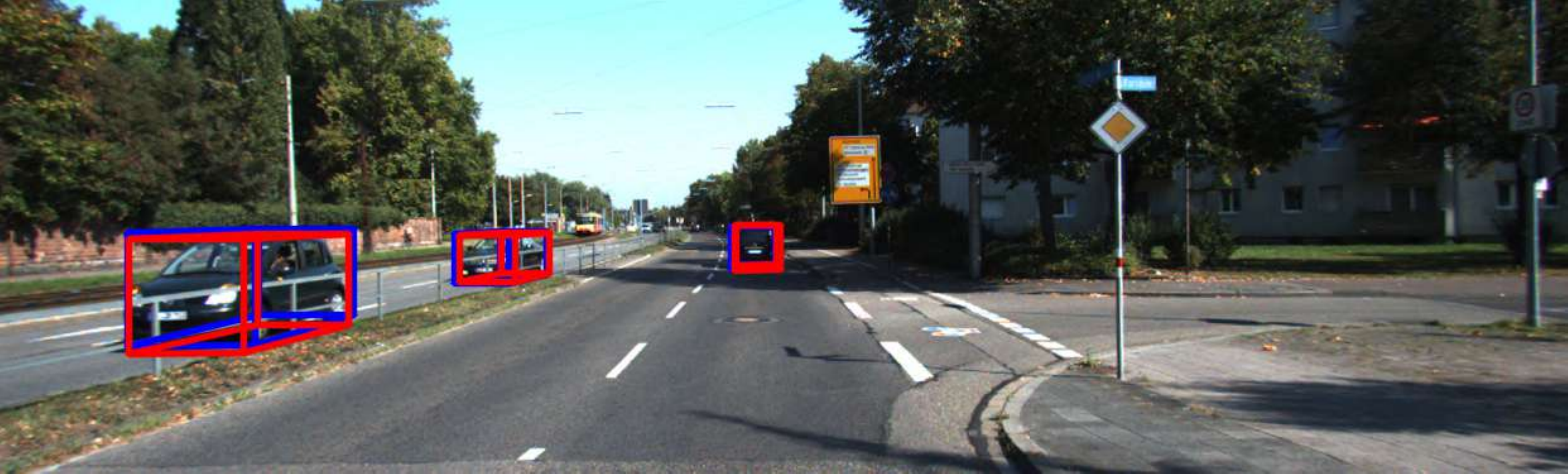}
\includegraphics[width=0.32\linewidth]{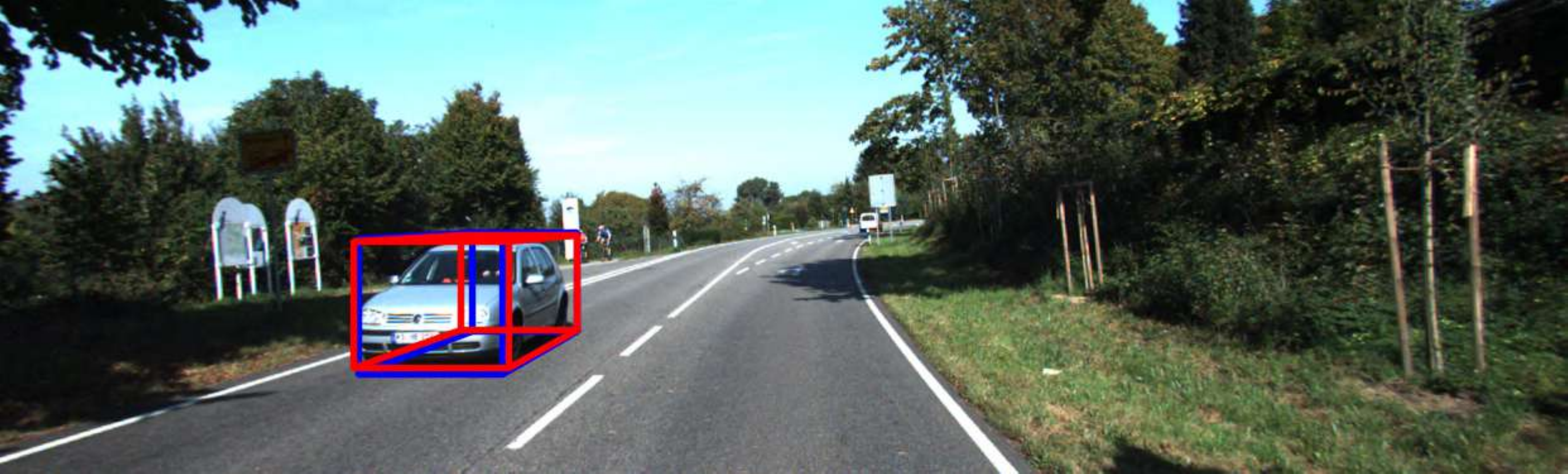}
\includegraphics[trim=4cm 2cm 0cm 3cm, clip=true, width=0.32\linewidth]{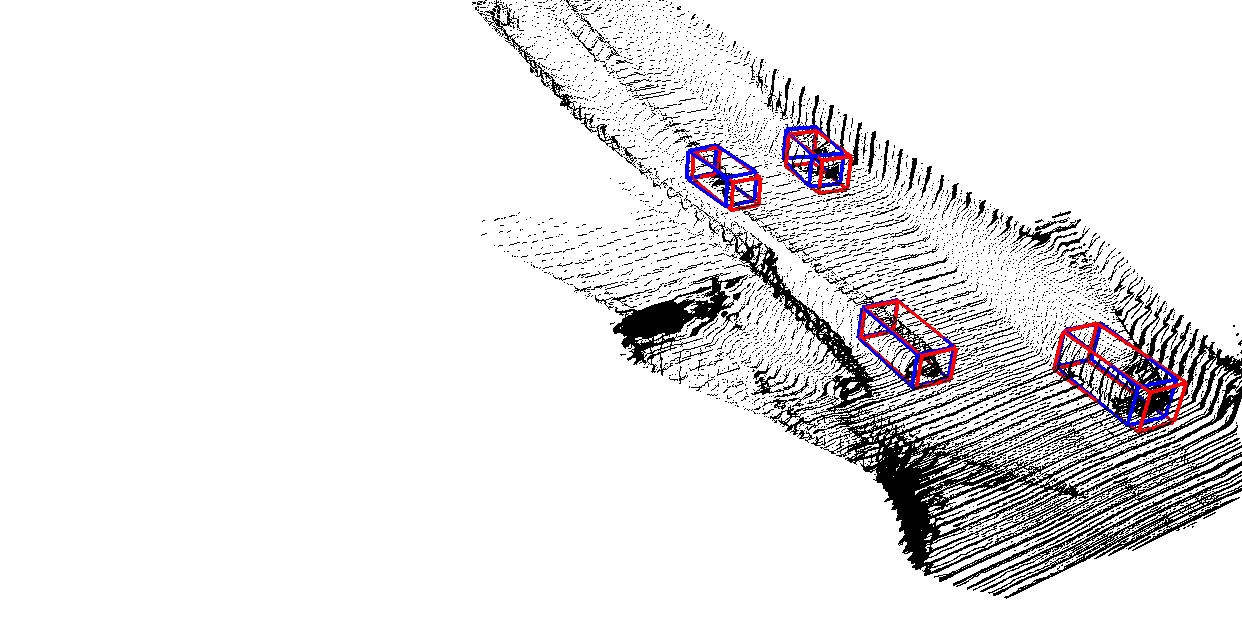}
\includegraphics[trim=4cm 2cm 0cm 3cm, clip=true, width=0.32\linewidth]{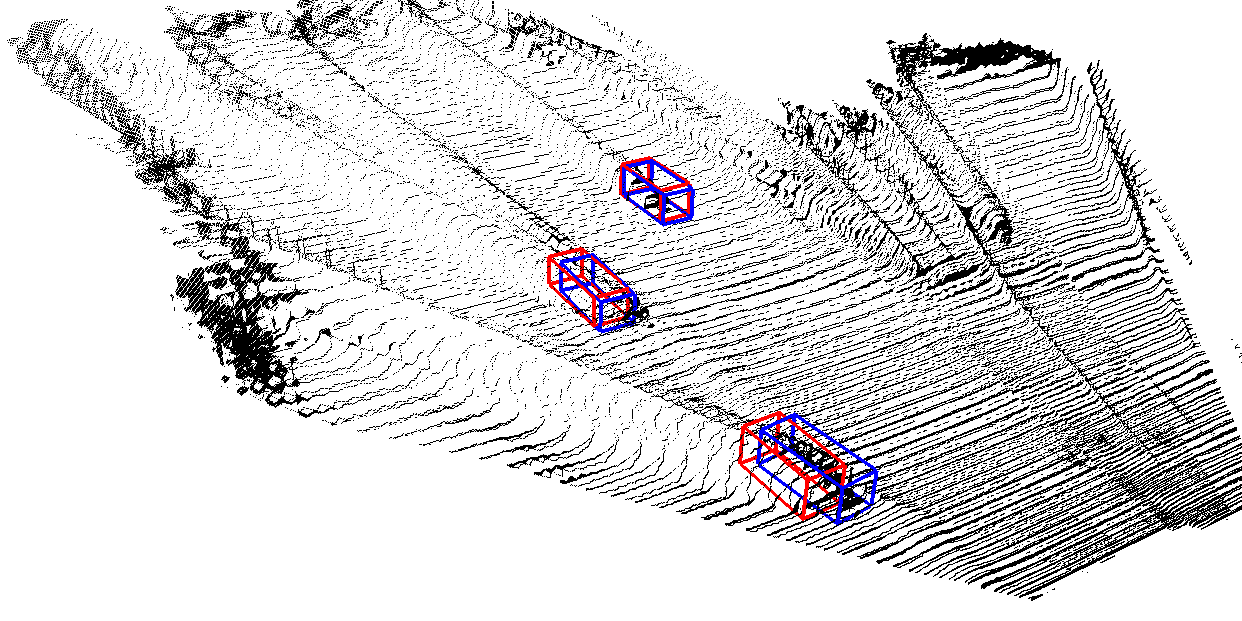}
\includegraphics[trim=4cm 2cm 0cm 3cm, clip=true, width=0.32\linewidth]{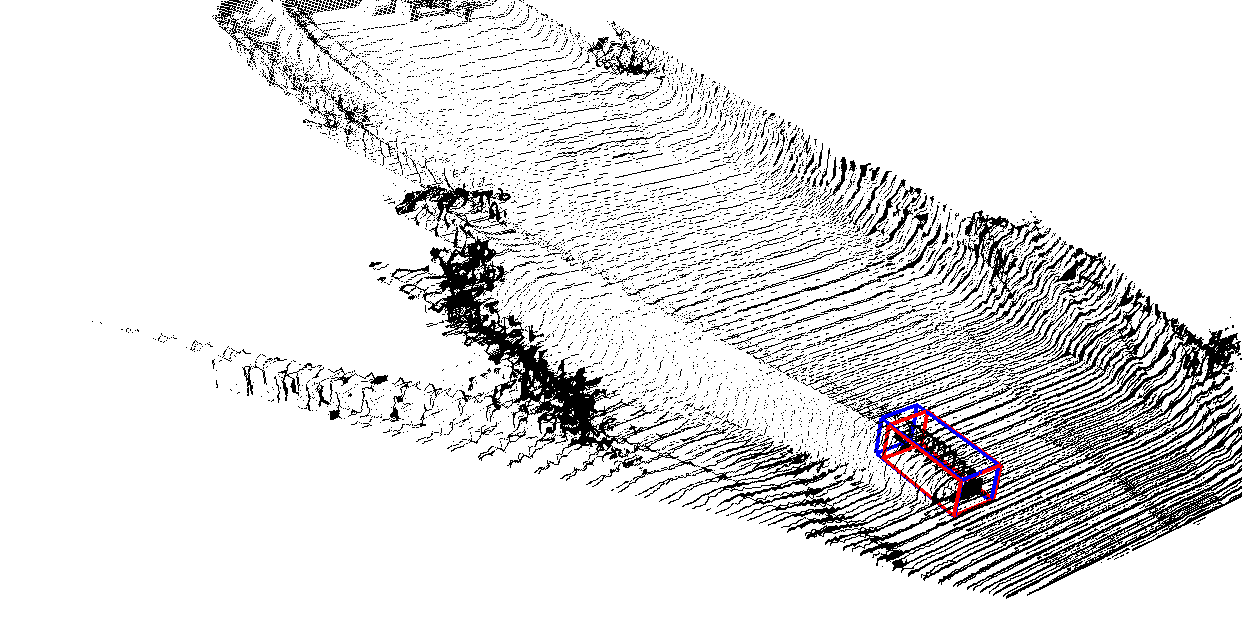}
   
\vspace{0.15cm}
\includegraphics[width=0.32\linewidth]{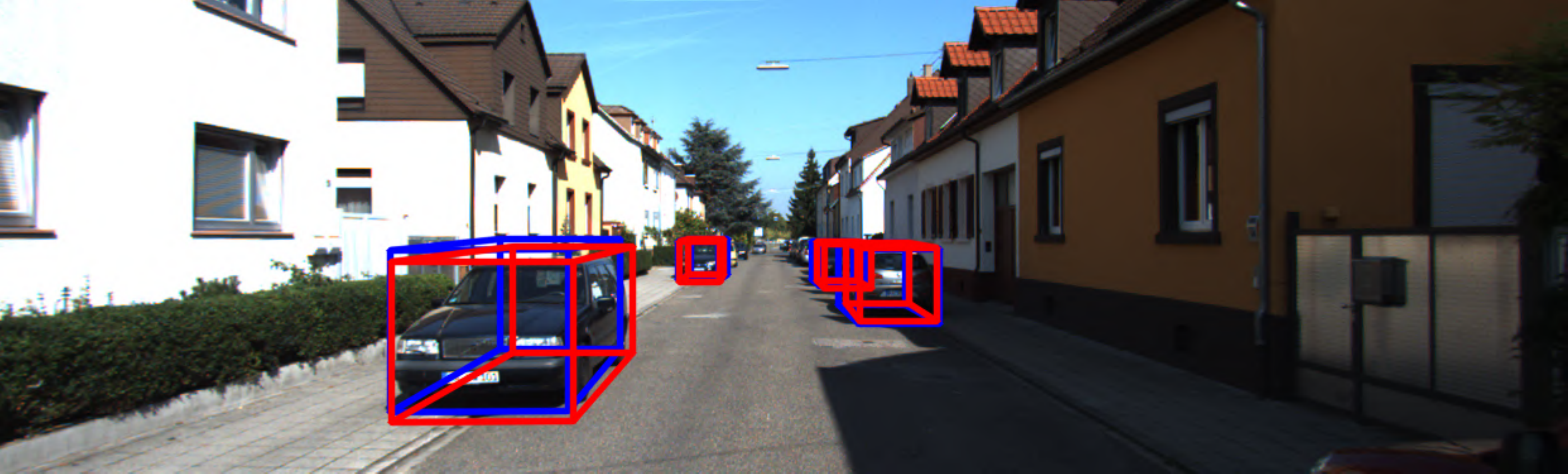}
\includegraphics[width=0.32\linewidth]{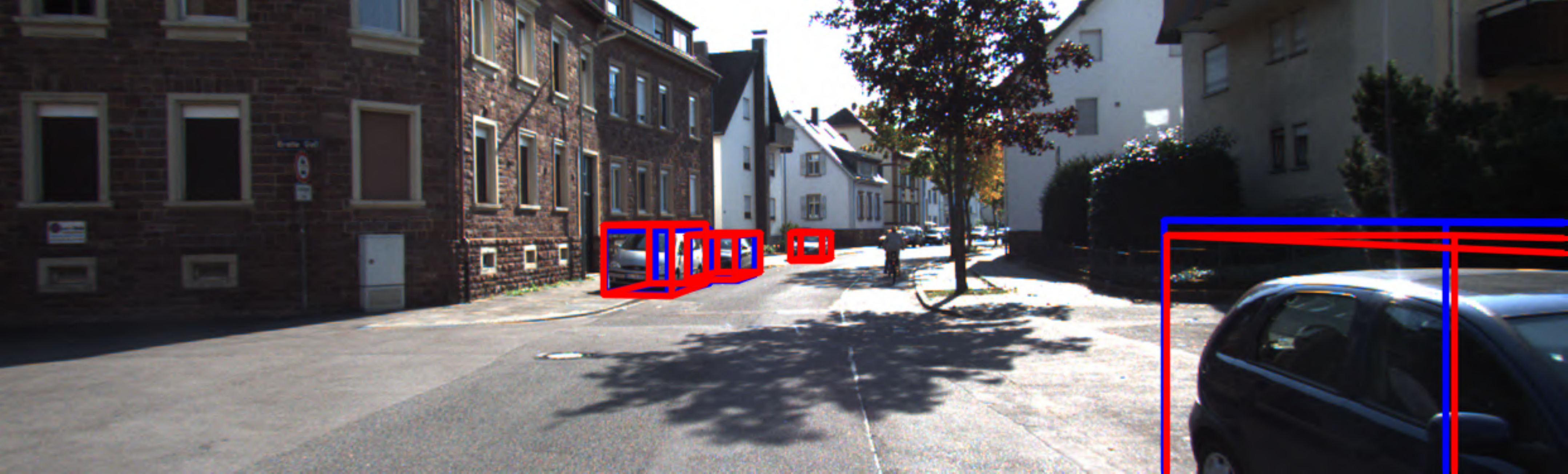}
\includegraphics[width=0.32\linewidth]{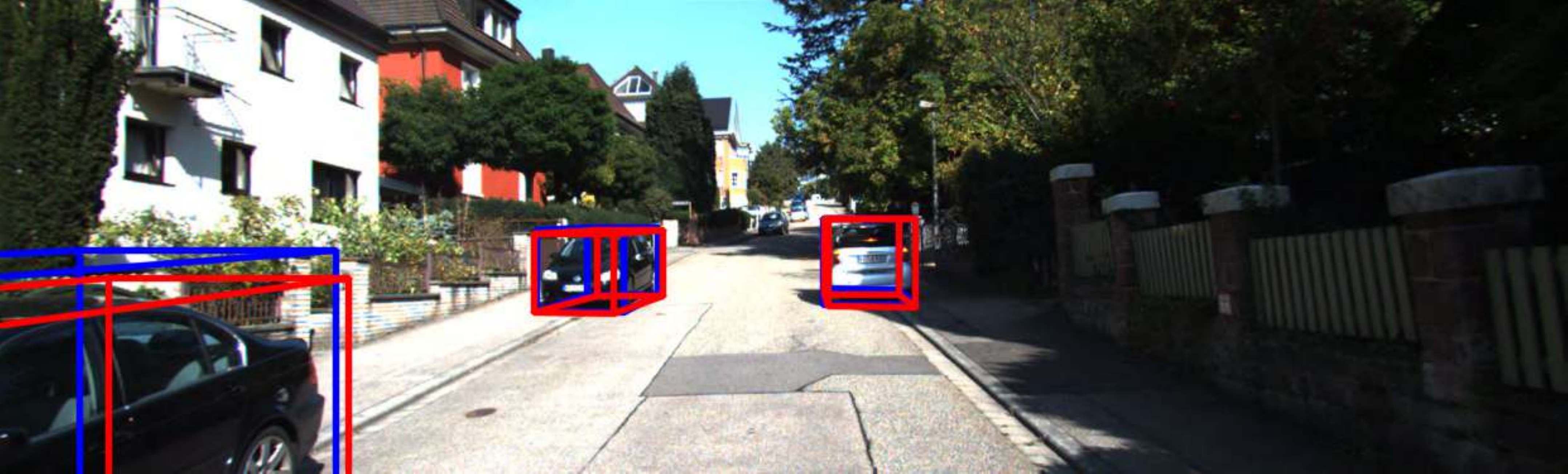}
\includegraphics[trim=4cm 2cm 0cm 3cm, clip=true, width=0.32\linewidth]{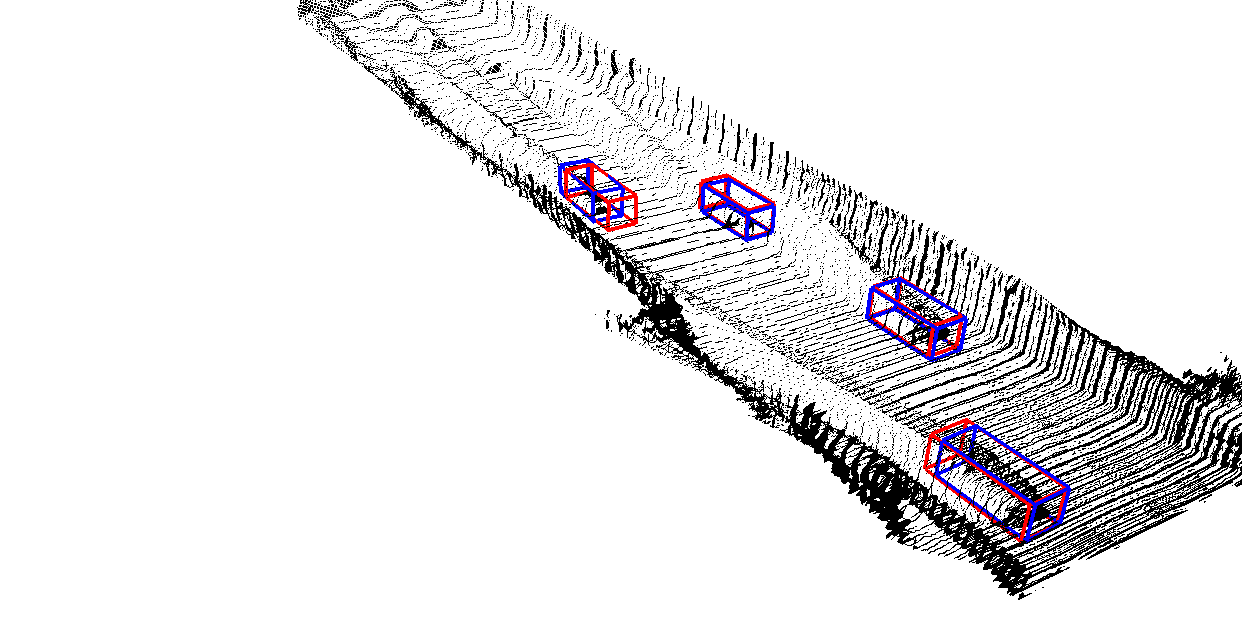}
\includegraphics[trim=4cm 2cm 0cm 3cm, clip=true, width=0.32\linewidth]{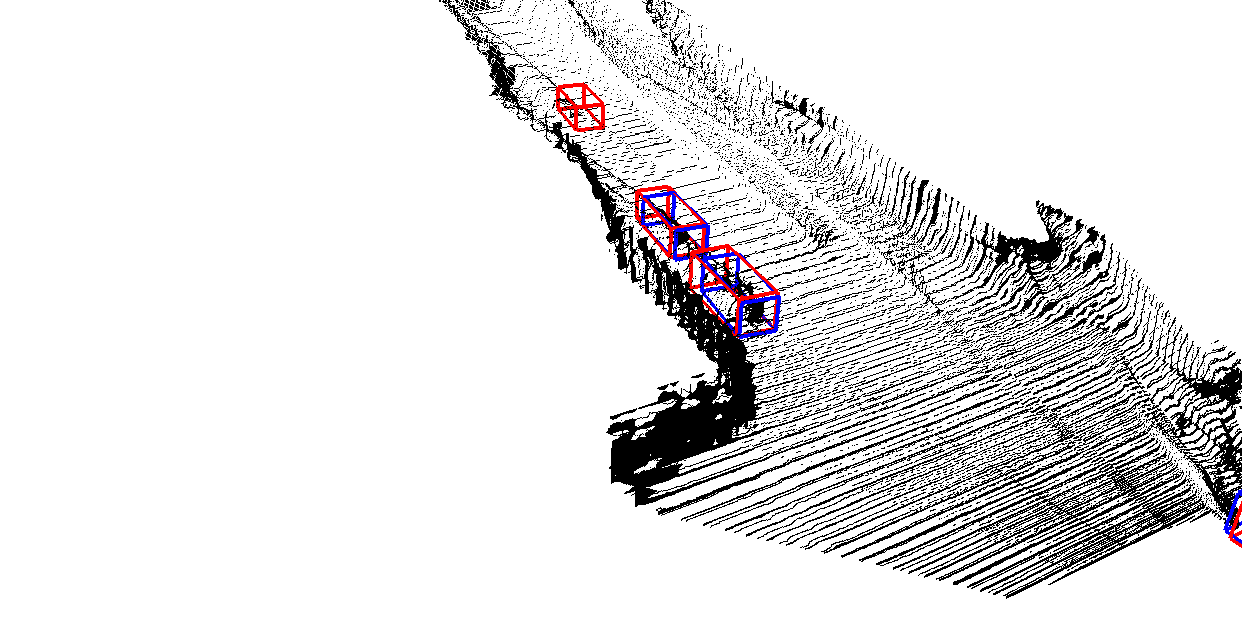}
\includegraphics[trim=4cm 2cm 0cm 3cm, clip=true, width=0.32\linewidth]{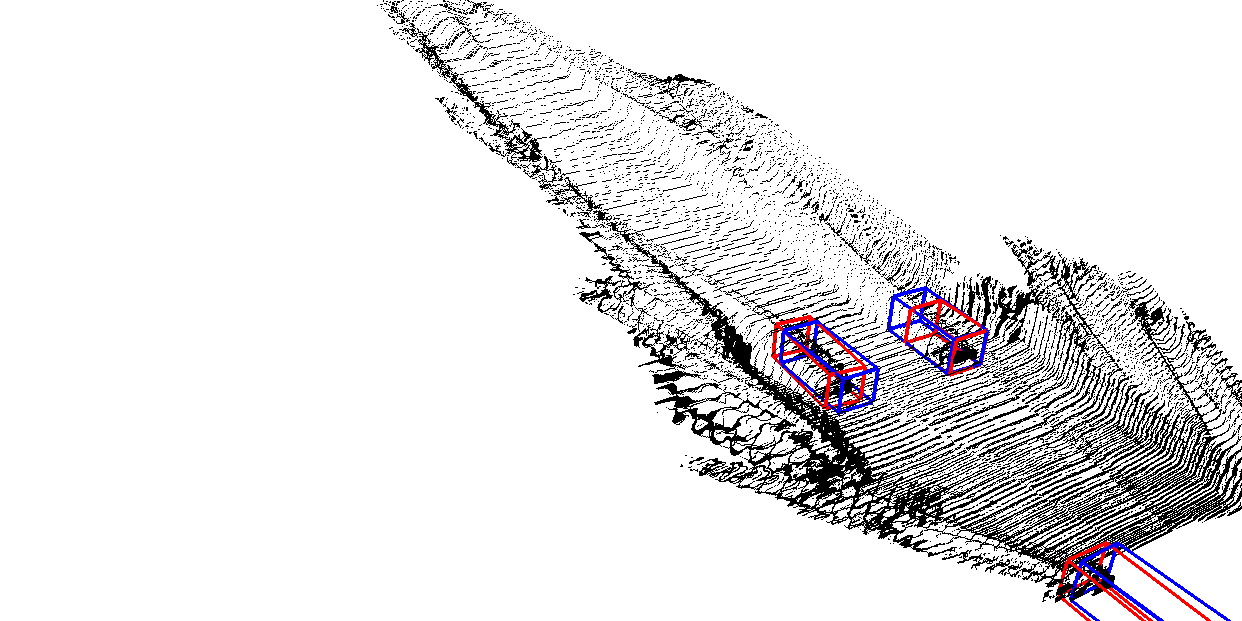}
   
\vspace{0.15cm}
\includegraphics[width=0.32\linewidth]{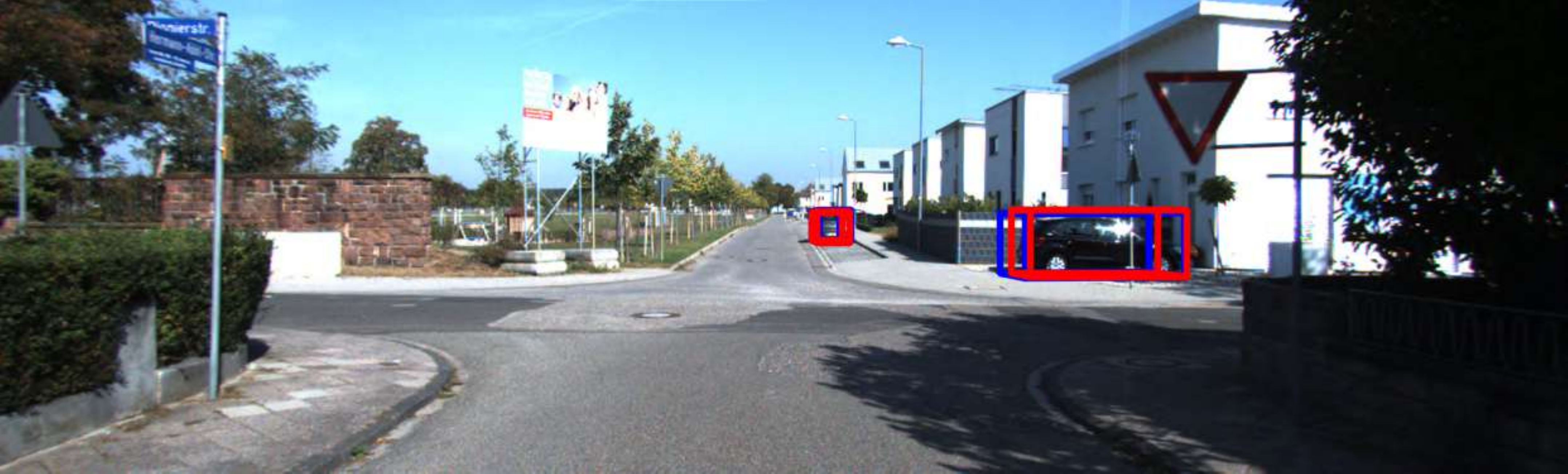}
\includegraphics[width=0.32\linewidth]{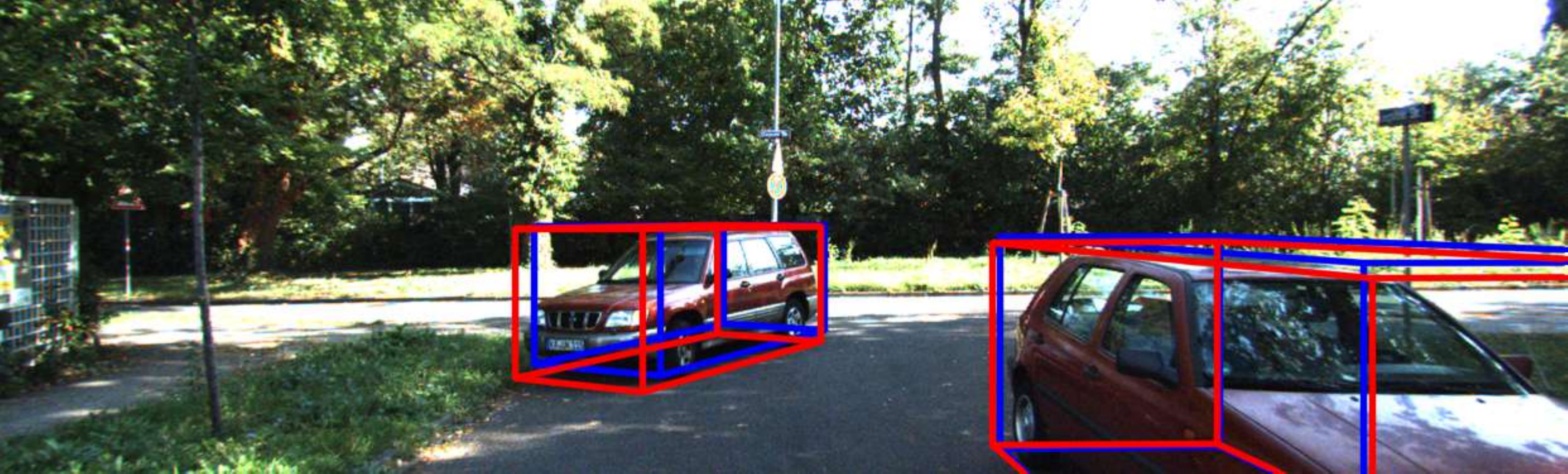}
\includegraphics[width=0.32\linewidth]{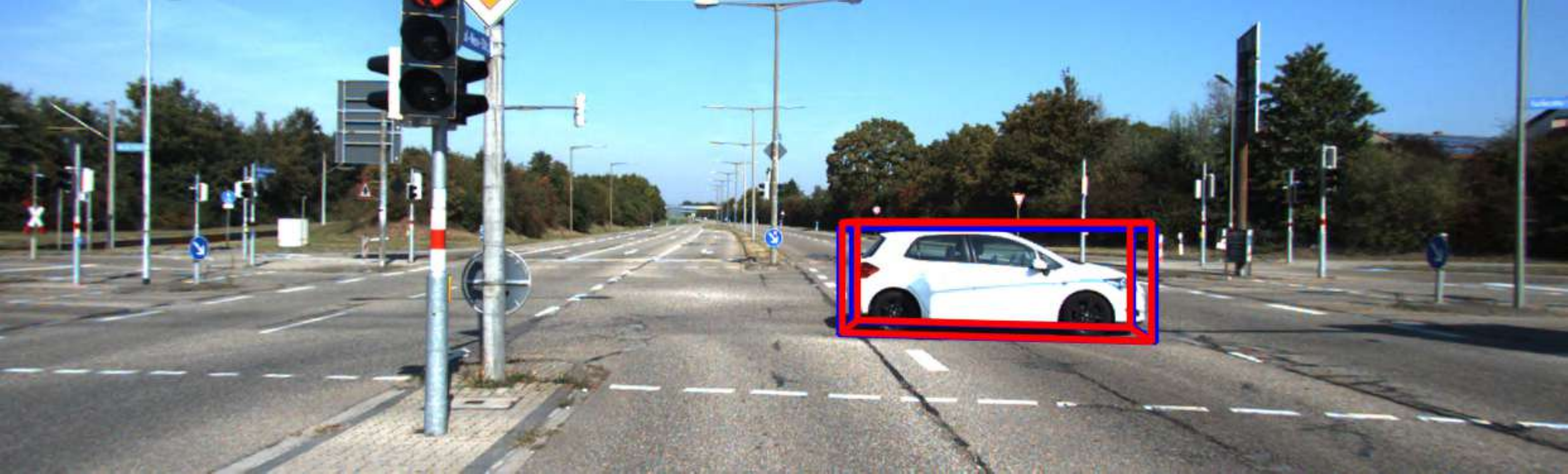}
\includegraphics[trim=4cm 2cm 0cm 3cm, clip=true, width=0.32\linewidth]{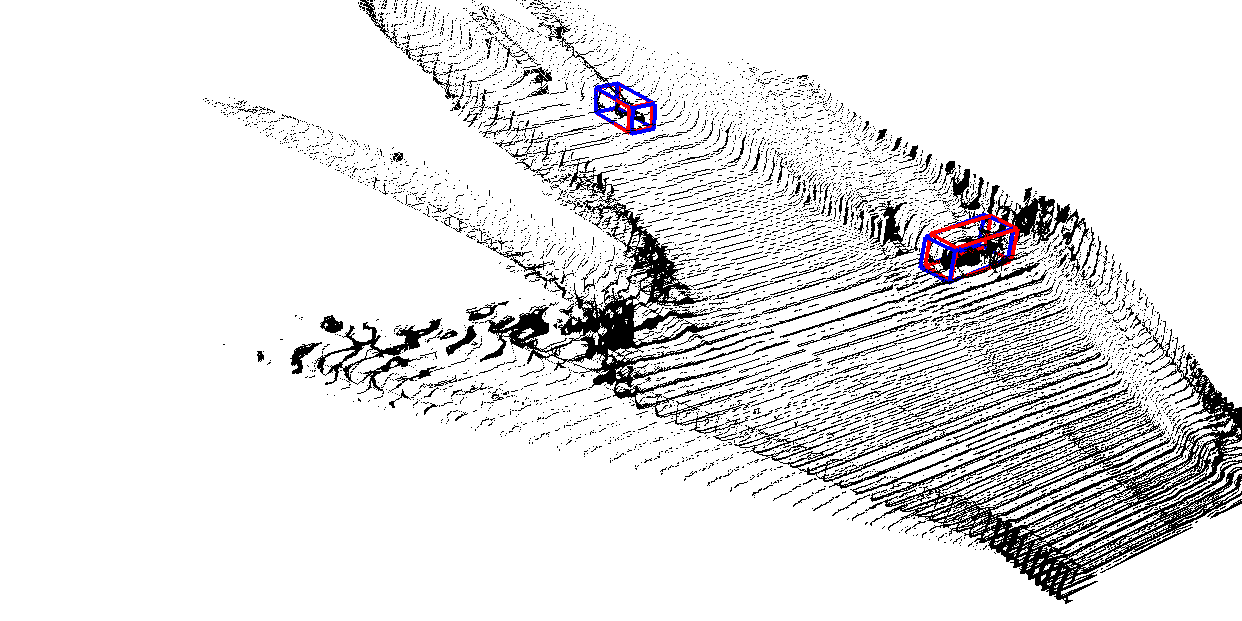}
\includegraphics[trim=4cm 2cm 0cm 3cm, clip=true, width=0.32\linewidth]{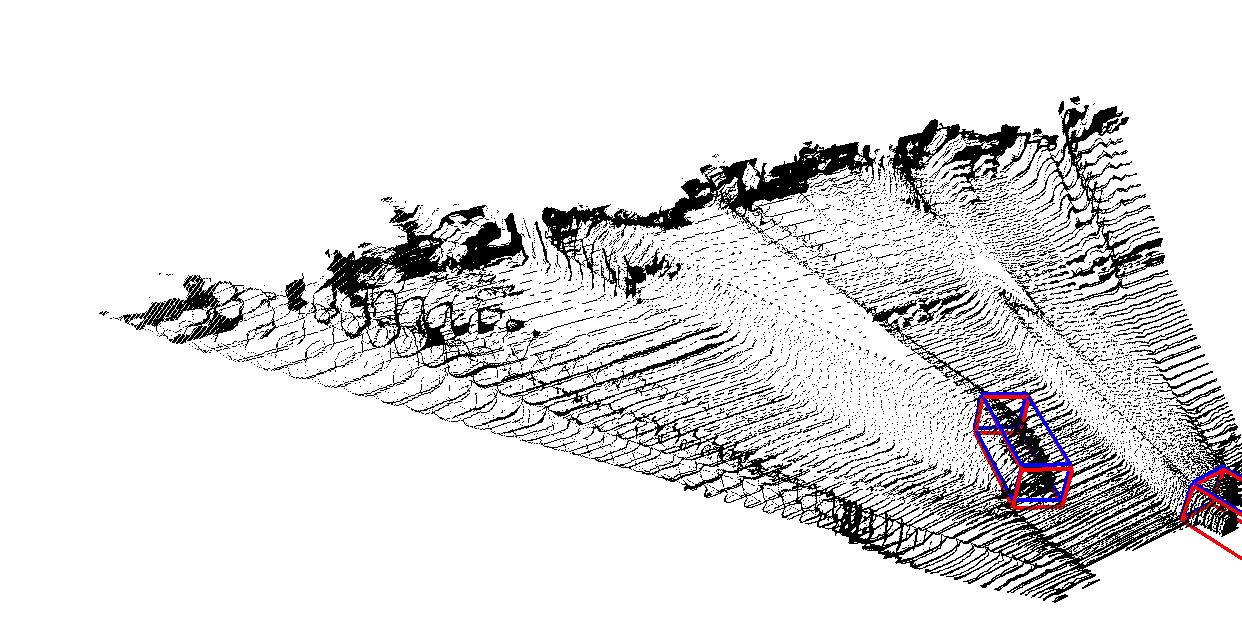}
\includegraphics[trim=4cm 2cm 0cm 3cm, clip=true, width=0.32\linewidth]{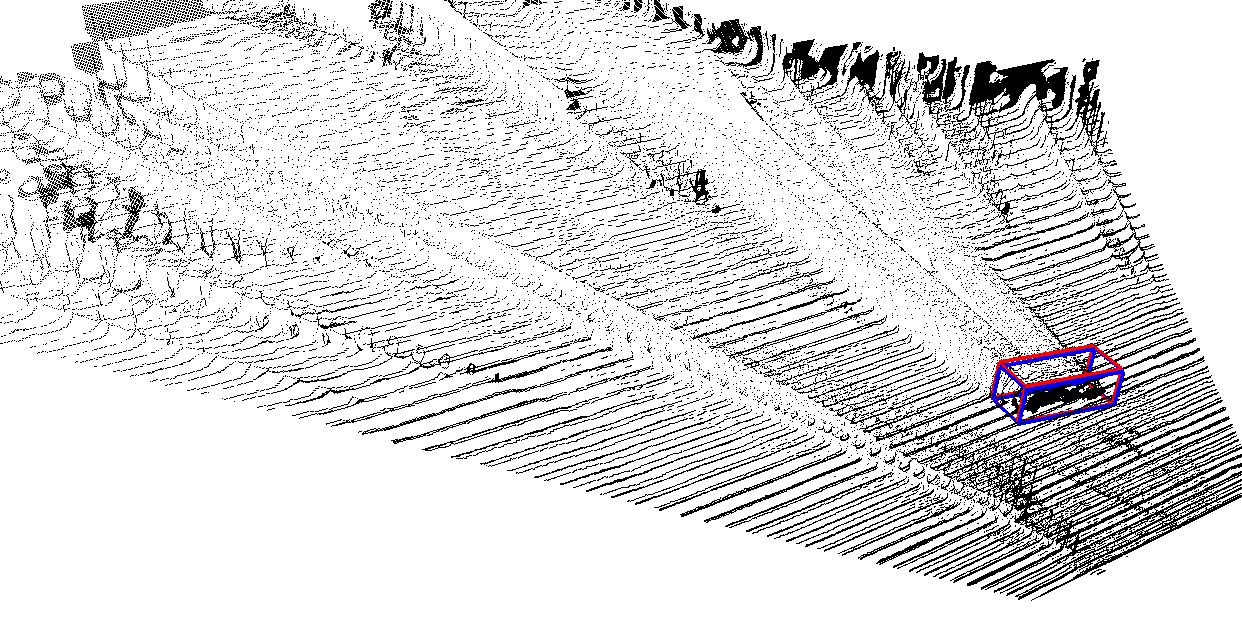}

\vspace{0.15cm}
\includegraphics[width=0.32\linewidth]{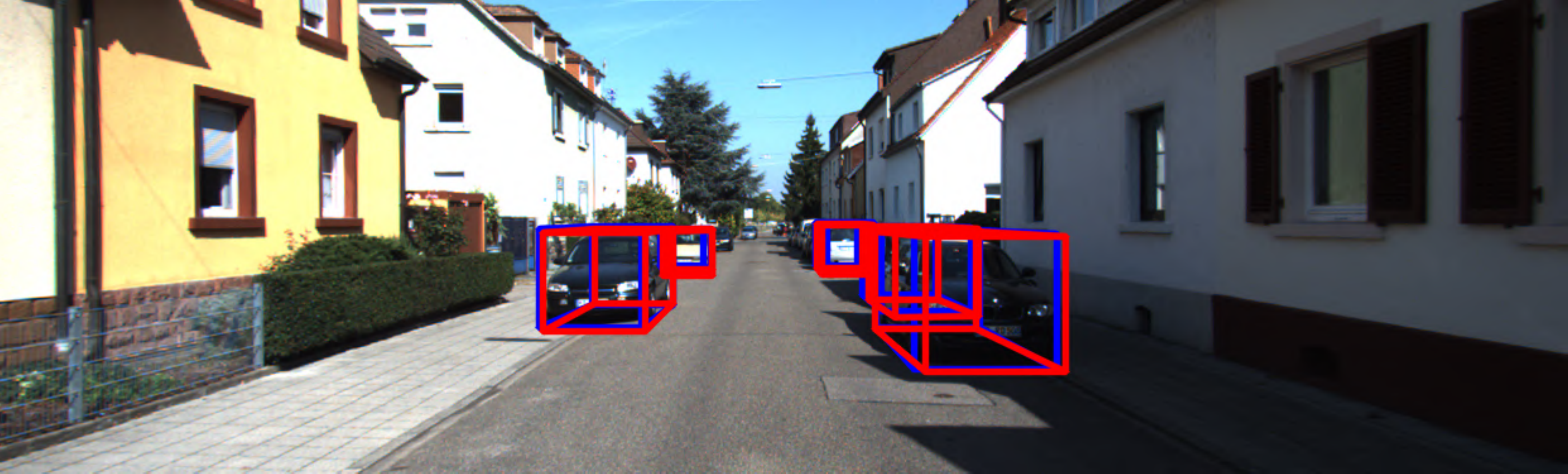}
\includegraphics[width=0.32\linewidth]{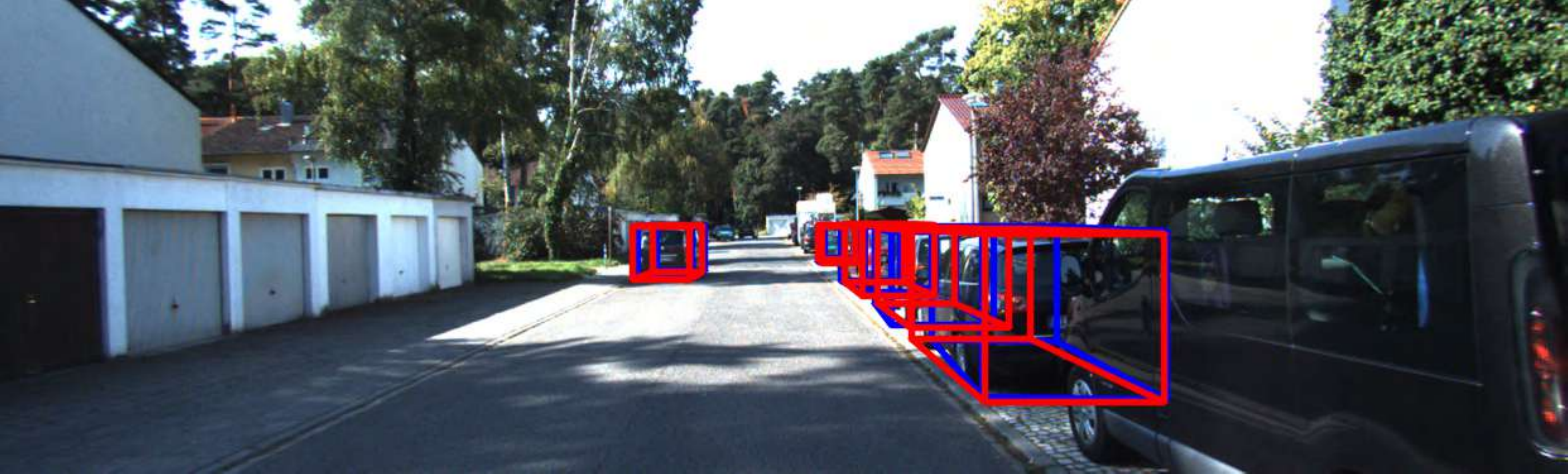}
\includegraphics[width=0.32\linewidth]{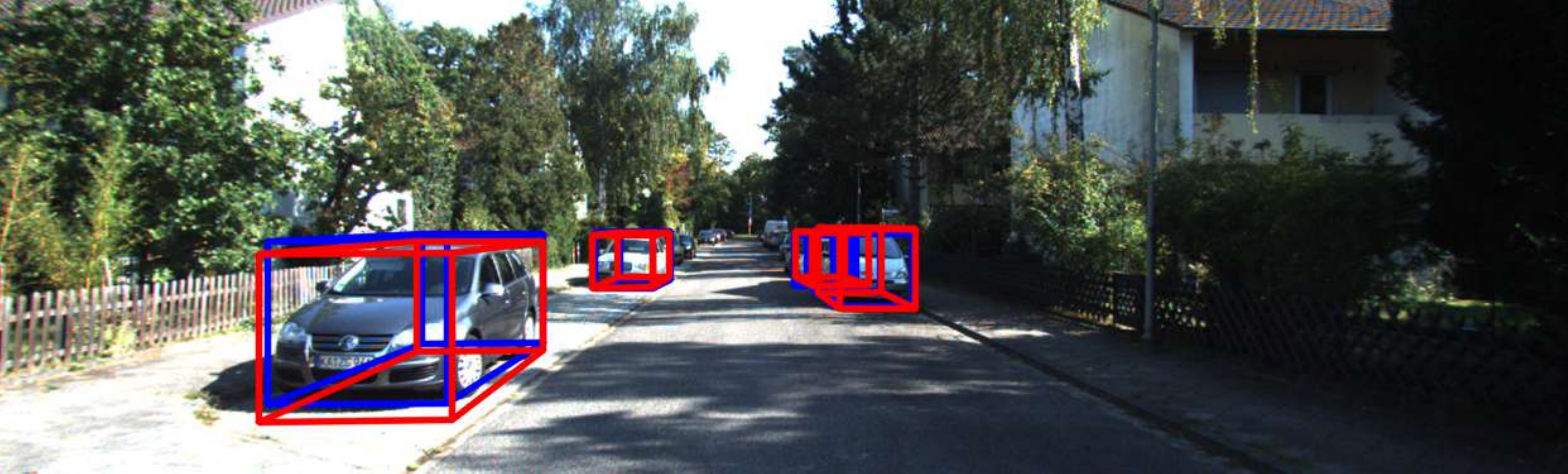}
\includegraphics[trim=4cm 2cm 0cm 3cm, clip=true, width=0.32\linewidth]{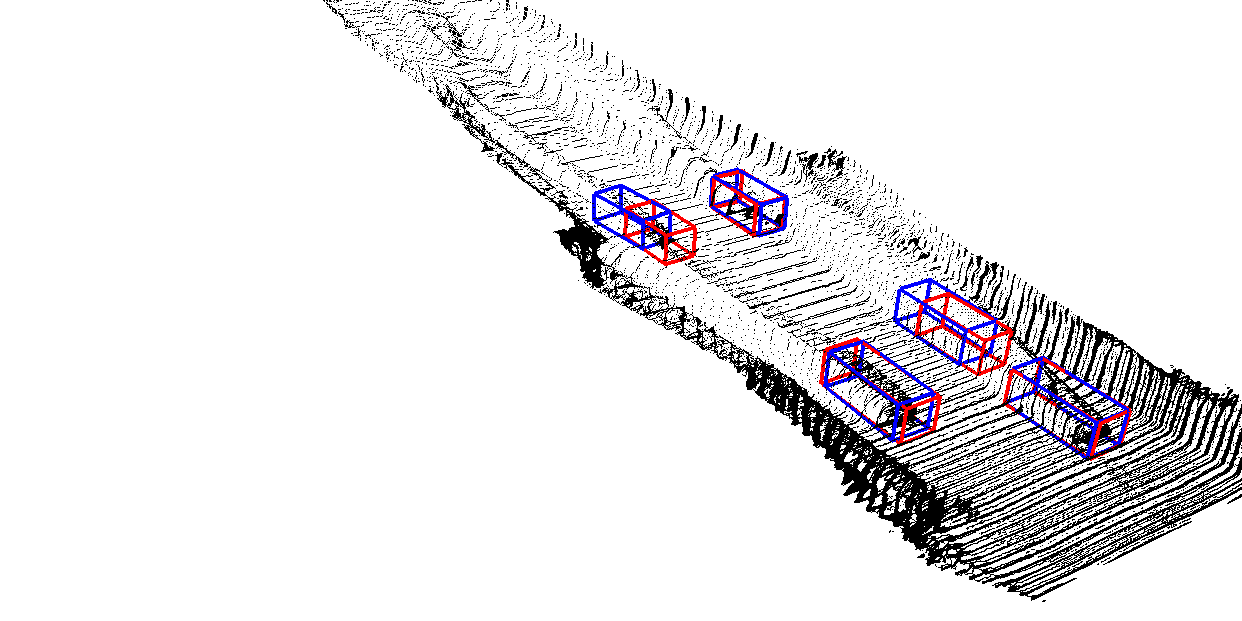}
\includegraphics[trim=4cm 2cm 0cm 3cm, clip=true, width=0.32\linewidth]{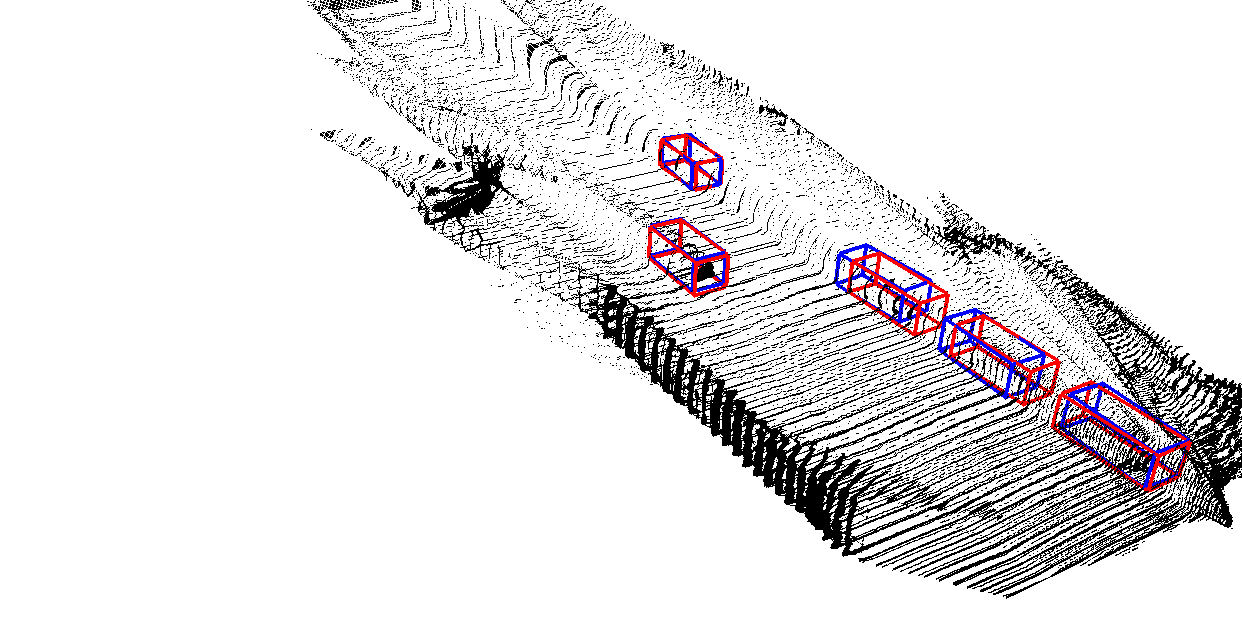}
\includegraphics[trim=4cm 2cm 0cm 3cm, clip=true, width=0.32\linewidth]{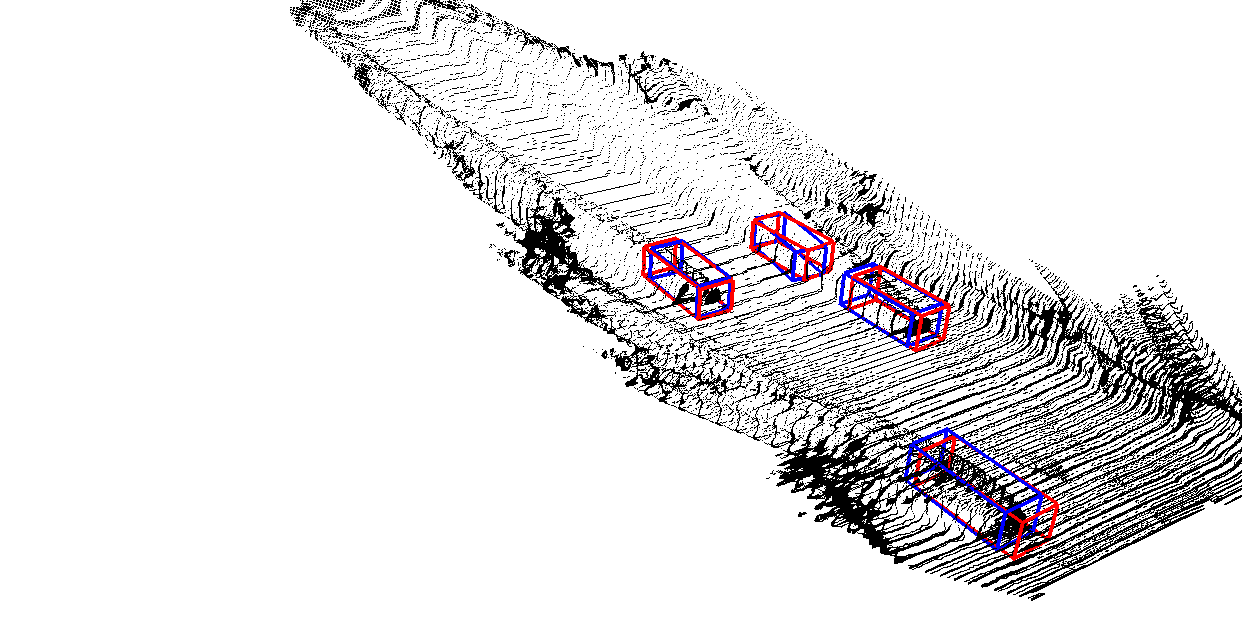}

\vspace{0.15cm}
\includegraphics[width=0.32\linewidth]{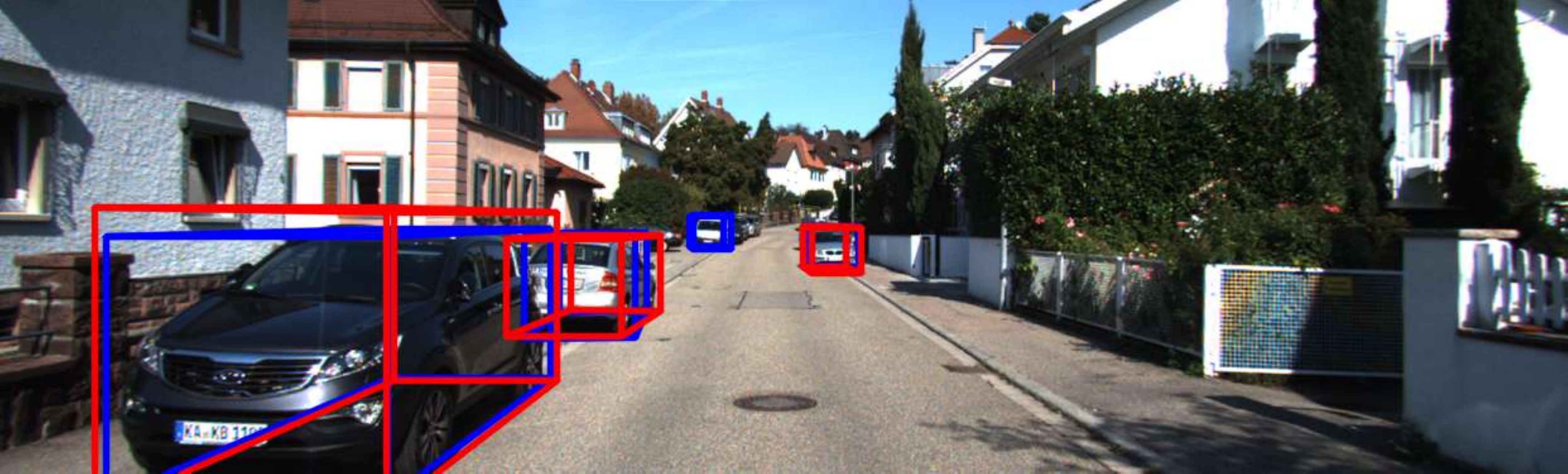}
\includegraphics[width=0.32\linewidth]{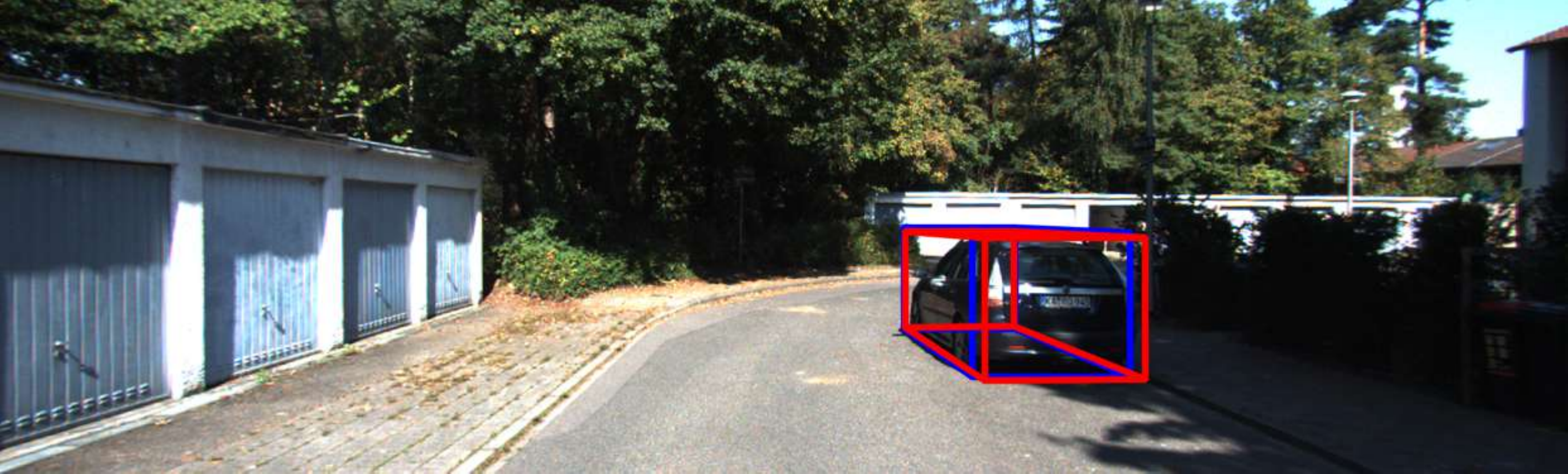}
\includegraphics[width=0.32\linewidth]{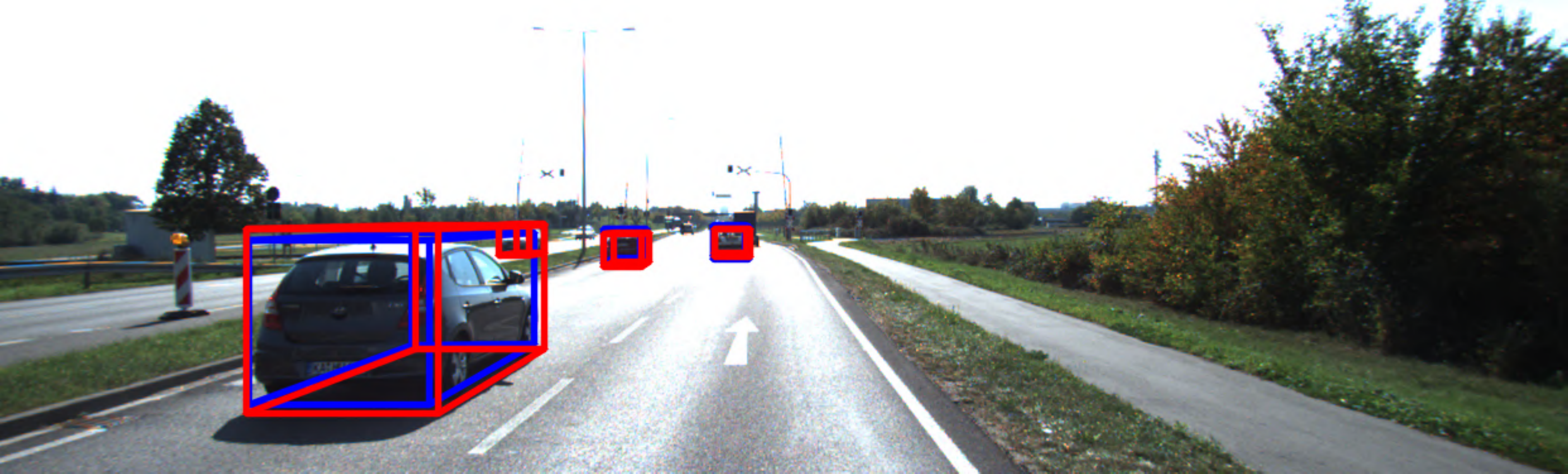}
\includegraphics[trim=4cm 2cm 0cm 3cm, clip=true, width=0.32\linewidth]{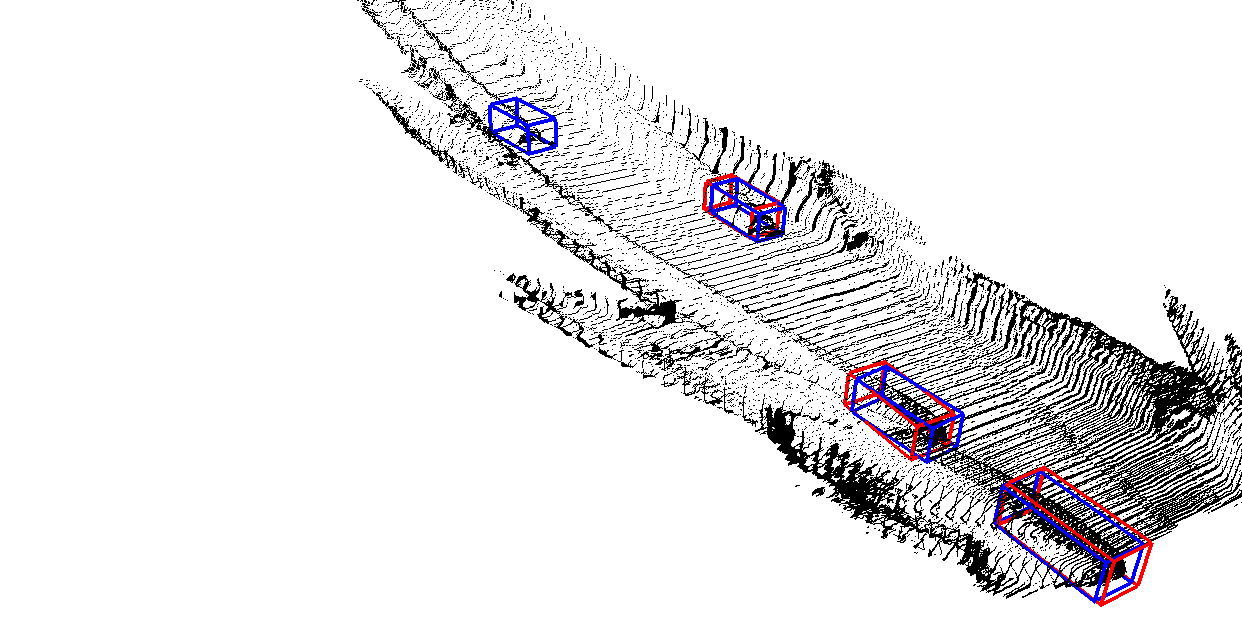}
\includegraphics[trim=4cm 2cm 0cm 3cm, clip=true, width=0.32\linewidth]{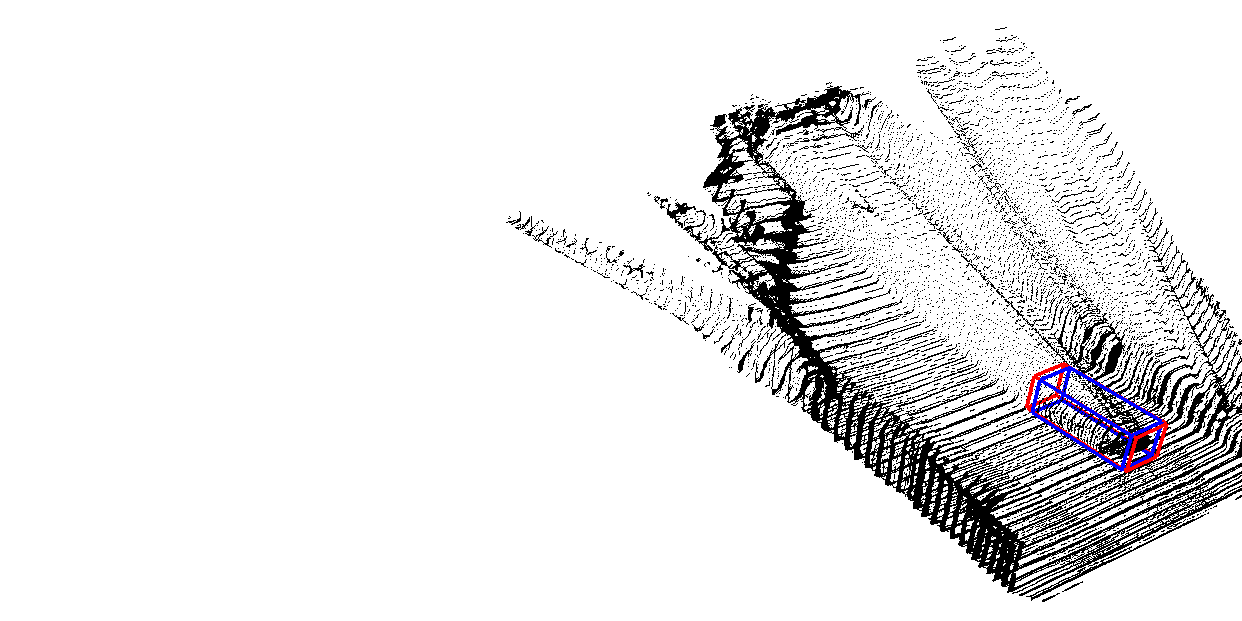}
\includegraphics[trim=4cm 2cm 0cm 3cm, clip=true, width=0.32\linewidth]{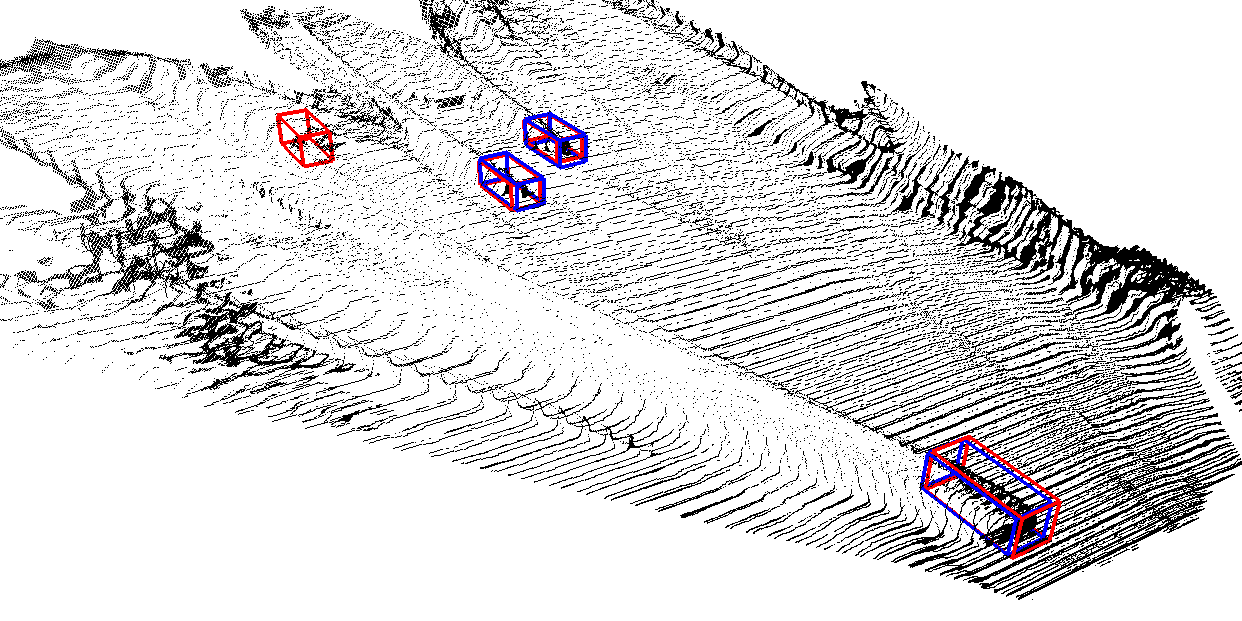}
\end{center}
\vspace{-0.2cm}
\caption{Additional qualitative results of our method on KITTI \textbf{val} set. We visualize our 3D bounding box estimate (in \textbf{\textcolor{blue}{blue}}) and ground truth (in \textbf{\textcolor{red}{red}}) on the frontal images (1st and 3rd rows) and pseudo-LiDAR point cloud (2nd and 4th rows).}
\vspace{-0.3cm}
\label{fig:qua}
\end{figure*}

\begin{figure*}[ht]
\begin{center}
\includegraphics[trim=0cm 10cm 24cm 0cm, clip=true, width=0.04\linewidth]{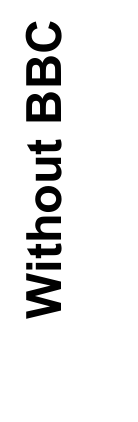}
\includegraphics[trim=12cm 12.5cm 12cm 0cm, clip=true, width=0.31\linewidth]{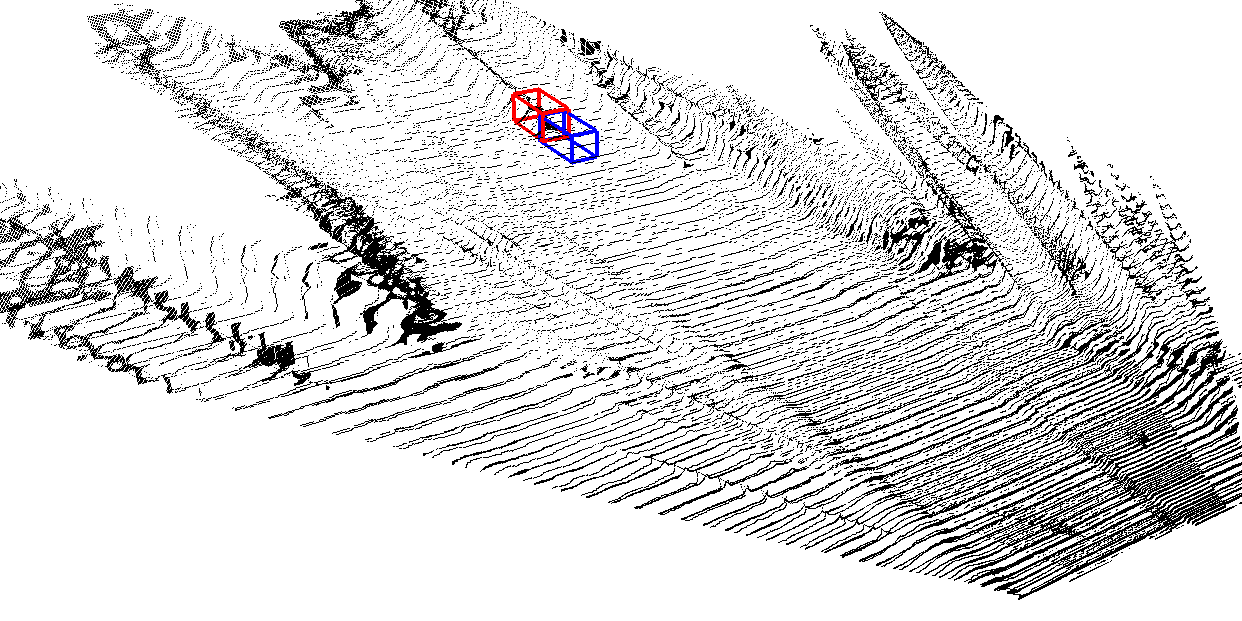}
\includegraphics[trim=20cm 7.5cm 4cm 5cm, clip=true, width=0.31\linewidth]{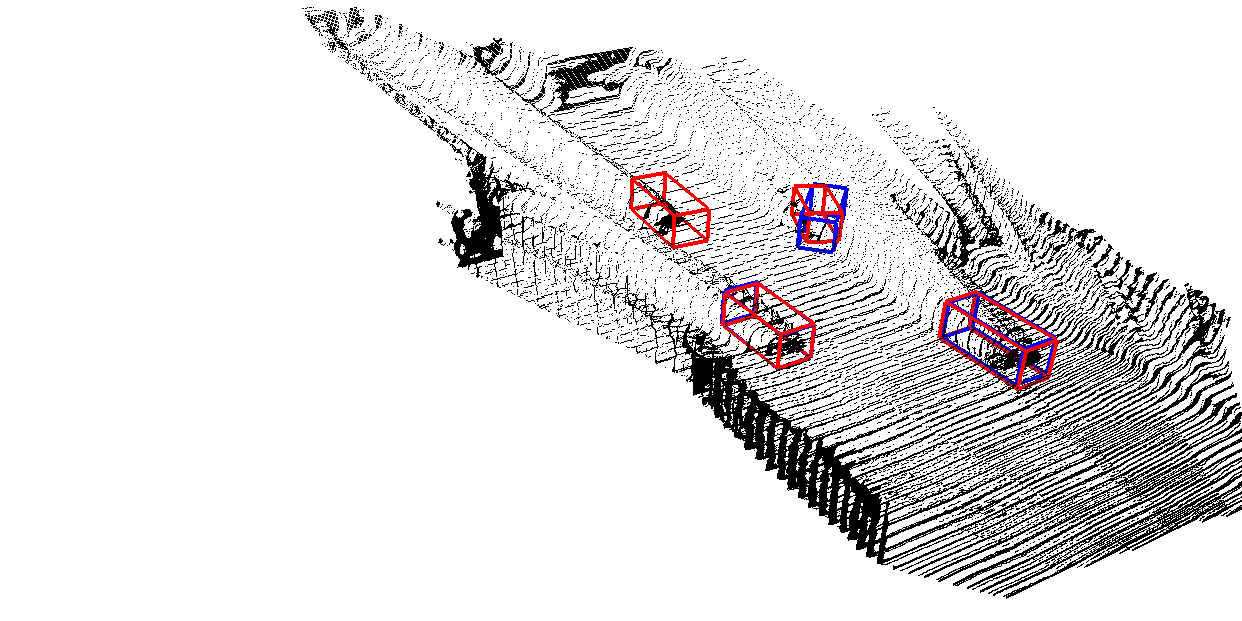}
\includegraphics[trim=6cm 4cm 14cm 7cm, clip=true, width=0.31\linewidth]{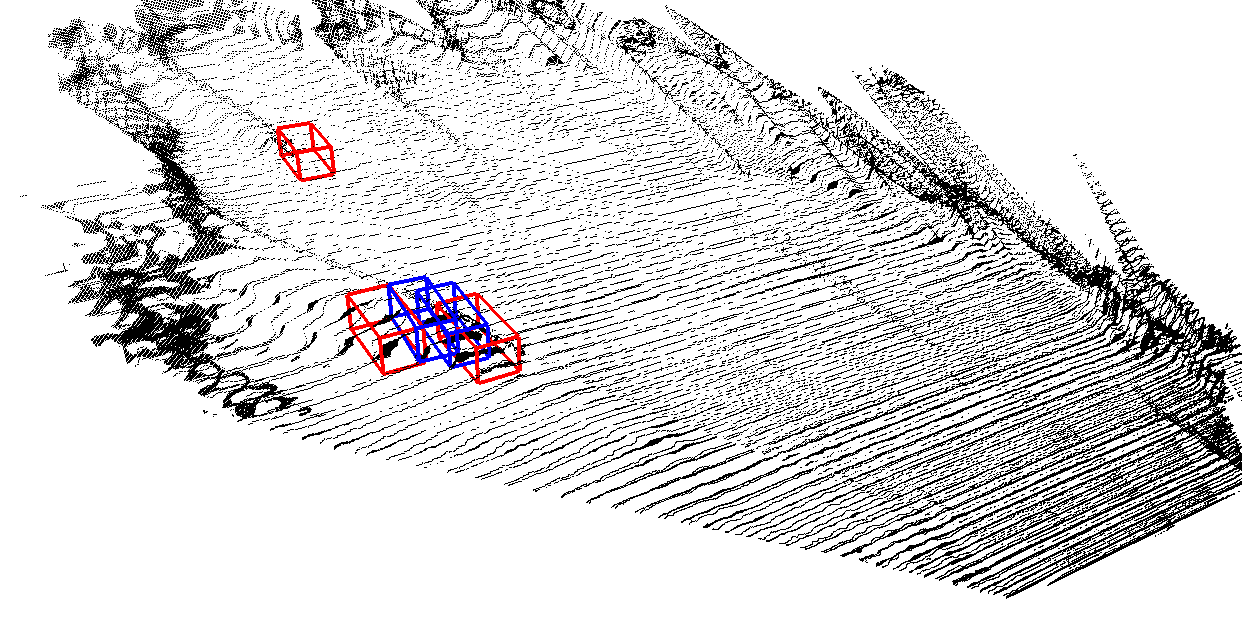}\\
\includegraphics[trim=0cm 10cm 24cm 0cm, clip=true, width=0.04\linewidth]{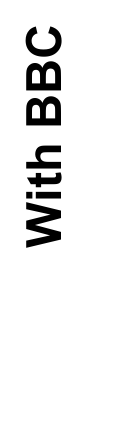}
\includegraphics[trim=12cm 12.5cm 12cm 0cm, clip=true, width=0.31\linewidth]{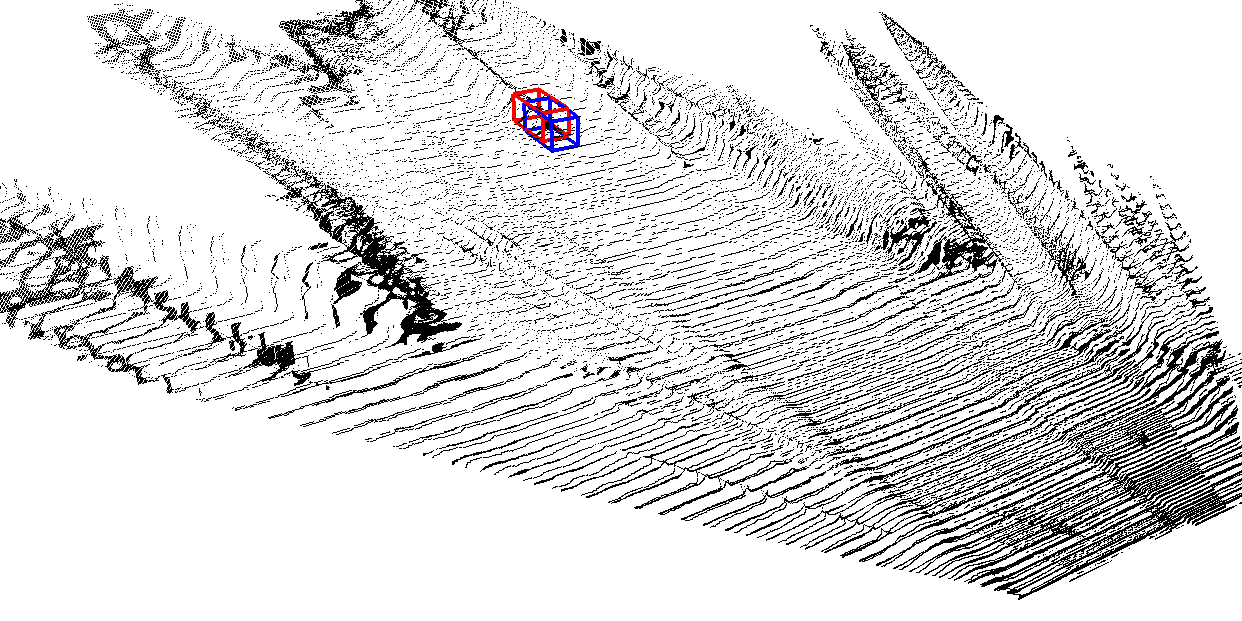}
\includegraphics[trim=20cm 7.5cm 4cm 5cm, clip=true, width=0.31\linewidth]{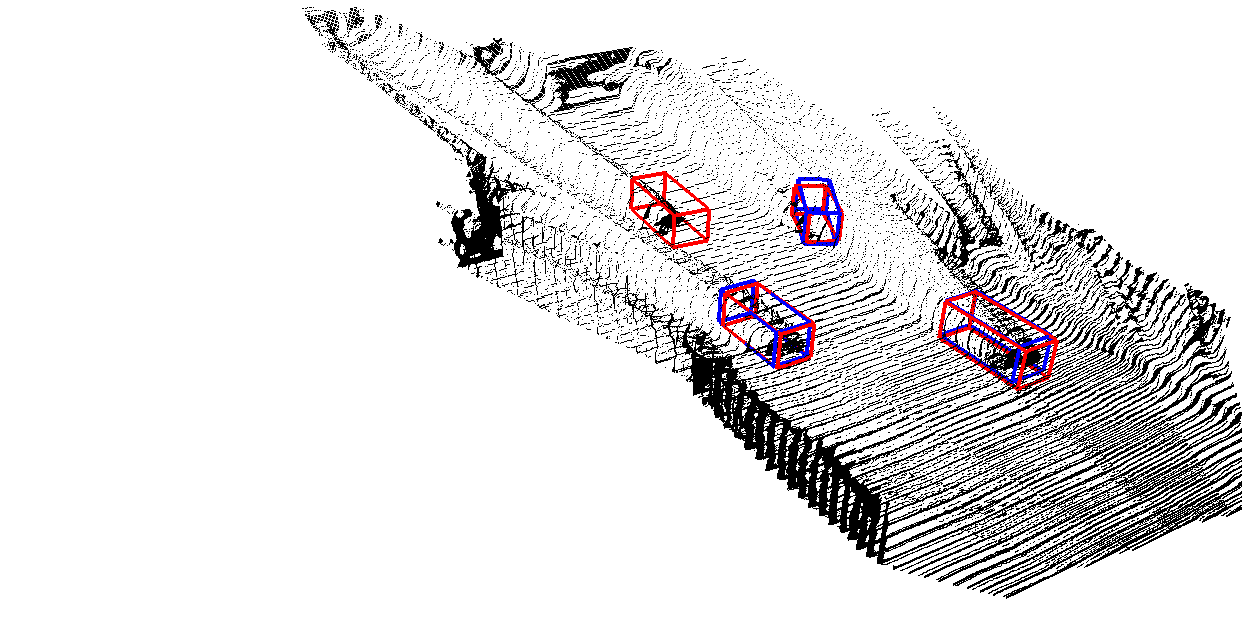}
\includegraphics[trim=6cm 4cm 14cm 7cm, clip=true, width=0.31\linewidth]{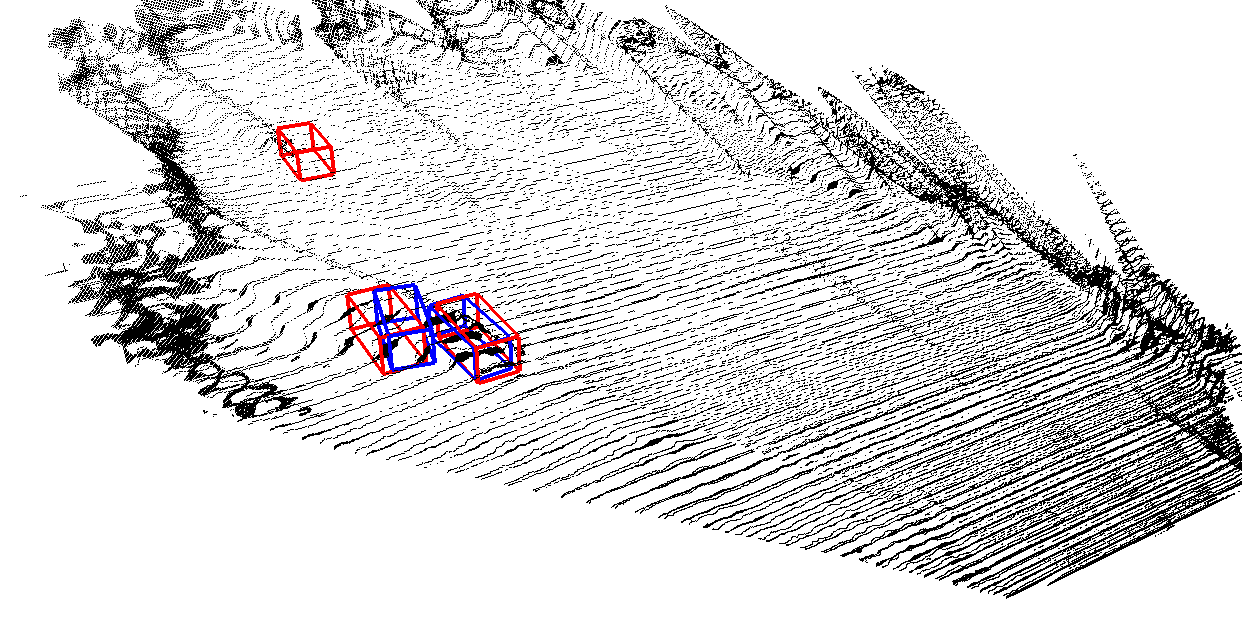}
   
\vspace{0.15cm}
\includegraphics[trim=0cm 10cm 24cm 0cm, clip=true, width=0.04\linewidth]{supp/bbc_more/without_BBC.pdf}
\includegraphics[trim=10cm 6cm 14cm 7cm, clip=true, width=0.31\linewidth]{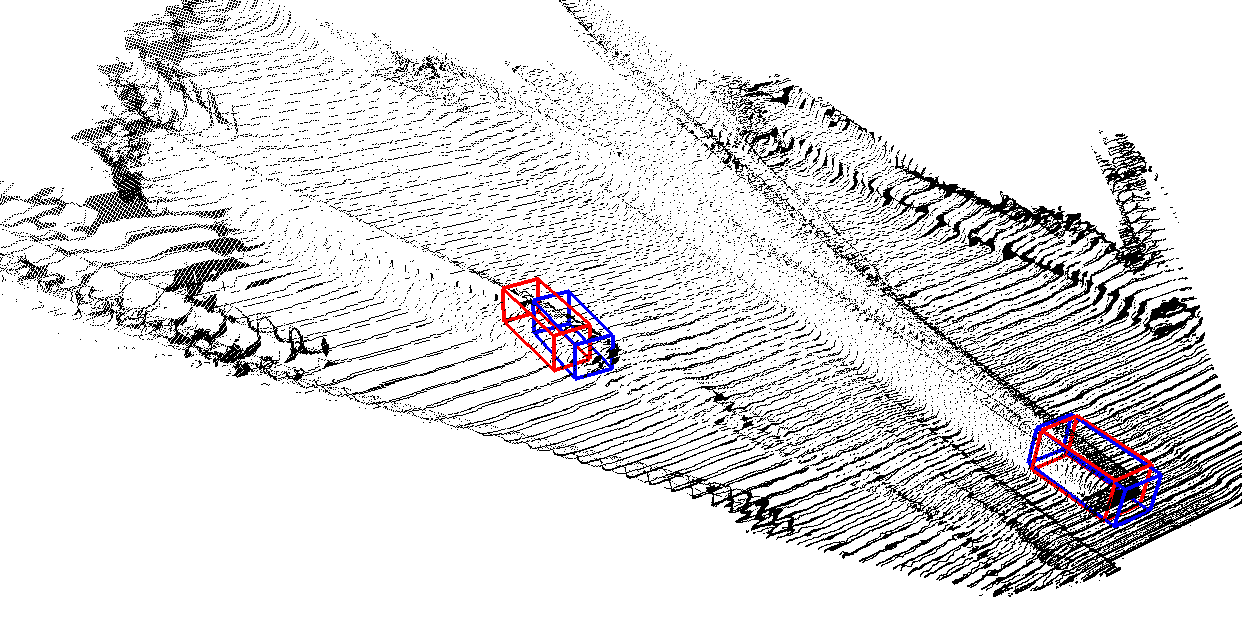}
\includegraphics[trim=11cm 10cm 14cm 3.3cm, clip=true, width=0.31\linewidth]{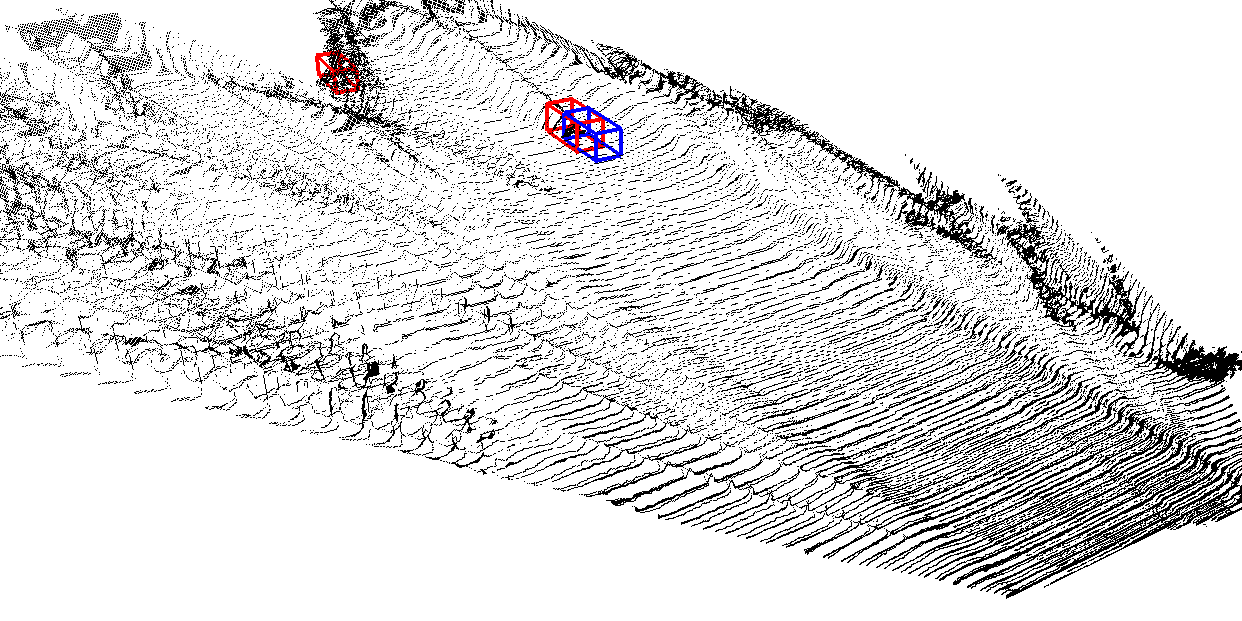}
\includegraphics[trim=8cm 3.3cm 17cm 10cm, clip=true, width=0.31\linewidth]{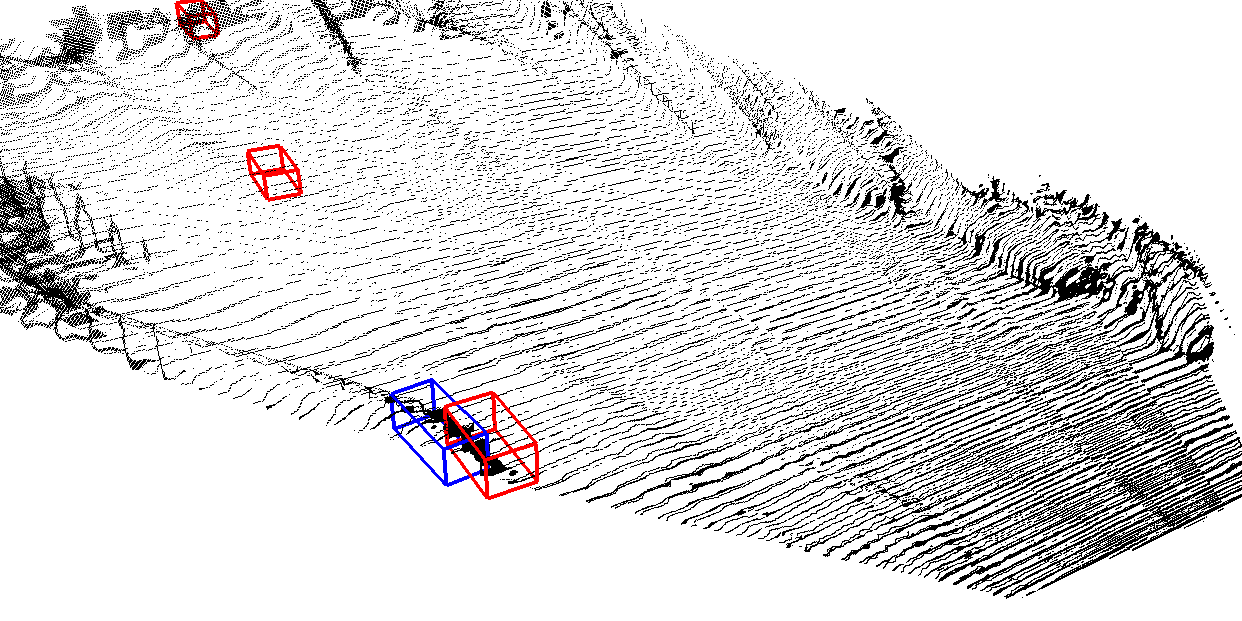}\\
\includegraphics[trim=0cm 10cm 24cm 0cm, clip=true, width=0.04\linewidth]{supp/bbc_more/with_BBC.pdf}
\includegraphics[trim=10cm 6cm 14cm 7cm, clip=true, width=0.31\linewidth]{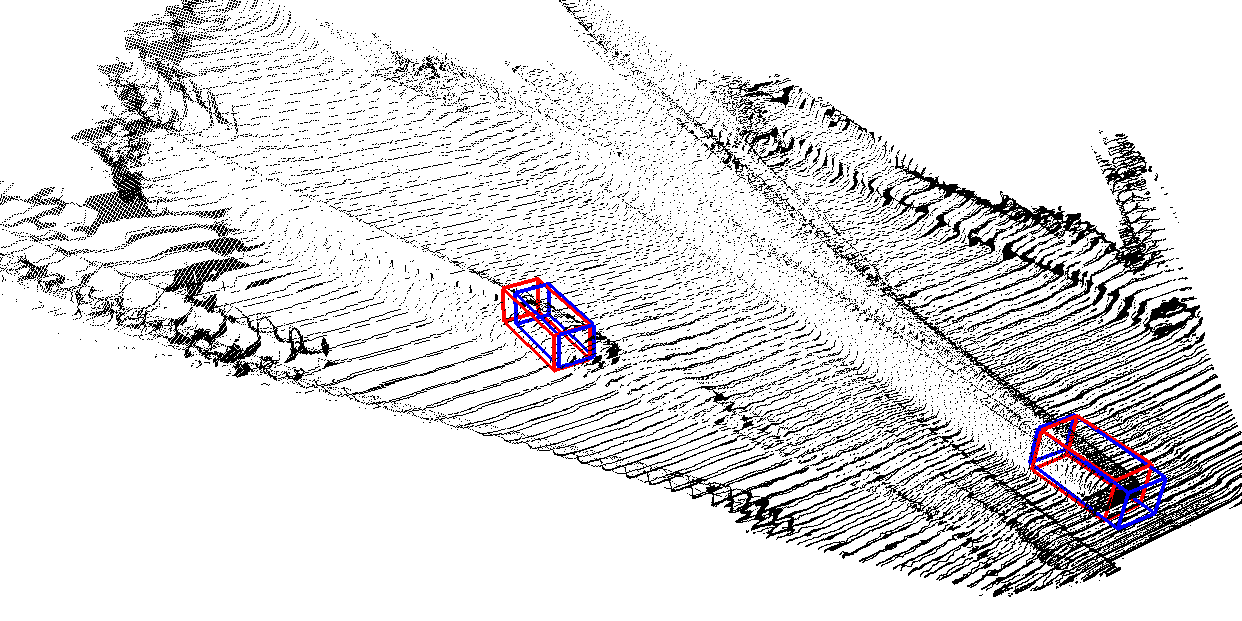}
\includegraphics[trim=11cm 10cm 14cm 3.3cm, clip=true, width=0.31\linewidth]{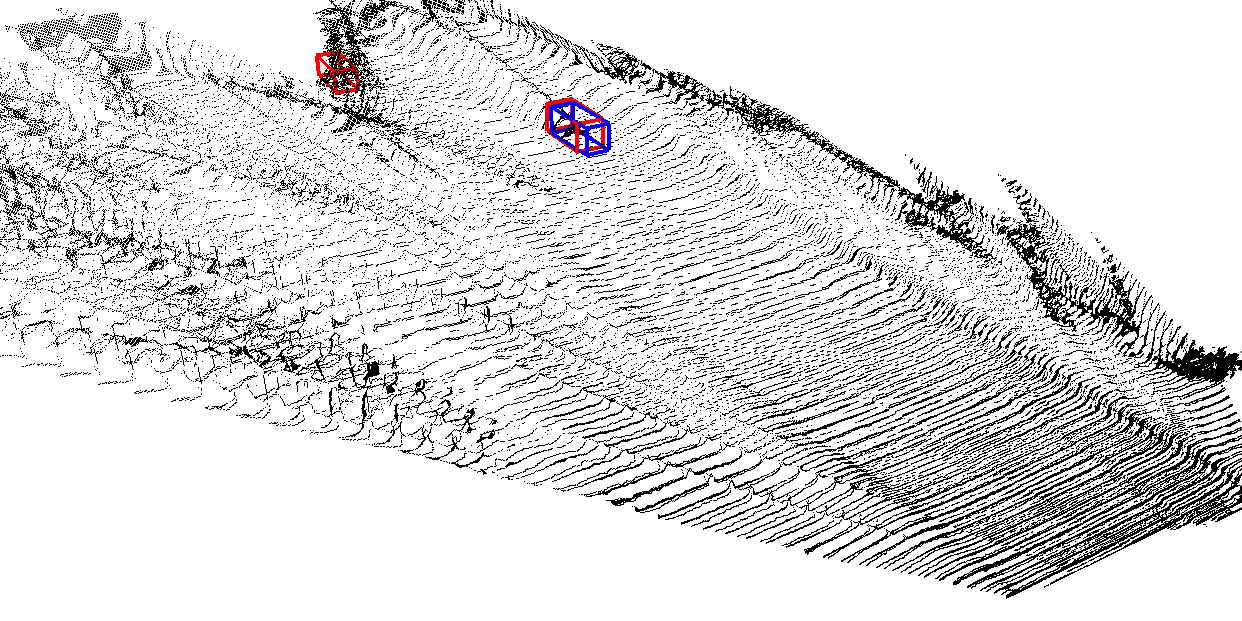}
\includegraphics[trim=8cm 3.3cm 17cm 10cm, clip=true, width=0.31\linewidth]{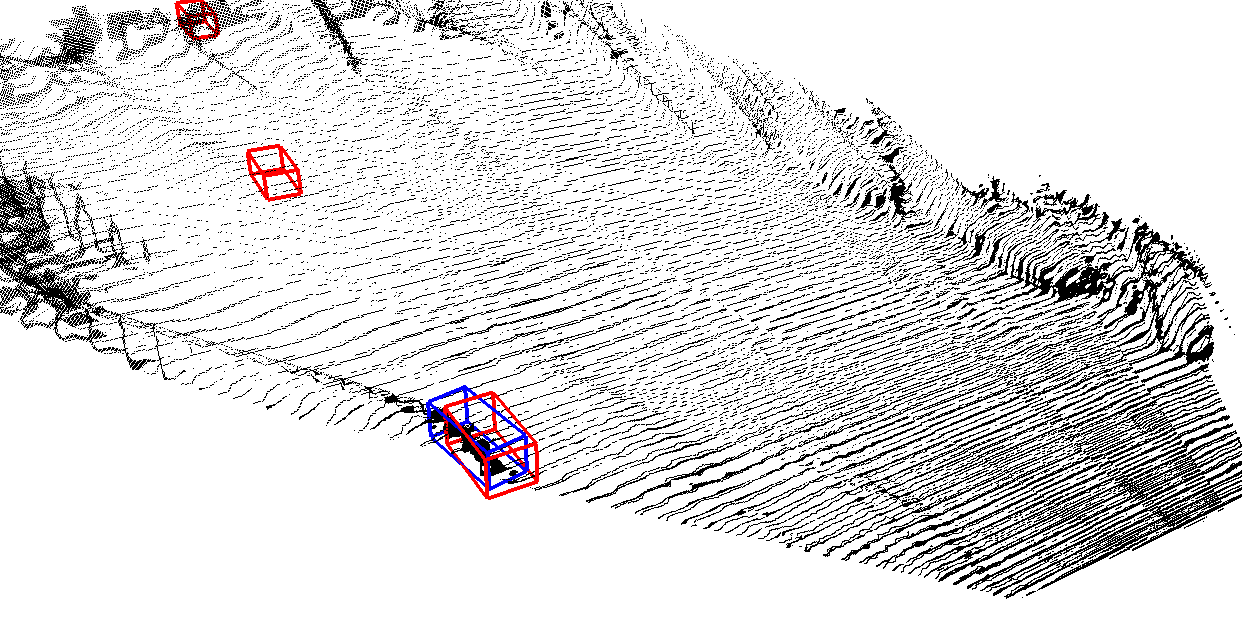}  
\end{center}
\vspace{-0.5cm}
\caption{\textbf{Additional Visualization about the Effect of Bounding Box Consistency (BBC)}. We visualize our 3D bounding box estimate (\textbf{\textcolor{blue}{blue}}) without BBC (in 1st and 3rd rows) and with BBC (in 2nd and 4th rows). Ground truth is shown in \textbf{\textcolor{red}{red}}. We show that using the bounding box consistency improves the 3D IoU between the 3D box estimate and the ground truth.}
\vspace{-0.3cm}
\label{fig:bbc}
\end{figure*}

\vspace{-0.1cm}
\section{Amodal 3D Object Detection}
\vspace{-0.1cm}


\noindent\textbf{Network Architecture of the 3D Box Correction Module}

We show the network architecture in Figure \ref{fig:architecture}. Similar to the 3D box estimation module proposed in \cite{Qi2018}, we also use a PointNet-based network for our 3D box correction module. The major difference is that the 3D box estimation module predicts the 3D box parameters while our 3D box correction module outputs the correction to the prediction (\emph{i.e.}, residual of the parameters). In addition, we concatenate the features extracted from the 3D box estimation module with the global feature extracted from the 3D box correction module for predicting the residual of the parameters.


\vspace{-0.1cm}
\section{2D-3D Bounding Box Consistency (BBC)}



\vspace{-0.1cm}\noindent\textbf{Additional Visualization about the Effect of Bounding Box Consistency}

We provide the extra visual comparison between the 3D bounding box estimate with and without using the BBC in Figure \ref{fig:bbc}. The 3D bounding box results shown in the 2nd and 4th rows, which are estimated from the model trained with bounding box consistency loss and post-processed with bounding box consistency optimization, clearly improve the 3D IoU over the 3D bounding box results without using the BBC, shown in the 1st and 3rd rows.


\vspace{-0.1cm}
\section{Experiments}
\vspace{-0.1cm}

\noindent\textbf{Additional Visualization of 3D Object Detection Results}

We provide additional qualitative results in Figure \ref{fig:qua}. We show that, from only a single RGB image, the 3D bounding box detection for the car category can be very accurate, even for the challenging faraway objects (\emph{e.g.}, in the 6th row 1st column and 8th row 2nd column of the Figure \ref{fig:qua}).

{\small
\bibliographystyle{ieee}
\bibliography{supplementary}
}